\documentclass[%
 reprint,
 amsmath,amssymb,
 aps,
]{revtex4-2}

\usepackage{amsfonts}
\usepackage{amsmath}
\usepackage{amssymb}

\usepackage[utf8]{inputenc}
\usepackage[english]{babel}

\newcommand{\w}{{\mathbf w}}
\newcommand{\x}{{\mathbf x}}

\newcommand{\cC}{\mathcal{C}}

\newcommand{\cF}{\mathcal{F}}

\newcommand{\cH}{\mathcal{H}}

\newcommand{\cL}{\mathcal{L}}

\newcommand{\cN}{\mathcal{N}}

\newcommand{\cQ}{\mathcal{Q}}

\newcommand{\cS}{\mathcal{S}}

\newcommand{\bbE}{\mathbb{E}}

\newcommand{\bbR}{\mathbb{R}}

\newcommand{\Var}{{\mathrm{Var}}}
\newcommand{\cov}{{\mathrm{Cov}}}

\newcommand{\transpose}{\mathsf{T}}

\newcommand{\pa}{\partial}

\usepackage{graphicx}% Include figure files
\usepackage{dcolumn}% Align table columns on decimal point
\usepackage{bm}% bold math
\usepackage[utf8]{inputenc} % allow utf-8 input
\usepackage[T1]{fontenc}    % use 8-bit T1 fonts
\usepackage{hyperref}       % hyperlinks
\usepackage{url}            % simple URL typesetting
\usepackage{booktabs}       % professional-quality tables
\usepackage{amsfonts}       % blackboard math symbols
\usepackage{nicefrac}       % compact symbols for 1/2, etc.
\usepackage{microtype}      % microtypography

\usepackage{microtype}
\usepackage{graphicx}
\usepackage{booktabs} % for professional tables
\usepackage{float}

\usepackage{color}
\usepackage{physics}
\usepackage{cancel}

\usepackage{hyperref}
\hypersetup{
    colorlinks=true,
    linkcolor=red,
    filecolor=magenta,      
    urlcolor=cyan,
    citecolor=blue,
}

\usepackage{amsfonts}
\usepackage{amsmath}

\usepackage{xr}

\begin{document}

\preprint{APS/123-QED}

\title{A self consistent theory of Gaussian Processes \\ captures feature learning effects in finite CNNs}% Force line breaks with \\

\author{Gadi Naveh$^{1,2}$ and Zohar Ringel$^{1}$}
\affiliation{
$^{1}$Racah Institute of Physics, Hebrew University, Jerusalem 91904, Israel~\\
$^{2}$Edmond and Lily Safra Center for Brain Sciences, Hebrew University, Jerusalem 91904, Israel
}

\date{\today}% It is always \today, today,
             %  but any date may be explicitly specified

\begin{abstract}
Deep neural networks (DNNs) in the infinite width/channel limit have received much attention recently, as they provide a clear analytical window to deep learning via mappings to Gaussian Processes (GPs). 
Despite its theoretical appeal, this viewpoint lacks a crucial ingredient of deep learning in finite DNNs, laying at the heart of their success --- \textit{feature learning}. 
Here we consider DNNs trained with noisy gradient descent on a large training set and derive a self consistent Gaussian Process theory accounting for strong finite-DNN and feature learning effects. 
Applying this to a toy model of a two-layer linear convolutional neural network (CNN) shows good agreement with experiments.
We further identify, both analytical and numerically, a sharp transition between a feature learning regime and a lazy learning regime in this model. 
Strong finite-DNN effects are also derived for a non-linear two-layer fully connected network. 
Our self consistent theory provides a rich and versatile analytical framework for studying feature learning and other non-lazy effects in finite DNNs.
\end{abstract}

%%%%%%%%%%%%%%%%%%%%%%%%%%%%%%%%%%%%%%%%%%%%%%%%%%%%%%%%%%%%%%%%%%%%%%%%%%%%%%%%%%%%%%%%%%%%%%%%%%%%%%%%%%%%%%%%%%%%%%%%%%%%%%%%%%

\maketitle
\onecolumngrid

\section{Introduction}
\label{section: intro}
The correspondence between Gaussian Processes (GPs) and deep neural networks (DNNs) has been instrumental in advancing our understanding of these complex algorithms. 
Early results related randomly initialized strongly over-parameterized DNNs with GP priors  \cite{neal1996priors,lee2017deep, matthews2018gaussian}. 
More recent results considered training using gradient flow (or noisy gradients), where DNNs map to Bayesian inference on GPs governed by the neural tangent kernel  \cite{Jacot2018, lee2019wide} (or the NNGP kernel \cite{naveh2020predicting}). These correspondences carry over to a wide variety of architectures, going beyond fully connected networks (FCNs) to convolutional neural networks (CNNs) \cite{Arora2019, novak2018bayesian}, recurrent neural networks (RNNs) \cite{alemohammad2020recurrent} and even attention networks \cite{hron2020infinite}. 
They provide us with closed analytical expressions for the outputs of strongly over-parameterized trained DNNs, which have been used to make accurate predictions for DNN learning curves \cite{Cohen2019,Canatar2021,bordelon2020spectrum}.

Despite their theoretical appeal, GPs are unable to capture \textit{feature learning} \cite{yu2013feature, Yosinski2014}, which is a well-observed key property of trained DNNs. 
Indeed, it was noticed \cite{Jacot2018} that as the width tends to infinity, the neural tangent kernel (NTK) tends to a constant kernel that does not evolve during training and the weights in hidden layers change  infinitesimally from their initialization values. This regime of training was thus dubbed \textit{lazy training} \cite{chizat2019lazy}. 
Other studies showed that for CNNs trained on image classification tasks, the feature learning regime generally tends to outperform the lazy regime
\cite{geiger2020disentangling, geiger2021landscape, lee2020finite}. 
Clearly, working in the feature learning regime is also crucial for performing \textit{transfer learning} \cite{yang2020feature, Yosinski2014}.

It is therefore desirable to have a theoretical approach to deep learning which enjoys the generality and analytical power of GPs while capturing feature learning effects in finite DNNs. 
Here we make several contributions towards this goal:
\begin{enumerate}
    \item 
    We show that the mean predictor of a finite DNN trained on a large data set with noisy gradients, weight decay and MSE loss, can be obtained from GP regression on a shifted target (§\ref{section: target shift}). 
    Central to our approach is a non-linear self consistent equation involving the higher cumulants of the finite DNN (at initialization) which predicts this target shift.  

    \item 
    Using this machinery on a toy model of a two-layer linear CNN in a teacher-student setting, we derive explicit analytical predictions which are in very good agreement with experiments even well away from the GP/lazy-learning regime (large number of channels, $C$) thus accounting for \textit{strong} finite-DNN corrections (§\ref{section: toy model}).
    Similarly strong corrections to GPs, yielding qualitative improvements in performance, are demonstrated for the quadratic two-layer fully connected model of Ref. \cite{mannelli2020optimization}.
 
    \item 
    We show how our framework can be used to study statistical properties of weights in hidden layers.
    In particular, in the CNN toy model, we identify, both analytically and numerically, a sharp transition between a feature learning phase and a lazy learning phase (§\ref{subsection: Feature learning phase transition}). 
    We define the feature learning phase as the regime where the features of the teacher network leave a clear signature in the spectrum of the student's hidden weights posterior covariance matrix. 
    In essence, this phase transition is analogous to the transition associated with the recovery of a low-rank signal matrix from a noisy matrix taken from the Wishart ensemble, when varying the strength of the low-rank component \cite{benaych2011eigenvalues}. 
    
\end{enumerate}

\subsection{Additional related work}
\label{subsection: previous work}
Several previous papers derived leading order finite-DNN corrections to the GP results \cite{naveh2020predicting, yaida2020non, Dyer2020Asymptotics}. 
While these results are in principle extendable to any order in perturbation theory, such high order expansions have not been studied much, perhaps due to their complexity. 
In contrast, we develop an analytically tractable non-perturbative approach which we find crucial for obtaining non-negligible feature learning and associated performance enhancement effects. 

Previous works \cite{geiger2020disentangling, geiger2021landscape} studied how the behavior of infinite DNNs depends on the scaling of the top layer weights with its width. 
In \cite{yang2020feature} it is shown that the standard and NTK parameterizations of a neural network do not admit an infinite-width limit that can learn features, and instead suggest an alternative parameterization which can learn features in this limit.
While unifying various viewpoints on infinite DNNs, this approach does not immediately lend itself to analytical analysis of the kind proposed here. 

Several works \cite{refinetti2021classifying, malach2021quantifying, ghorbani2020neural, andreassen2020asymptotics} show that finite width models can generalize either better or worse than their infinite width counterparts, and provide examples where the relative performance depends on the optimization details, the DNN architecture and the statistics of the data. 
Here we demonstrate analytically that finite DNNs outperform their GP counterparts when the latter have a prior that lacks some constraint found in the data (e.g. positive-definiteness \cite{mannelli2020optimization}, weight sharing \cite{novak2018bayesian}). 

Deep linear networks (FCNs and CNNs) similar to our CNN toy example have been studied in the literature \cite{baldi1989neural, saxe2013exact, lampinen2018analytic, gunasekar2018implicit}. 
These studies use different approaches and assumptions and do not discuss the target shift mechanism which applies also for non-linear CNNs. 
In addition, their analytical results hinge strongly on linearity whereas our approach could be useful whenever several leading cumulants of the DNN output are known. 

A concurrent work \cite{zavatone2021exact} derived exact expressions for the output priors of finite FCNs induced by Gaussian priors over their weights.
However, these results only apply to the limited case of a \textit{prior} over a \textit{single} training point and only for a FCN.
In contrast, our approach applies to the setting of a large training set, it is not restricted to FCNs and yields results for the posterior predictions, not the prior. Focusing on deep linear fully connected DNNs, recent work \cite{li2020statistical} derived analytical finite-width renormalization results for the GP kernel, by sequentially integrating out the weights of the DNN, starting from the output layer and working backwards towards the input. Our analytical approach, its scope, and the models studied here differ substantially from that work.

%%%%%%%%%%%%%%%%%%%%%%%%%%%%%%%%%%%%%%%%%%%%%%%%%%%%%%%%%%%%%%%%%%%%%%%

\section{Preliminaries}
\label{section: Preliminaries} 
We consider a fixed set of $n$ training inputs $\left\{ \x_{\mu}\right\} _{\mu=1}^{n} \subset \bbR^d$ and a single test point $\x_*$ over which we wish to model the distribution of the outputs of a DNN. 
We consider a generic DNN architecture where for simplicity we assume a scalar output $f(\x) \in \bbR$. 
The learnable parameters of the DNN that determine its output, are collected into a single vector $\theta$. 
We pack the outputs evaluated on the training set and on the test point into a vector 
$\vec{f}\equiv\left(f\left(\x_{1}\right),\dots,f\left(\x_{n}\right),f\left(\x_{n+1}\right)\right)\in\bbR^{n+1}$, where we denoted the test point as $\x_{*}=\x_{n+1}$. 
We train the DNNs using \textit{full-batch} gradient decent with weight decay and external white Gaussian noise. The discrete dynamics of the parameters are thus
\begin{equation}
\label{eq:discrete Langevin wd}
    \theta_{t+1} - \theta_{t}
     = - \left( \gamma \theta_{t} + \nabla_{\theta} \cL \left(f_{\theta}\right) \right) \eta + 2 \sigma \sqrt{\eta}\xi_{t}
\end{equation}
where $\theta_t$ is the vector of all network parameters at time step $t$, $\gamma$ is the strength of the weight decay, $\cL(f_\theta)$ is the loss as a function of the DNN output $f_\theta$ (where we have emphasized the dependence on the parameters $\theta$), $\sigma$ is the magnitude of noise, $\eta$ is the learning rate and $\xi_t \sim \cN (0, I)$. 
As $\eta \to 0$ this discrete-time dynamics converge to the continuous-time Langevin equation given by
$
\dot{\theta}\left(t\right) = -\nabla_{\theta} \left(\frac{\gamma}{2} ||\theta(t)||^2 + \cL \left(f_{\theta}\right)\right) + 2 \sigma \xi\left(t\right)
$ 
with 
$
\left\langle \xi_i(t) \xi_j(t') \right\rangle = \delta_{ij} \delta\left(t-t'\right)
$, so that as $t \to \infty$ the DNN parameters $\theta$ will be sampled from the equilibrium Gibbs distribution $P(\theta)$. 

As shown in \cite{naveh2020predicting}, the parameter distribution $P(\theta)$ induces a posterior distribution over the trained DNN outputs $P(\vec{f})$ with the following partition function: 
\begin{equation}
\label{eq: Z posterior orig}
    Z\left(\vec{J}\right) = 
    \int d\vec{f}P_{0}\left(\vec{f}\right)\exp\left(-\frac{1}{2\sigma^{2}}\cL\left(\left\{ f_{\mu}\right\} _{\mu=1}^{n}, \left\{ g_{\mu}\right\} _{\mu=1}^{n}\right)+\sum_{\mu=1}^{n+1}J_{\mu}f_{\mu}\right)
\end{equation}
Here $P_0(\vec{f})$ is the prior generated by the finite-DNN with $\theta$ drawn from ${\mathcal N}(0, 2 \sigma^2/\gamma)$ where the weight decay $\gamma$ may be layer-dependent, $\left\{ g_{\mu}\right\} _{\mu=1}^{n}$ are the training targets and $\vec{J}$ are source terms used to calculate the statistics of $f$.
We keep the loss function $\cL$ arbitrary at this point, committing to a specific choice in the next section.
As standard \cite{helias2019statistical}, to calculate the posterior mean at any of the training points or the test point $\x_{n+1}$ from this partition function one uses
\begin{equation}
\label{eq: pa_J log Z}    
    \forall\mu\in\left\{ 1,\dots,n+1\right\}: \qquad
    \left\langle f_{\mu}\right\rangle = \partial_{J_{\mu}}\eval{\log Z\left(\vec{J}\right)}_{\vec{J}=\vec{0}}
\end{equation}

%%%%%%%%%%%%%%%%%%%%%%%%%%%%%%%%%%%%%%%%%%%%%%%%%%%%%%%%%%%%%%%%%%%%%%%

\section{A self consistent theory for the posterior mean and covariance}
\label{section: target shift}
In this section we show that for a large training set, the posterior mean predictor (Eq. \ref{eq: pa_J log Z}) amounts to GP regression on a shifted target ($g_{\mu} \to g_{\mu} - \Delta g_{\mu}$). 
This shift to the target ($\Delta g_{\mu}$) is determined by solving certain self-consistent equations involving the cumulants of the prior $P_0(\vec{f})$.
For concreteness, we focus here on the MSE loss 
$\cL = \sum_{\mu=1}^{n}\left(f_{\mu}-g_{\mu}\right)^{2}$ and comment on extensions to other losses, e.g. the cross entropy, in App. \ref{appendix: TS derivs}. 
To this end, consider first the prior of the output of a finite DNN. Using standard manipulations (see App. \ref{appendix: TS}), it can be expressed as follows 
\begin{equation}
\label{eq: prior via cumulants}    
    P_{0}\left(\vec{f}\right)\propto\int_{\bbR^{n+1}} d\vec{t}\exp\left(-\sum_{\mu=1}^{n+1}it_{\mu}f_{\mu}+\sum_{r=2}^{\infty}\frac{1}{r!}\sum_{\mu_{1},\dots,\mu_{r}=1}^{n+1}\kappa_{\mu_{1},\dots,\mu_{r}}it_{\mu_{1}}\cdots it_{\mu_{r}}\right)
\end{equation}
where $\kappa_{\mu_1,\dots,\mu_r}$ is the $r$'th multivariate cumulant of $P_{0}(\vec{f})$ \cite{Mccullagh2017}. The second term in the exponent is the cumulant generating function ($\cC$) corresponding to $P_0$. As discussed in App. \ref{appendix: Edgeworth review} and Ref. \cite{naveh2020predicting}, for standard initialization protocols the $r$'th cumulant will scale as $1/C^{(r/2-1)}$, where $C$ controls the over-parameterization. 
The second ($r=2$) cumulant which is $C$-independent, describes the NNGP kernel of the finite DNN and is denoted by $K(\x_{\mu_1}, \x_{\mu_2}) = \kappa_{\mu_1,\mu_2}$. 

Consider first the case of $C\rightarrow \infty$ \cite{lee2017deep, matthews2018gaussian, neal1996priors} where all $r>2$ cumulants vanish. 
Here one can explicitly perform the integration in Eq. \ref{eq: prior via cumulants} to obtain the standard GP prior 
$P_{0}\left(\vec{f}\right)\propto\exp\left(-\frac{1}{2}\sum^{n+1}_{\mu_{1},\mu_{2}=1}\kappa_{\mu_{1},\mu_{2}}f_{\mu_{1}}f_{\mu_{2}}\right)$. 
Plugging this prior into Eq. \ref{eq: Z posterior orig} with MSE loss, one recovers standard GP regression formulas \cite{Rasmussen2005}. 
In particular, the predictive mean at $\x_*$ is: 
$\left\langle f\left(\x_{*}\right)\right\rangle =\sum_{\mu,\nu=1}^{n}K_{\mu}^{*}\tilde{K}_{\mu\nu}^{-1}g_{\nu}$
where $K_{\mu}^{*}=K\left(\x_{*},\x_{\mu}\right)$ and $\tilde{K}_{\mu\nu}=K\left(\x_{\mu},\x_{\nu}\right)+\sigma^{2}\delta_{\mu\nu}$. 
Another set of quantities we shall find useful are the \textit{discrepancies in GP prediction}, which for the training set read
\begin{align}
\label{eq: discrepancies}
    \boxed{\forall\mu\in\left\{ 1,\dots,n\right\} :\quad \langle\hat{\delta}g_{\mu}\rangle\equiv g_{\mu}-\left\langle f\left(\x_{\mu}\right)\right\rangle = 
    g_{\mu}-\sum_{\nu,\nu'=1}^{n}K_{\mu\nu'}\tilde{K}_{\nu',\nu}^{-1}g_{\nu}
    }
\end{align}

\paragraph{Saddle point approximation for the mean predictor.}
For a DNN with finite $C$, the prior $P_0(\vec{f})$ will no longer be Gaussian and cumulants with $r>2$ would contribute. 
This renders the partition function in Eq. \ref{eq: Z posterior orig} intractable and so some approximation is needed to make progress.
To this end we note that $f$ can be integrated out (see App. \ref{appendix: TS dual Z}) to yield a partition function of the form
\begin{equation}
\label{eq: Z=exp(-S)}
    Z\left(\vec{J}\right)\propto \int_{\bbR^n} dt_{1}\cdots dt_{n}e^{-\cS\left(\vec{t},\vec{J}\right)}    
\end{equation}
where $\cS(\vec{t},\vec{J})$ is the action whose exact form is given in Eq. \ref{appEq: S of t and J GP}. 
Interestingly, the $it_{\mu}$ variables appearing above are closely related to the discrepancies $\hat{\delta}g_{\mu}$, in particular $\langle it_{\mu} \rangle = \langle \hat{\delta}g_{\mu} \rangle / \sigma^2$. 

To proceed analytically we adopt the \textit{saddle point (SP) approximation} \cite{Daniels1954} which, as argued in App. \ref{appendix: heuristic SP criterion}, relies on the fact that the non-linear terms in the action comprise of a sum of many $it_{\mu}$'s. 
Given that this sum is dominated by collective effects coming from all data points, expanding $\cS(\vec{t},\vec{J})$ around the saddle point yields terms with increasingly negative powers of $n$. 

For the training points $\mu\in\left\{ 1,\dots,n\right\}$, taking the saddle point approximation amounts to setting 
$\eval{\pa_{it_{\mu}}\cS\left(\vec{t},\vec{J}\right)}_{\vec{J}=\vec{0}} = 0$.    
This yields a set of equations that has precisely the form of Eq. \ref{eq: discrepancies}, but where the target is shifted as $g_{\nu}\to g_{\nu}-\Delta g_{\nu}$ and the target shift is determined self consistently by
\begin{equation}
\label{eq: target shift SP}
    \boxed{\Delta g_{\nu}=\sum_{r=3}^{\infty}\frac{1}{\left(r-1\right)!}\sum_{\mu_{1},\dots,\mu_{r-1}=1}^{n}\kappa_{\nu,\mu_{1},\dots,\mu_{r-1}}\left\langle \sigma^{-2}\hat{\delta}g_{\mu_{1}}\right\rangle \cdots\left\langle \sigma^{-2}\hat{\delta}g_{\mu_{r-1}}\right\rangle}
\end{equation}
Equation \ref{eq: target shift SP} is thus an implicit equation for $\Delta g_{\nu}$ involving all training points, and it holds for the training set and the test point $\nu\in\left\{ 1,\dots, n+1 \right\}$. 
Once solved, either analytically or numerically, one calculates the predictions on the test point via 
\begin{equation}
\label{eq: SC test SP}
    \boxed{\left\langle f_{*}\right\rangle =\Delta g_{*}+\sum_{\mu,\nu=1}^{n}K_{\mu}^{*}\tilde{K}_{\mu\nu}^{-1}\left(g_{\nu}-\Delta g_{\nu}\right)}
\end{equation}

Equation \ref{eq: discrepancies} with $g_{\nu}\to g_{\nu}-\Delta g_{\nu}$ along with Eqs. \ref{eq: target shift SP} and \ref{eq: SC test SP} are the first main result of this paper. 
Viewed as an algorithm, the procedure to predict the finite DNN's output on a test point $\x_*$ is as follows: 
we shift the target in Eq. \ref{eq: discrepancies} as $g \to g - \Delta g$ with $\Delta g$ as in Eq. \ref{eq: target shift SP}, arriving at a closed equation for the average discrepancies $\langle \hat{\delta}g_\mu \rangle$ on the training set.
For some models, the cumulants $\kappa_{\nu, \mu_2, \dots, \mu_r}$ can be computed for any order $r$ and it can be possible to sum the entire series, while for other models several leading cumulants might already give a reasonable approximation due to their $1/C^{r/2 - 1}$ scaling. 
The resulting coupled non-linear equations can then be solved numerically, to obtain $\Delta g_{\mu}$ from which predictions on the test point are calculated using Eq. \ref{eq: SC test SP}.

Notwithstanding, solving such equations analytically is challenging and one of our main goals here is to provide concrete analytical insights. 
Thus, in §\ref{subsubsection: EK SC eqs} we propose an additional approximation wherein to leading order we replace all summations over data-points with integrals over the measure from which the data-set is drawn. 
This approximation, taken in some cases beyond leading order as in Ref. \cite{Cohen2019}, will yield analytically tractable equations which we solve for two simple toy models, one of a linear CNN and the other of a non-linear FCN.

\paragraph{Saddle point plus Gaussian fluctuations for the posterior covariance.} 
The SP approximation can be extended to compute the predictor variance by expanding the action $\cS$ to quadratic order in $it_\mu$ around the SP value (see App. \ref{appendix: cov SP + fluct}).
Due to the saddle-point being an extremum this leads to $\cS \approx \cS_{\rm{SP}} + \frac{1}{2}t_{\mu}A^{-1}_{\mu \nu} t_{\nu}$. 
This leaves the previous SP approximation for the posterior mean on the training set unaffected (since the mean and maximizer of a Gaussian coincide), but is necessary to get sensible results for the posterior covariance.
Empirically, in the toy models we considered in §\ref{section: toy model} we find that the finite DNN corrections to the variance are much less pronounced than those for the mean. 
Using the standard Gaussian integration formula, one finds that $A_{\mu \nu}$ is the covariance matrix of $it_{\mu}$. 
Performing such an expansion one finds 
\begin{align}
\label{eq: Ainv Delta K train}
    A^{-1}_{\mu \nu} &= -\left(\sigma^{2}\delta_{\mu\nu} + K_{\mu\nu} + \Delta K_{\mu\nu}\right)  \\ \nonumber
    \Delta K_{\mu \nu} &= \partial_{it_{\mu}}\Delta g_{\nu}\left(it_{1},\dots,it_{n}\right)
\end{align}
where the $it_{\mu}$ on the r.h.s. are those of the saddle point. This gives an expression for the posterior covariance matrix on the training set:
\begin{align}
\label{eq: SP posterior cov train}    
    \Sigma_{\mu\nu}=\left\langle f_{\mu}f_{\nu}\right\rangle -\left\langle f_{\mu}\right\rangle \left\langle f_{\nu}\right\rangle =-\sigma^{4}\left[\sigma^{2}I + K +\Delta K\right]_{\mu\nu}^{-1}+\sigma^{2}\delta_{\mu\nu}
\end{align}
where the r.h.s. coincides with the posterior covariance of a GP with a kernel equal to $K+\Delta K$ \cite{Rasmussen2005}. 
The variance on the test point is given by (repeating indices are summed over the training set)
\begin{align}
\label{eq: posterior var test}    
    \Sigma_{**}=K_{**}-K_{\mu}^{*}A_{\mu\nu}^{-1}K_{\nu}^{*}+\left\langle \partial_{it_{*}}^{2}\tilde{\cC}|_{it_{*}=0}\right\rangle +2\left(\left\langle \Delta g_{*}K_{\mu}^{*}it_{\mu}\right\rangle -\left\langle \Delta g_{*}\right\rangle \left\langle K_{\mu}^{*}it_{\mu}\right\rangle \right)+\Var\left(\Delta g_{*}\right)
\end{align}
where here $\Delta g_*$ is as in Eq. \ref{eq: target shift SP} but where the $\left\langle \sigma^{-2}\hat{\delta}g_{\mu}\right\rangle$'s are replaced the $it_\mu$'s that have Gaussian fluctuations, and $\tilde{\cC}$ is $\cC$ without the second cumulant (see App. \ref{appendix: TS dual Z}). 
The first two terms in Eq. \ref{eq: posterior var test} yield the standard result for the GP posterior covariance matrix on a test point \cite{Rasmussen2005}, for the case of $\Delta K = 0$ (see Eq. \ref{eq: Ainv Delta K train}).
The rest of the terms can be evaluated by the SP plus Gaussian fluctuations approximation, where the details would depend on the model at hand.

%%%%%%%%%%%%%%%%%%%%%%%%%%%%%%%%%%%%%%%%%%%%%%%%%%%%%%%%%%%%%%%%%%%%%%%%%%%%%%%%%%%%%%%%%%

\section{Two toy models}
\label{section: toy model}
\subsection{The two layer linear CNN and its properties}
\label{section: two layer linear CNN}

Here we define a teacher-student toy model showing several qualitative real-world aspects of feature learning and analyze it via our self-consistent shifted target approach. 
Concretely, we consider the simplest student CNN $f(\x)$, having one hidden layer with linear activation, and a corresponding teacher CNN, $g(\x)$
\begin{equation}
\label{eq: student teacher CNN linear}
    f\left(\x \right) = \sum_{i=1}^{N}\sum_{c=1}^{C}a_{i,c} \w_{c}\cdot\tilde{\x}_{i}
    \qquad
    g\left(\x \right) = \sum_{i=1}^{N}\sum_{c=1}^{C^*}a^*_{i,c} \w^*_{c}\cdot\tilde{\x}_{i}
\end{equation}
This describes a CNN that performs 1-dimensional convolution where the convolutional weights for each channel are $\w_{c} \in \bbR ^S$. 
These are dotted with a convolutional window of the input 
$
\tilde{\x}_{i} = \left(x_{S\left(i-1\right)+1},\dots,x_{S\cdot i}\right)^{\transpose} \in\bbR^S
$
and there are no overlaps between them so that 
$
\x = \left(x_{1}, \dots, x_{N\cdot S}\right)^{\transpose} = \left(\tilde{\x}_{1}, \dots, \tilde{\x}_{N}\right)^{\transpose} \in \bbR^{N \cdot S}
$. 
Namely, the input dimension is $d = NS$, where $N$ is the number of (non-overlapping) convolutional windows, $S$ is the stride of the conv-kernel and it is also the length of the conv-kernel, hence there is no overlap between the strides. 
The inputs $\x$ are sampled from $\cN(0, I_{d})$.

Despite its simplicity, this model distils several key differences between feature learning models and lazy learning or GP models. 
Due to the lack of pooling layers, the GP associated with the student fails to take advantage of the weight sharing property of the underlying CNN \cite{novak2018bayesian}. 
In fact, here it coincides with a GP of a fully-connected DNN which is quite inappropriate for the task. We thus expect that the finite network will have good performance already for $n = C^*(N + S)$ whereas the GP will need $n$ of order of the dimension ($NS$) to learn well \cite{Cohen2019}. Thus, for $N+S \ll NS$ there should be a broad regime in the value of $n$ where the finite network substantially outperforms the corresponding GP. 
We later show (§\ref{subsection: Feature learning phase transition}) that this performance boost over GP is due to feature learning, as one may expect.

Conveniently, the cumulants of the student DNN of any order can be worked out exactly. 
Assuming $\gamma$ and $\sigma^2$ of the noisy GD training are chosen such that\footnote{Generically this requires $C$ dependent and layer dependent weight decay.} 
$
a_{i,c}\sim\cN\left(0,\sigma_{a}^{2}/CN\right),\quad \w_{c} \sim \cN \left(\boldsymbol{0},  \frac{\sigma_{w}^{2}}{S}I_S \right)
$ 
(and similarly for the teacher DNN) the covariance function for the associated GP reads 
\begin{equation}
    K\left(\x,\x'\right)=\frac{\sigma_{a}^{2}\sigma_{w}^{2}}{NS}\sum_{i=1}^{N}\tilde{\x}_{i}^{\transpose}\tilde{\x}_{i}'=\frac{\sigma_{a}^{2}\sigma_{w}^{2}}{NS}\x_{i}^{\transpose}\x_{i}'
\end{equation}
Denoting $\lambda:=\frac{\sigma_{a}^{2}}{N}\frac{\sigma_{w}^{2}}{S}$, the even cumulant of arbitrary order $2m$ is (see App. \ref{appendix: cumulants linear CNN}):
\begin{equation}
\label{eq: gen cumulant linear CNN}
    \kappa_{2m}\left(\x_{1},\dots,\x_{2m}\right)=\frac{\lambda^{m}}{C^{m-1}}\sum_{i_{1},\dots,i_{m}=1}^{N}\left(\bullet_{i_{1}},\bullet_{i_{2}}\right)\cdots\left(\bullet{}_{i_{m-2}},\bullet{}_{i_{m-1}}\right)\left(\bullet{}_{i_{m-1}},\bullet{}_{i_{m}}\right)\cdots\left[\left(2m-1\right)!\right]
\end{equation}
while all odd cumulants vanish due to the sign flip symmetry of the last layer. 
In this notation, we mean that the $\bullet$'s stand for integers in $\left\{ 1,\dots,2m\right\} $ and e.g. 
$\left(1_{i_{1}},2_{i_{2}}\right)\equiv\left(\tilde{\x}_{i_{1}}^{1}\cdot\tilde{\x}_{i_{2}}^{2}\right)$
and the bracket notation
$\left[\left(2m-1\right)!\right]$  stands for the number of ways to pair the integers $\left\{ 1,...,2m\right\}$ into the above form.  This result can then be plugged in \ref{eq: target shift SP} to obtain the self consistent (saddle point) equations on the training set. See App. \ref{appendix: heuristic SP criterion} for a convergence criterion for the saddle point, supporting its application here. 

\subsection{Self consistent equation in the limit of a large training set}
\label{subsubsection: EK SC eqs}
In §\ref{section: target shift} our description of the self consistent equations was for a finite and fixed training set. 
Further analytical insight can be gained if we consider the limit of a large training set, known in the GP literature as the Equivalent Kernel (EK) limit \cite{Rasmussen2005, sollich2004understanding}. For a short review of this topic, see App. \ref{appendix: EK review}.
In essence, in the EK limit we replace the discrete sums over a specific draw of training set, as in Eqs. \ref{eq: discrepancies}, \ref{eq: target shift SP}, \ref{eq: SC test SP},
with integrals over the entire input distribution $\mu(\x)$. 
Given a kernel that admits a spectral decomposition in terms of its eigenvalues and eigenfunctions: 
$K\left(\x,\x'\right)=\sum_{s}\lambda_{s}\psi_{s}\left(\x\right)\psi_{s}\left(\x'\right)$, the standard result for the GP posterior mean at a test point is approximated by \cite{Rasmussen2005}
\begin{equation}
\label{eq: EK GP}    
   \left\langle f\left(\x_{*}\right)\right\rangle =\int d\mu\left(\mathbf{x}\right)h\left(\mathbf{x}_{*},\x\right)g\left(\x\right);
   \qquad
   h\left(\mathbf{x}_{*},\x\right) = \sum_{s}\frac{\lambda_{s}}{\lambda_{s}+\sigma^{2}/n}\psi_{s}\left(\x_{*}\right)\psi_{s}\left(\x\right)
\end{equation}
This has several advantages, already at the level of GP analysis.
From a theoretical point of view, the integral expressions retain the symmetries of the kernel $K(\x, \x')$ unlike the discrete sums that ruin these symmetries. 
Also, Eq. \ref{eq: EK GP} does not involve computing the inverse matrix $\tilde{K}^{-1}$ which is costly for large matrices. 

In the context of our theory, the EK limit allows for a derivation of a simple analytical form for the self consistent equations. As shown in App. \ref{appendix: technical SC eqs EK limit} in our toy CNN both $\Delta g$ and $\hat{\delta} g$ become linear in the target. 
Thus the self-consistent equations can be reduced to a single equation governing the proportionality factor ($\alpha$) between $\hat{\delta}g$ and $g$ 
($\hat{\delta}g = \alpha g$). 
Thus starting from the general self consistent equations, \ref{eq: discrepancies}, \ref{eq: target shift SP}, \ref{eq: SC test SP}, taking their EK limit, and plugging in the general cumulant for our toy model (\ref{eq: gen cumulant linear CNN}) we arrive at the following equation for $\alpha$
\begin{equation}
\label{eq: SC alpha all kappas no SL}
    \alpha=\frac{\sigma^{2}/n}{\lambda+\sigma^{2}/n}+\frac{\left(1-q\right)\lambda}{\lambda+\sigma^{2}/n}+\left(q\frac{\lambda}{\lambda+\sigma^{2}/n}-1\right)\frac{\lambda^{2}}{C}\left(\frac{\alpha}{\sigma^{2}/n}\right)^{3}\left[1-\frac{\lambda}{C}\left(\frac{\alpha}{\sigma^{2}/n}\right)^{2}\right]^{-1}
\end{equation}
%We comment that to leading order the large $C$ limit, we may keep only the leading term involving the fourth cumulant $\kappa_4$ and then the only change in the equation would be to ignore the $\left[\cdots\right]^{-1}$. 
%This would give a leading order perturbative result for finite width corrections, as in \cite{naveh2020predicting, yaida2020non}.
Setting for simplicity $\sigma_{a}^{2}=1=\sigma_{w}^{2}$ 
we have $\lambda=1/\left(NS\right)$ and we also introduced the constant
$ q \equiv \lambda^{-1}(1 - \hat{\alpha}_{\rm{GP}}) (\lambda + \sigma^2/n) 
$ 
where $\hat{\alpha}_{\mathrm{GP}}$ is computed using the empirical GP predictions on either the training set or test set:
$\hat{\alpha}_{\rm{GP}} \equiv 1 - \left(\sum_{\mu}f_{\mu}^{\rm{GP}}g_{\mu}\right)/\left(\sum_{\mu}g_{\mu}^{2}\right)$, 
or analytically in the perturbation theory approach developed in \cite{Cohen2019}.
The quantity $q$ has an interpretation as $1/n$ corrections to the EK approximation \cite{Cohen2019} but here can be considered as a fitting parameter. It is non-negative and is typically $O(1)$; for more details and analytical estimates see App. \ref{appendix: Fudge factor}. 

Equation \ref{eq: SC alpha all kappas no SL} is the second main analytical result of this work. 
It simplifies the highly non-linear inference problem to a single equation that embodies strong non-linear finite-DNN effect and feature learning (see also §\ref{subsection: Feature learning phase transition}). 
In practice, to compute $\alpha_\mathrm{test}$ we numerically solve \ref{eq: SC alpha all kappas no SL} using $q_\mathrm{train}$ for the training set to get $\alpha_\mathrm{train}$, and then set $\alpha = \alpha_\mathrm{train}$ in the r.h.s. of \ref{eq: SC alpha all kappas no SL} but use $q = q_\mathrm{test}$. 
Equation \ref{eq: SC alpha all kappas no SL} can also be used to bound $\alpha$ analytically on both the training set and test point, given the reasonable assumption that $\alpha$ changes continuously with $C$. 
Indeed, at large $C$ the pole in this equation lays at $\alpha_{\rm{pole}} = (\sigma^2/n) (C/\lambda)^{1/2} \gg 1$ whereas $\alpha \approx \alpha_{\rm{GP}} < \alpha_{\rm{pole}}$. 
As $C$ diminishes, continuity implies that $\alpha$ must remain smaller than $\alpha_{\rm{pole}}$. The latter decays as $\sigma^2\sqrt{CNS}/n$ implying that the amount of data required for good performance scales as $\sqrt{CNS}$ rather than as $NS$ in the GP case.

%%%%%%%%%%%%%%%%%%%%%%%%%%%%%%%%%%%%%%%%%%%%%%%%%%%%%%%%%%%%%%%%%%%%%%%%%%%%%%%%%%%%%%%%%%

\subsection{Numerical verification}
\label{section: Numerical}
In this section we numerically verify the predictions of the self consistent theory of Sec. §\ref{subsubsection: EK SC eqs}, by training linear shallow student CNNs on a teacher with $C^* = 1$ as in Eq. \ref{eq: student teacher CNN linear}, using noisy gradients as in Eq. \ref{eq:discrete Langevin wd}, and averaging their outputs across noise realizations and across dynamics after reaching equilibrium. 

For simplicity we used $N=S$ and $n\in\left\{ 62,200,650\right\}, S\in\left\{ 15,30,60\right\}$ so that $n \propto S^{1.7}$. 
The latter scaling places us in the poorly performing regime of the associated GP while allowing good performance of the CNN. 
Indeed, as aforementioned, the GP here requires $n$ of the scale of $\lambda^{-1} = NS = O(S^2)$ for good performance \cite{Cohen2019}, while the CNN requires $n$ of scale of the number of parameters ($C(N+S)=O(S)$).      

\begin{figure}[h]
\vspace*{-0.1in}
\begin{center}
(A)
\includegraphics[height=4cm]{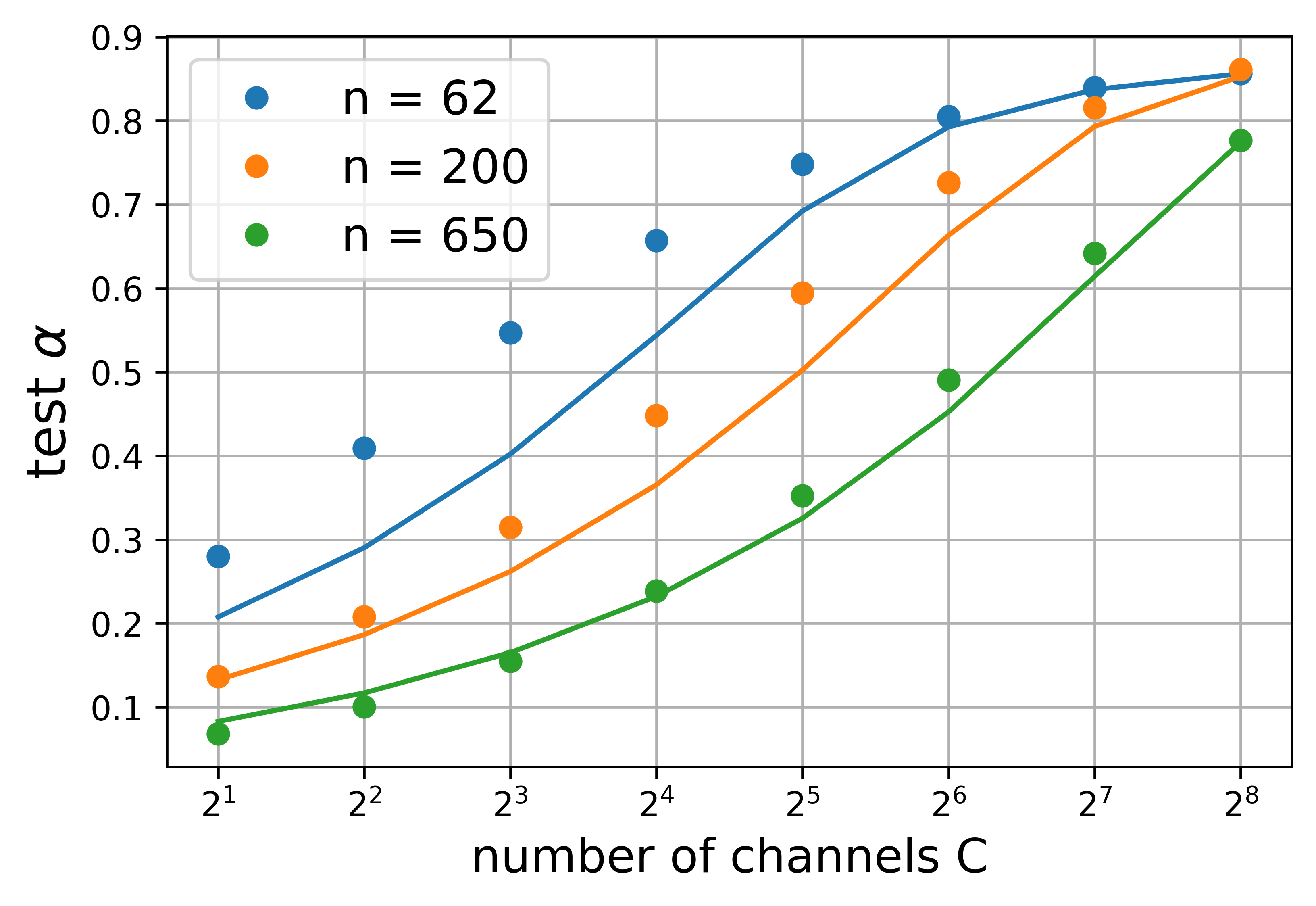}
(B)
\includegraphics[height=4cm]{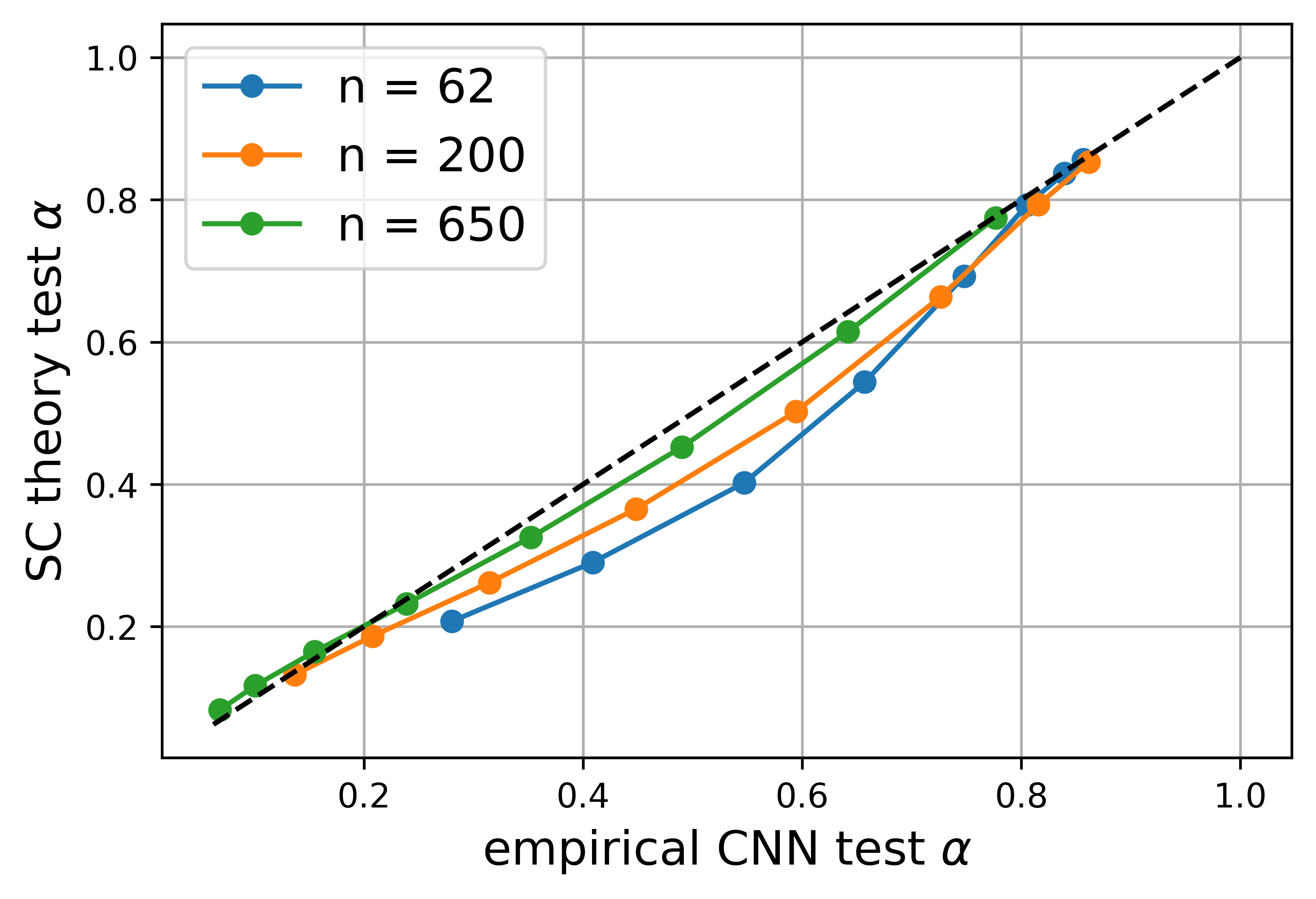}

\end{center}
% \vspace*{-0.15in}
\caption{ 
\textbf{(A)} 
The CNNs' cosine distance $\alpha$, defined by $\left\langle f\right\rangle  = (1 - \alpha) g$
between the ensemble-averaged prediction $\left\langle f\right\rangle $ and ground truth $g$ plotted vs. number of channels $C$ for the test set (for the train set, see App. \ref{appendix: details numerics}).  
As $n$ increases, the solution of the self consistent equation \ref{eq: SC alpha all kappas no SL} (solid line) yields an increasingly accurate prediction of these empirical values (dots).
\textbf{(B)} 
Same data as in (A), presented as empirical $\alpha$ vs. predicted $\alpha$.
As $n$ grows, the two converge to the identity line (dashed black line). Solid lines connecting the dots here are merely for visualization purposes.  
\label{fig: SC theory verification}}
\end{figure}

The results are shown in Fig \ref{fig: SC theory verification} where we compare the theoretical predictions given by the solutions of the self consistent equation (\ref{eq: SC alpha all kappas no SL}) to the empirical values of $\alpha$ obtained by training actual CNNs and averaging their outputs across the ensemble.

%%%%%%%%%%%%%%%%%%%%%%%%%%%%%%%%%%%%%%%%%%%%%%%%%%%%%%%%%%%%%%%%%%%%%%%%%%%%%%%%%%%%%%%%%%

\subsection{Feature learning phase transition in the CNN model}
\label{subsection: Feature learning phase transition}
At this point there is evidence that our self-consistent shifted target approach works well within the feature learning regime of the toy model. Indeed GP is sub-optimal here, since it does not represent the CNN's weight sharing present in the teacher network. Weight sharing is intimately tied with feature learning in the first layer, since it aggregates the information coming from all convolutional windows to refine a single set of repeating convolution-filters. Empirically, we observed a large performance gap of finite $C$ CNNs over the infinite-$C$ (GP) limit, which was also observed previously in more realistic settings \cite{lee2020finite, geiger2020disentangling, novak2018bayesian}. 
Taken together with the existence of a clear feature in the teacher, a natural explanation for this performance gap is that feature learning, which is completely absent in GPs, plays a major role in the behavior of finite $C$ CNNs.

To analyze this we wish to track how the feature of the teacher $\w^*$ are reflected in the student network's first layer weights $\w_c$ across training time (after reaching equilibrium) and across training realizations. 
However, as our formalism deals with \textit{ensembles of DNNs}, computing averages of $\w_c$ with respect to these ensembles would simply give zero. Indeed, the chance of a DNN with specific parameters $\theta=\left\{ a_{i,c},\, \w_{c}\right\} $ appearing is the same as that of $-\theta$. 
Consequently, to detect feature learning the first reasonable object to examine is the empirical covariance matrix 
$\Sigma_W = \frac{S}{C} WW^\transpose $,  
where the matrix $W \in \bbR^{S \times C}$ has $\w_c$ as its $c$'th column. 
This $\Sigma_W$ is invariant under such a change of signs and provides important information on the statistics of $\w_c$. 

As shown in App. \ref{appendix: FL phase trans}, using our field-theory or function-space formulation, we find that to leading order in $1/C$ the ensemble average of the empirical covariance matrix, for a teacher with a single feature $\w^*$, is
\begin{equation}
\label{eq: Sigma_W rank-1 pert}
    \left\langle \left[\Sigma_{W}\right]_{ss'}\right\rangle = 
    \left(1 + \left(\frac{1}{\lambda}+\frac{n}{\sigma^{2}}\right)^{-1} \right)
    \delta_{ss'} + \frac{2}{C}\frac{\lambda}{\left(\lambda+\sigma^{2}/n\right)^{2}}w_{s}^{*}w_{s'}^{*}
    + O(1/C^2)
\end{equation}
A first conclusion that could be drawn here, is that given access to an ensemble of such trained CNNs, feature learning happens for any finite $C$ as a statistical property. We turn to discuss the more common setting where one wishes to use the features learned by a specific randomly chosen CNN from this ensemble. 

\begin{figure}[h]
    \centering
    (A)
    \includegraphics[height=4cm]{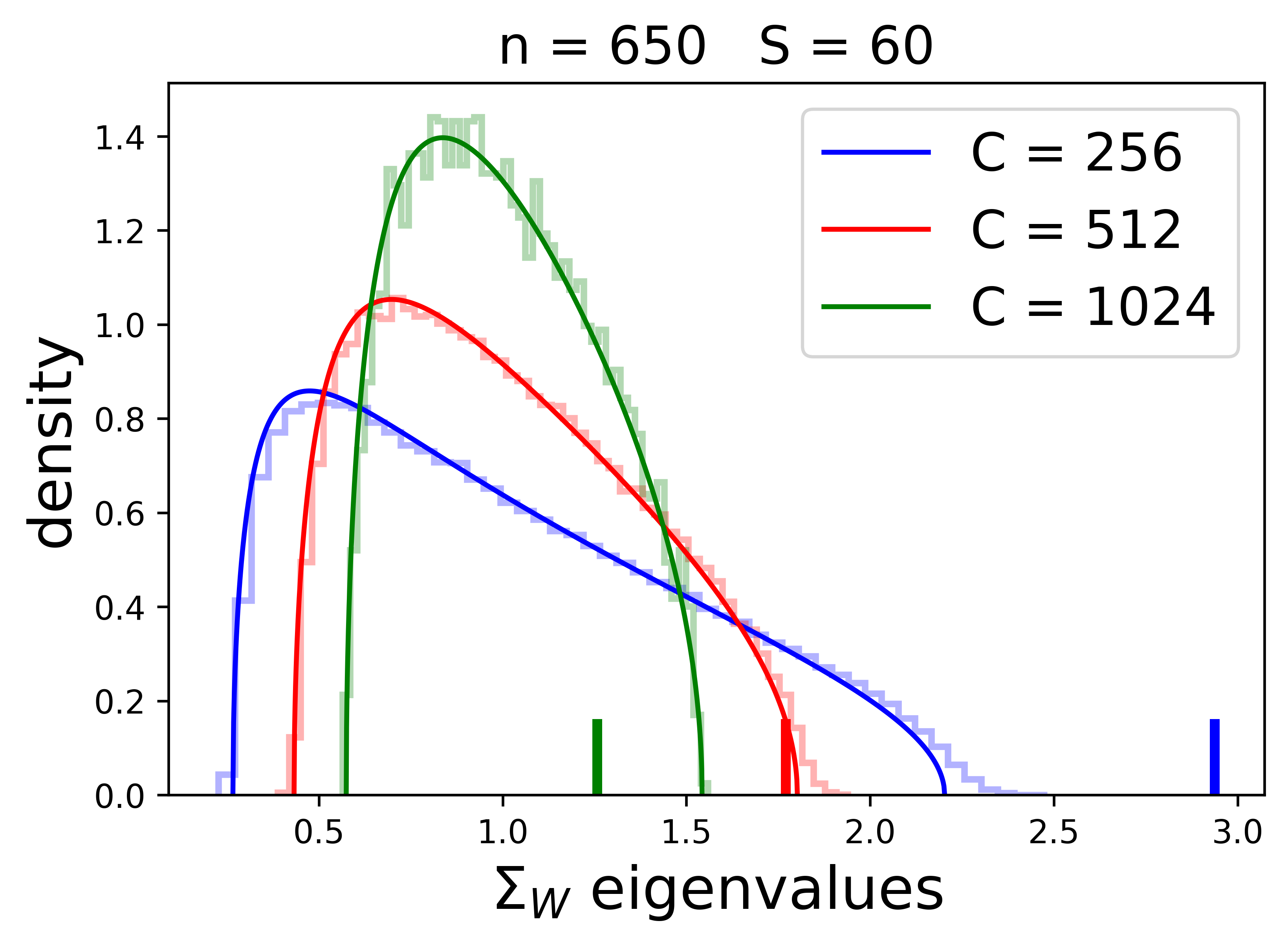}
    (B)
    \includegraphics[height=4cm]{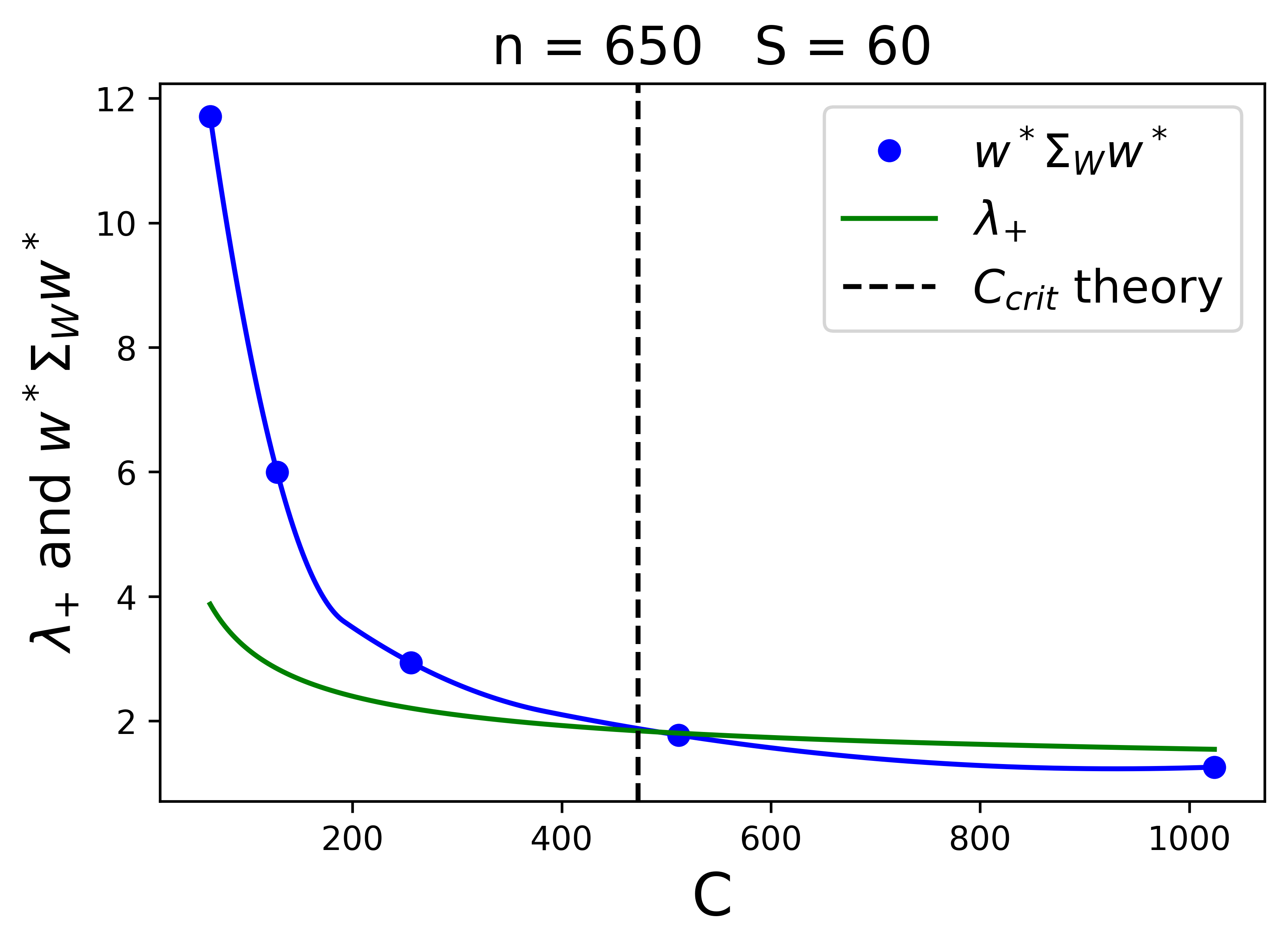}
    \caption{
\textbf{(A)}
Aggregated histograms of $\Sigma_W$ eigenvalues where $\Sigma_W = \frac{S}{C} WW^\transpose $ is the normalized empirical covariance matrix of the hidden layer weights during training. Different colors indicate varying number of channels, $C$. Solid smooth lines indicate the corresponding Marchenko-Pastur (MP) distributions with support on $\left[\lambda_-, \lambda_+ \right]$ where: $\lambda_{\pm} = \left( 1 \pm \sqrt{S/C}\right)^2$. 
The quantity $\cQ \equiv \w^{*\transpose} \Sigma_W \w^*$, which correlates with the SNR of the feature $\w^*$, is represented by thick short bars. 
For large $C$, $\cQ$ remains within the MP bulk whereas for small $C$ it pops out.
\textbf{(B)}
The theoretical $\lambda_+$ curve and interpolated curve of $\cQ$ and intersect very close to the theoretically predicted value given in Eq. \ref{eq: C_crit}, here given by $C_{\rm{crit}} = 473$ (dashed vertical line).
}
\label{fig: phase trans}
\end{figure}

To this end, we follow Ref. \cite{Mahoney2018} and model $\Sigma_{W}$ as a Wishart matrix with a rank-one perturbation. 
The variance of the matrix and details of the rank one perturbation are then determined by the above equation. 
Consequently the eigenvalue distribution is expected to follow a spiked Marchenko-Pastur (MP), which was studied extensively in \cite{benaych2012singular}. 
To test this modeling assumption, for each snapshot of training time (after reaching equilibrium) and noise realization we compute $\Sigma_{W}$'s eigenvalues and aggregate these across the ensemble.
In Fig. \ref{fig: phase trans} we plot the resulting empirical spectral distribution for varying values of $C$ while keeping $S$ fixed. Note that, differently from the usual spiked-MP model, varying $C$ here changes both the distribution of the MP bulk (which is determined by the ratio $S/C$) as well as the strength of the low-rank perturbation. 

Our main finding is a phase transition between two regimes which becomes sharp as one takes $n, S \rightarrow \infty$. In the regime of large $C$ the eigenvalue distribution of $\Sigma_W$ is indistinguishable from the MP distribution, whereas in the regime of small $C$ an outlier eigenvalue $\lambda_m$ departs from the support of the bulk MP distribution and the associated top eigenvector has a non-zero overlap with $\w^*$, see Fig. \ref{fig: phase trans}.
We refer to the latter as the feature-learning regime, since the feature $\w^*$ is manifested in the spectrum of the students weights, whereas the former is the non-feature learning regime.
We use the quantity $\cQ \equiv \w^{*\transpose} \Sigma_W \w^*$ as a surrogate for $\lambda_m$, as it is valid on both sides of the transition. Having established the correspondence to the MP plus low rank model, we can use the results of \cite{benaych2012singular} to find the exact location of the phase transition, which occurs at the critical value $C_{\rm{crit}}$ given by
\begin{equation}
\label{eq: C_crit}    
    C_{\rm{crit}} = \frac{4}{S\left(S^{-1}+\left(\sigma^{2}/n\right)S\right)^{4}}
    \left(1 + \left(S^2 + \frac{n}{\sigma^{2}}\right)^{-1} \right)
    + O\left(1 + \left(\frac{1}{\lambda}+\frac{n}{\sigma^{2}}\right)^{-1}\right)
\end{equation}
where we assumed for simplicity $N=S$ so that $\lambda=S^{-2}$. 

%%%%%%%%%%%%%%%%%%%%%%%%%%%%%%%%%%%%%%%%%%%%%%%%%%%%%%%%%%%%%%%%%%%%%%%%%%%%%%%%%%%%%%%%%%

\subsection{Two-layer FCN with average pooling and quadratic activations}
\label{subsect: 2-layer FCN}
Another setting where GPs are expected to under-perform finite-DNNs is the case of quadratic fully connected teacher and student DNNs where the teacher is rank-1, also known as the phase retrieval problem \cite{mannelli2020optimization}. 
Here we consider some positive target of the form 
$g(\x)=(\w_* \cdot \x)^2 - \sigma_w^2 ||\x||^2$ where $\w_*,\x \in {\bbR}^d$ and a student DNN given by 
$f(\x) = \sum_{m=1}^M (\w_m \cdot \x)^2 - \sigma_w^2 ||\x||^2$. We consider training this DNN on $n$ train points  $\left\{ \x_{\mu}\right\} _{\mu=1}^{n}$ using noisy GD training with weight decay $\gamma = 2M \sigma^2/\sigma_w^2$. 

Similarly to the previous toy model, here too the GP associated with the student at large $M$ (and finite $\sigma^2$) overlooks a qualitative feature of the finite DNN --- the fact that the first term in $f(\x)$ is non-negative. 
Interestingly, this feature provides a strong performance boost \cite{mannelli2020optimization} in the $\sigma^2 \rightarrow 0$ limit compared to the associated GP. Namely the DNN, even at large $M$, performs well for $n > 2 d$ \cite{mannelli2020optimization} whereas the associated GP is expected to work well only for $n=O(d^2)$ \cite{Cohen2019}. 

We wish to solve for the predictions of this model with our self consistent GP based approach. 
As shown in App. \ref{appendix: LenkasModel}, the cumulants of this model can be obtained from the following cumulant generating function  
\begin{align}
    \cC(t_1,...,t_{n+1}) &= - \frac{M}{2}\Tr\left(\log\left[I -2M^{-1} \sigma_w^2\sum_{\mu} it_{\mu} \x_{\mu}\x_{\mu}^\transpose \right]\right) - \sum_{\mu=1}^{n+1} it_{\mu}\sigma_w^2 ||\x_{\mu}||^2
\end{align}
The associated GP kernel is given by $K(\x_{\mu},\x_{\nu}) = 2 M^{-1} \sigma_w^4 (\x_{\mu}\cdot \x_{\nu})^2$. 
Following this, the target shift equation, at the saddle point level, appears as 
\begin{align}
\Delta g_{\nu} &= -\sum_{\mu} K(\x_{\nu},\x_{\mu}) \frac{\hat{\delta} g_{\mu}}{\sigma^2} + \sigma_w^2 \x_{\nu}^\transpose \left[I- 2 M^{-1} \sigma_w^2\sum_{\mu} \frac{\hat{\delta} g_{\mu}}{\sigma^2} \x_{\mu}\x_{\mu}^\transpose \right]^{-1} \x_{\nu}  -\sigma_w^2 ||\x_{\nu}||^2 
\end{align}

In App. \ref{appendix: LenkasModel}, we solve these equations numerically for $\sigma^2=10^{-5}$ and show that our approach captures the correct $n = 2d$ threshold value. 
An analytic solution of these equations at low $\sigma^2$ using EK or other continuum approximations is left for future work (see Refs. \cite{Cohen2019, bordelon2020spectrum, Canatar2021} for potential approaches). 
As a first step towards this goal, in App. \ref{appendix: LenkasModel} we consider the simpler case of $\sigma^2=1$ and derive the asymptotics of the learning curves which deviate strongly from those of GP for $M \ll d$.    

%To proceed analytically we again use the continuum EK limit and find that the discrepancy remains of the form $\hat{\delta}g_{\mu} = \alpha g(\x_{\mu}) + \beta \sigma_w^2 ||\x_{\mu}||^2$. 
%The target shift equations then become two coupled non-linear equation for $\alpha$ and $\beta$ (see App. \ref{appendix: LenkasModel}) which one can Taylor expand around good performance ($\alpha,\beta=0$) to yield the scaling of $\alpha$ and $\beta$ with $n$. 
%For $d\gg 1$ and $n \gg \sigma^2 Md \sigma_w^{-4}$ we find  
%\begin{align}
%\alpha &= \frac{5}{18}\frac{\sigma^2 Md}{2\sigma_w^4 n} \quad \beta = \frac{4}{18}\frac{\sigma^2 Md}{2\sigma_w^4 n}
%\end{align}
%Thus $n$ scaling as $Md$ indeed ensures good performance. 

%%%%%%%%%%%%%%%%%%%%%%%%%%%%%%%%%%%%%%%%%%%%%%%%%%%%%%%%%%%%%%%%%%%%%%%

\section{Discussion}
\label{section: discussion}
In this work we presented a correspondence between ensembles of finite DNNs trained with noisy gradients and GPs trained on a shifted target. 
The shift in the target can be found by solving a set of self consistent equations for which we give a general form.
We found explicit expressions for these equations for the case of a 2-layer linear CNN and a non-linear FCN, and solved them analytically and numerically. 
For the former model, we performed numerical experiments on CNNs that agree well with our theory both in the GP regime \textit{and well away from it}, i.e. for small number of channels $C$, thus accounting for strong finite $C$ effects. 
For the latter model, the numerical solution of these equations capture a remarkable and subtle effect in these DNNs which the GP approach completely overlooks --- the $n=2d$ threshold value. 

Considering feature learning in the CNN model, we found that averaging over ensembles of such networks always leads to a form of feature learning. Namely, the teacher always leaves a signature on the statistics of the student's weights.
However, feature learning is usually considered at the level of a single DNN instance rather than an ensemble of DNNs. 
Focusing on this case, we show numerically that the eigenvalues of $\Sigma_W$, the student hidden weights covariance matrix, follow a Marchenko–Pastur distribution plus a rank-1 perturbation. 
We then use our approach to derive the critical number of channels $C_{\rm{crit}}$ below which the student is in a feature learning regime. 

There are many directions for future research. 
Our toy models where chosen to be as simple as possible in order to demonstrate the essence of our theory on problems where lazy learning grossly under-performs finite-DNNs. 
Even within this setting, various extensions are interesting to consider such as adding more features to the teacher CNN (e.g. biases or a subset of linear functions which are more favorable), studying linear CNNs with overlapping convolutional windows, or deeper linear CNNs. 
As for non-linear CNNs, we believe it is possible to find the exact cumulants of any order for a variety of toy CNNs involving, for example, quadratic activation functions. For other cases it may be useful to develop methods for characterizing and approximating the cumulants. 

More generally, we advocated here a physics-style methodology using approximations, self-consistency checks, and experimental tests. As DNNs are very complex experimental systems, we believe this mode of research is both appropriate and necessary. Nonetheless we hope the insights gained by our approach would help generate a richer and more relevant set of toy models on which mathematical proofs could be made.

%%%%%%%%%%%%%%%%%%%%%%%%%%%%%%%%%%%%%%%%%%%%%%%%%%%%%%%%%%%%%%%%%%%%%%%%%%%%%%%%%%%%%%%%%%%%%%%%%%%%%%%%%%%%%%%%%%%%%%%%%%%%%%%%%%

\bibliographystyle{natbib.bst}
\bibliography{PRE_main}% Produces the bibliography via BibTeX.

\begin{thebibliography}{}

\bibitem[Alemohammad {\em et~al.}(2020)Alemohammad, Wang, Balestriero, and
  Baraniuk]{alemohammad2020recurrent}
Alemohammad, S., Wang, Z., Balestriero, R., and Baraniuk, R. (2020).
\newblock The recurrent neural tangent kernel.
\newblock {\em arXiv preprint arXiv:2006.10246\/}.

\bibitem[Andreassen and Dyer(2020)Andreassen and
  Dyer]{andreassen2020asymptotics}
Andreassen, A. and Dyer, E. (2020).
\newblock Asymptotics of wide convolutional neural networks.
\newblock {\em arXiv preprint arXiv:2008.08675\/}.

\bibitem[{Arora} {\em et~al.}(2019){Arora}, {Du}, {Hu}, {Li}, {Salakhutdinov},
  and {Wang}]{Arora2019}
{Arora}, S., {Du}, S.~S., {Hu}, W., {Li}, Z., {Salakhutdinov}, R., and {Wang},
  R. (2019).
\newblock {On Exact Computation with an Infinitely Wide Neural Net}.
\newblock {\em arXiv e-prints\/}, page arXiv:1904.11955.

\bibitem[Baldi and Hornik(1989)Baldi and Hornik]{baldi1989neural}
Baldi, P. and Hornik, K. (1989).
\newblock Neural networks and principal component analysis: Learning from
  examples without local minima.
\newblock {\em Neural networks\/}, {\bf 2}(1), 53--58.

\bibitem[Benaych-Georges and Nadakuditi(2011)Benaych-Georges and
  Nadakuditi]{benaych2011eigenvalues}
Benaych-Georges, F. and Nadakuditi, R.~R. (2011).
\newblock The eigenvalues and eigenvectors of finite, low rank perturbations of
  large random matrices.
\newblock {\em Advances in Mathematics\/}, {\bf 227}(1), 494--521.

\bibitem[Benaych-Georges and Nadakuditi(2012)Benaych-Georges and
  Nadakuditi]{benaych2012singular}
Benaych-Georges, F. and Nadakuditi, R.~R. (2012).
\newblock The singular values and vectors of low rank perturbations of large
  rectangular random matrices.
\newblock {\em Journal of Multivariate Analysis\/}, {\bf 111}, 120--135.

\bibitem[Bordelon {\em et~al.}(2020)Bordelon, Canatar, and
  Pehlevan]{bordelon2020spectrum}
Bordelon, B., Canatar, A., and Pehlevan, C. (2020).
\newblock Spectrum dependent learning curves in kernel regression and wide
  neural networks.

\bibitem[Canatar {\em et~al.}(2021)Canatar, Bordelon, and
  Pehlevan]{Canatar2021}
Canatar, A., Bordelon, B., and Pehlevan, C. (2021).
\newblock Spectral bias and task-model alignment explain generalization in
  kernel regression and infinitely wide neural networks.
\newblock {\em Nature Communications\/}, {\bf 12}(1).

\bibitem[Chizat {\em et~al.}(2019)Chizat, Oyallon, and Bach]{chizat2019lazy}
Chizat, L., Oyallon, E., and Bach, F. (2019).
\newblock On lazy training in differentiable programming.
\newblock In {\em Advances in Neural Information Processing Systems\/}, pages
  2937--2947.

\bibitem[{Cohen} {\em et~al.}(2019){Cohen}, {Malka}, and {Ringel}]{Cohen2019}
{Cohen}, O., {Malka}, O., and {Ringel}, Z. (2019).
\newblock {Learning Curves for Deep Neural Networks: A Gaussian Field Theory
  Perspective}.
\newblock {\em arXiv e-prints\/}, page arXiv:1906.05301.

\bibitem[Daniels(1954)Daniels]{Daniels1954}
Daniels, H.~E. (1954).
\newblock {Saddlepoint Approximations in Statistics}.
\newblock {\em The Annals of Mathematical Statistics\/}, {\bf 25}(4), 631 --
  650.

\bibitem[Dyer and Gur-Ari(2020)Dyer and Gur-Ari]{Dyer2020Asymptotics}
Dyer, E. and Gur-Ari, G. (2020).
\newblock Asymptotics of wide networks from feynman diagrams.
\newblock In {\em International Conference on Learning Representations\/}.

\bibitem[Geiger {\em et~al.}(2020)Geiger, Spigler, Jacot, and
  Wyart]{geiger2020disentangling}
Geiger, M., Spigler, S., Jacot, A., and Wyart, M. (2020).
\newblock Disentangling feature and lazy training in deep neural networks.
\newblock {\em Journal of Statistical Mechanics: Theory and Experiment\/}, {\bf
  2020}(11), 113301.

\bibitem[Geiger {\em et~al.}(2021)Geiger, Petrini, and
  Wyart]{geiger2021landscape}
Geiger, M., Petrini, L., and Wyart, M. (2021).
\newblock Landscape and training regimes in deep learning.
\newblock {\em Physics Reports\/}.

\bibitem[Ghorbani {\em et~al.}(2020)Ghorbani, Mei, Misiakiewicz, and
  Montanari]{ghorbani2020neural}
Ghorbani, B., Mei, S., Misiakiewicz, T., and Montanari, A. (2020).
\newblock When do neural networks outperform kernel methods?
\newblock {\em arXiv preprint arXiv:2006.13409\/}.

\bibitem[Gunasekar {\em et~al.}(2018)Gunasekar, Lee, Soudry, and
  Srebro]{gunasekar2018implicit}
Gunasekar, S., Lee, J., Soudry, D., and Srebro, N. (2018).
\newblock Implicit bias of gradient descent on linear convolutional networks.
\newblock {\em arXiv preprint arXiv:1806.00468\/}.

\bibitem[Helias and Dahmen(2019)Helias and Dahmen]{helias2019statistical}
Helias, M. and Dahmen, D. (2019).
\newblock Statistical field theory for neural networks.
\newblock {\em arXiv preprint arXiv:1901.10416\/}.

\bibitem[Hron {\em et~al.}(2020)Hron, Bahri, Sohl-Dickstein, and
  Novak]{hron2020infinite}
Hron, J., Bahri, Y., Sohl-Dickstein, J., and Novak, R. (2020).
\newblock Infinite attention: Nngp and ntk for deep attention networks.
\newblock In {\em International Conference on Machine Learning\/}, pages
  4376--4386. PMLR.

\bibitem[{Jacot} {\em et~al.}(2018){Jacot}, {Gabriel}, and
  {Hongler}]{Jacot2018}
{Jacot}, A., {Gabriel}, F., and {Hongler}, C. (2018).
\newblock {Neural Tangent Kernel: Convergence and Generalization in Neural
  Networks}.
\newblock {\em arXiv e-prints\/}, page arXiv:1806.07572.

\bibitem[Lampinen and Ganguli(2018)Lampinen and Ganguli]{lampinen2018analytic}
Lampinen, A.~K. and Ganguli, S. (2018).
\newblock An analytic theory of generalization dynamics and transfer learning
  in deep linear networks.
\newblock {\em arXiv preprint arXiv:1809.10374\/}.

\bibitem[Lee {\em et~al.}(2018)Lee, Sohl-dickstein, Pennington, Novak,
  Schoenholz, and Bahri]{lee2017deep}
Lee, J., Sohl-dickstein, J., Pennington, J., Novak, R., Schoenholz, S., and
  Bahri, Y. (2018).
\newblock Deep neural networks as gaussian processes.
\newblock In {\em International Conference on Learning Representations\/}.

\bibitem[Lee {\em et~al.}(2019)Lee, Xiao, Schoenholz, Bahri, Novak,
  Sohl-Dickstein, and Pennington]{lee2019wide}
Lee, J., Xiao, L., Schoenholz, S.~S., Bahri, Y., Novak, R., Sohl-Dickstein, J.,
  and Pennington, J. (2019).
\newblock Wide neural networks of any depth evolve as linear models under
  gradient descent.
\newblock {\em arXiv preprint arXiv:1902.06720\/}.

\bibitem[Lee {\em et~al.}(2020)Lee, Schoenholz, Pennington, Adlam, Xiao, Novak,
  and Sohl-Dickstein]{lee2020finite}
Lee, J., Schoenholz, S.~S., Pennington, J., Adlam, B., Xiao, L., Novak, R., and
  Sohl-Dickstein, J. (2020).
\newblock Finite versus infinite neural networks: an empirical study.
\newblock {\em arXiv preprint arXiv:2007.15801\/}.

\bibitem[Li and Sompolinsky(2020)Li and Sompolinsky]{li2020statistical}
Li, Q. and Sompolinsky, H. (2020).
\newblock Statistical mechanics of deep linear neural networks: The
  back-propagating renormalization group.
\newblock {\em arXiv preprint arXiv:2012.04030\/}.

\bibitem[Malach {\em et~al.}(2021)Malach, Kamath, Abbe, and
  Srebro]{malach2021quantifying}
Malach, E., Kamath, P., Abbe, E., and Srebro, N. (2021).
\newblock Quantifying the benefit of using differentiable learning over tangent
  kernels.
\newblock {\em arXiv preprint arXiv:2103.01210\/}.

\bibitem[Mannelli {\em et~al.}(2020)Mannelli, Vanden-Eijnden, and
  Zdeborov{\'a}]{mannelli2020optimization}
Mannelli, S.~S., Vanden-Eijnden, E., and Zdeborov{\'a}, L. (2020).
\newblock Optimization and generalization of shallow neural networks with
  quadratic activation functions.
\newblock {\em arXiv preprint arXiv:2006.15459\/}.

\bibitem[{Martin} and {Mahoney}(2018){Martin} and {Mahoney}]{Mahoney2018}
{Martin}, C.~H. and {Mahoney}, M.~W. (2018).
\newblock {Implicit Self-Regularization in Deep Neural Networks: Evidence from
  Random Matrix Theory and Implications for Learning}.
\newblock {\em arXiv e-prints\/}, page arXiv:1810.01075.

\bibitem[Matthews {\em et~al.}(2018)Matthews, Rowland, Hron, Turner, and
  Ghahramani]{matthews2018gaussian}
Matthews, A. G. d.~G., Rowland, M., Hron, J., Turner, R.~E., and Ghahramani, Z.
  (2018).
\newblock Gaussian process behaviour in wide deep neural networks.
\newblock {\em arXiv preprint arXiv:1804.11271\/}.

\bibitem[Mccullagh(2017)Mccullagh]{Mccullagh2017}
Mccullagh, P. (2017).
\newblock {\em Tensor Methods in Statistics\/}.
\newblock Dover Books on Mathematics.

\bibitem[Naveh {\em et~al.}(2020)Naveh, Ben-David, Sompolinsky, and
  Ringel]{naveh2020predicting}
Naveh, G., Ben-David, O., Sompolinsky, H., and Ringel, Z. (2020).
\newblock Predicting the outputs of finite networks trained with noisy
  gradients.
\newblock {\em arXiv preprint arXiv:2004.01190\/}.

\bibitem[Neal(1996)Neal]{neal1996priors}
Neal, R.~M. (1996).
\newblock Priors for infinite networks.
\newblock In {\em Bayesian Learning for Neural Networks\/}, pages 29--53.
  Springer.

\bibitem[Note1(????)Note1]{Note1}
Note1 (????).
\newblock Generically this requires $C$ dependent and layer dependent weight
  decay.

\bibitem[Note2(????)Note2]{Note2}
Note2 (????).
\newblock The $||{\protect \mathbf x}||^2$ shift is not part of the original
  model but has only a superficial shift effect useful for book-keeping later
  on.

\bibitem[{Novak} {\em et~al.}(2018){Novak}, {Xiao}, {Lee}, {Bahri}, {Yang},
  {Abolafia}, {Pennington}, and {Sohl-Dickstein}]{novak2018bayesian}
{Novak}, R., {Xiao}, L., {Lee}, J., {Bahri}, Y., {Yang}, G., {Abolafia}, D.~A.,
  {Pennington}, J., and {Sohl-Dickstein}, J. (2018).
\newblock {Bayesian Deep Convolutional Networks with Many Channels are Gaussian
  Processes}.
\newblock {\em arXiv e-prints\/}, page arXiv:1810.05148.

\bibitem[Rasmussen and Williams(2005)Rasmussen and Williams]{Rasmussen2005}
Rasmussen, C.~E. and Williams, C. K.~I. (2005).
\newblock {\em Gaussian Processes for Machine Learning (Adaptive Computation
  and Machine Learning)\/}.
\newblock The MIT Press.

\bibitem[Refinetti {\em et~al.}(2021)Refinetti, Goldt, Krzakala, and
  Zdeborov{\'a}]{refinetti2021classifying}
Refinetti, M., Goldt, S., Krzakala, F., and Zdeborov{\'a}, L. (2021).
\newblock Classifying high-dimensional gaussian mixtures: Where kernel methods
  fail and neural networks succeed.
\newblock {\em arXiv preprint arXiv:2102.11742\/}.

\bibitem[Saxe {\em et~al.}(2013)Saxe, McClelland, and Ganguli]{saxe2013exact}
Saxe, A.~M., McClelland, J.~L., and Ganguli, S. (2013).
\newblock Exact solutions to the nonlinear dynamics of learning in deep linear
  neural networks.
\newblock {\em arXiv preprint arXiv:1312.6120\/}.

\bibitem[Sollich and Williams(2004)Sollich and
  Williams]{sollich2004understanding}
Sollich, P. and Williams, C.~K. (2004).
\newblock Understanding gaussian process regression using the equivalent
  kernel.
\newblock In {\em International Workshop on Deterministic and Statistical
  Methods in Machine Learning\/}, pages 211--228. Springer.

\bibitem[Yaida(2020)Yaida]{yaida2020non}
Yaida, S. (2020).
\newblock Non-gaussian processes and neural networks at finite widths.
\newblock In {\em Mathematical and Scientific Machine Learning\/}, pages
  165--192. PMLR.

\bibitem[Yang and Hu(2020)Yang and Hu]{yang2020feature}
Yang, G. and Hu, E.~J. (2020).
\newblock Feature learning in infinite-width neural networks.
\newblock {\em arXiv preprint arXiv:2011.14522\/}.

\bibitem[Yosinski {\em et~al.}(2014)Yosinski, Clune, Bengio, and
  Lipson]{Yosinski2014}
Yosinski, J., Clune, J., Bengio, Y., and Lipson, H. (2014).
\newblock How transferable are features in deep neural networks?
\newblock In Z.~Ghahramani, M.~Welling, C.~Cortes, N.~Lawrence, and K.~Q.
  Weinberger, editors, {\em Advances in Neural Information Processing
  Systems\/}, volume~27. Curran Associates, Inc.

\bibitem[Yu {\em et~al.}(2013)Yu, Seltzer, Li, Huang, and Seide]{yu2013feature}
Yu, D., Seltzer, M., Li, J., Huang, J.-T., and Seide, F. (2013).
\newblock Feature learning in deep neural networks - studies on speech
  recognition.
\newblock In {\em International Conference on Learning Representations\/}.

\bibitem[Zavatone-Veth and Pehlevan(2021)Zavatone-Veth and
  Pehlevan]{zavatone2021exact}
Zavatone-Veth, J.~A. and Pehlevan, C. (2021).
\newblock Exact priors of finite neural networks.
\newblock {\em arXiv preprint arXiv:2104.11734\/}.

\end{thebibliography}

%%%%%%%%%%%%%%%%%%%%%%%%%%%%%%%%%%%%%%%%%%%%%%%%%%%%%%%%%%%%%%%%%%%%%%%%%%%%%%%%%%%%%%%%%%%%%%%%%%%%%%%%%%%%%%%%%%%%%%%%%%%%%%%%%%

\newpage
\appendix

\numberwithin{equation}{section}
\makeatletter 
% "activate" the preparatory code, but for section-level headers only
\newcommand{\section@cntformat}{Appendix \thesection:\ }
\makeatother

\section{Derivation of the target shift equations}
\label{appendix: TS}
\subsection{The partition function in terms of dual variables}
\label{appendix: TS dual Z}
Consider the general setting of Bayesian inference with Gaussian measurement noise (or equivalently a DNN trained with MSE loss, weight decay, and white noise added to the gradients). 
Let $\left\{ \x_{\mu}, g_{\mu}\right\} _{\mu=1}^{n}$ denote the inputs and targets on the training set and let $\x_* \equiv \x_{n+1}$  be the test point. 
Denote the prior (or equivalently the equilibrium distribution of a DNN trained with no data) by $P_0(\vec{f})$ where $f_{\mu}=f(\x_{\mu}) \in \bbR$ is the output of the model, and $\vec{f}\equiv\left(f_{1},\dots,f_{n},f_{n+1}\right) \in \bbR^{n+1}$. 
The model's predictions (or equivalently the ensemble averaged DNN output) on the point $\x_{\mu}$ can be obtained by
\begin{equation}
\label{appEq: pa_J log Z}
    \forall\mu\in\left\{ 1,\dots,n+1\right\}: \qquad
    \left\langle f_{\mu}\right\rangle = \partial_{J_{\mu}}\eval{\log Z\left(\vec{J}\right)}_{\vec{J}=\vec{0}}
\end{equation}
with the following partition function 
\begin{equation}
\label{EqApp:ZJ}
    Z\left(\vec{J}\right)=\int d\vec{f}P_{0}\left(\vec{f}\right)\exp\left(-\frac{1}{2\sigma^{2}}\sum_{\mu=1}^{n}\left(f_{\mu}-g_{\mu}\right)^{2}+\sum_{\mu}J_{\mu}f_{\mu}\right)    
\end{equation}
where unless explicitly written otherwise, summations over $\mu$ run from $1$ to $n+1$ (i.e. include the test point). 
Here we commit to the MSE loss which facilitates the derivation, and in App. \ref{appendix: TS derivs} we give an alternative derivation that may also be applied to other losses such as cross-entropy.
Our goal in this appendix is to establish that the target shift equations are in fact saddle point equations of the partition function \ref{EqApp:ZJ} following some transformations on the variables of integration. 
To this end, consider the cumulant generating function of $P_{0}\left(\vec{f}\right)$ given by \begin{align}
\label{appEq: cumulant gen func log}
\cC\left(\vec{t}\right)=\log\left(\int_{-\infty}^{\infty}d\vec{f}e^{i\sum_{\mu}t_{\mu}f_{\mu}}P_{0}\left(\vec{f}\right)\right)
\end{align}
or expressed via the cumulant tensors:
\begin{equation}
\label{appEq: cumulant gen func kappas}    
    \cC\left(\vec{t}\right)=\sum_{r=2}^{\infty}\frac{1}{r!}\sum_{\mu_{1},\dots,\mu_{r}=1}^{n+1}\kappa_{\mu_{1},\dots,\mu_{r}}it_{\mu_{1}}\cdots it_{\mu_{r}}
\end{equation}
where the sum over the cumulant tensors $\kappa_{\mu_{1},\dots,\mu_{r}}$ does not include $r=1$ since our DNN priors are assumed to have zero mean. Notably one can re-express $P_{0}\left(\vec{f}\right)$ as the inverse Fourier transform of $e^{\cC(\vec{t})}$: 
\begin{equation}
\label{appEq: prior via cumulants}    
    P_{0}\left(\vec{f}\right)\propto\int d\vec{t}\exp\left(-i\sum_{\mu}t_{\mu}f_{\mu}+\cC\left(\vec{t}\right)\right)
\end{equation}
Plugging this in Eq. \ref{EqApp:ZJ} we obtain
\begin{align}
    Z\left(\vec{J}\right)\propto\iint d\vec{f}d\vec{t}\exp\left(-\frac{1}{2\sigma^{2}}\sum_{\mu=1}^{n}\left(f_{\mu}-g_{\mu}\right)^{2}+\sum_{\mu}\left(J_{\mu}-it_{\mu}\right)f_{\mu}+\cC\left(\vec{t}\right)\right)
\end{align}
where for clarity we do not keep track of multiplicative $\pi$ factors that have no effect on moments of $f_{\mu}$. 
As the term in the exponent (the action) is quadratic in $\left\{ f_{\mu}\right\} _{\mu=1}^{n}$ and linear in $f_{n+1}$ these can be integrated out to yield an equivalent partition function phrased solely in terms of $t_{1},...,t_{n+1}$:
\begin{equation}
\label{appEq: Z(J) = e^(-s)}
    Z\left(\vec{J}\right)\propto\int dt_{1}\cdots dt_{n}e^{-\cS\left(\vec{t},\vec{J}\right)}
\end{equation}
where the action is now
\begin{align}
\label{appEq: action of t,J}
    \cS=-\cC\left(t_{1},\dots,t_{n},-iJ_{n+1}\right)+\sum_{\mu=1}^{n}\left[\frac{\sigma^{2}}{2}t_{\mu}^{2}+it_{\mu}g_{\mu}+J_{\mu}\left(i\sigma^{2}t_{\mu}-g_{\mu}\right)-\frac{\sigma^{2}}{2}J_{\mu}^{2}\right]
\end{align}
The identification $t_{n+1}=-iJ_{n+1}$ arises from the delta function:
\begin{equation}
    \frac{1}{2\pi}\int_{-\infty}^{\infty}df_{n+1}e^{-if_{n+1}\left(iJ_{n+1}+t_{n+1}\right)}=\delta\left(iJ_{n+1}+t_{n+1}\right)
\end{equation} 
Recall that 
$\left\langle f_{\mu}\right\rangle = \partial_{J_{\mu}}\eval{\log Z\left(\vec{J}\right)}_{\vec{J}=\vec{0}}$ 
and notice that the first term in Eq. \ref{appEq: action of t,J} (the cumulant generating function, $\cC$) depends on $J_{n+1}$ and not on $\left\{ J_{\mu}\right\} _{\mu=1}^{n}$ whereas the rest of the action depends on $\left\{ J_{\mu}\right\} _{\mu=1}^{n}$ and not on $J_{n+1}$. 
Thus, for training points $\left\langle f_{\mu}\right\rangle $ amounts to the average of $g_{\mu} -i \sigma^2 t_\mu$, and so we identify 
\begin{align}
\label{appEq: mean it delta g}
    \forall\mu\in\left\{ 1,\dots,n\right\} :\qquad\langle it_{\mu}\rangle=\frac{g_{\mu}-\langle f_{\mu}\rangle}{\sigma^{2}}\equiv\frac{\langle\hat{\delta}g_{\mu}\rangle}{\sigma^{2}}
\end{align}
where $\langle \cdots \rangle$ denotes an expectation value using $Z(\vec{J}=\vec{0})$. 
We comment that the above relation holds also for any (non-mixed) cumulants of $i t_{\mu}$ and $\hat{\delta} g_\mu / \sigma^2$ except the covariance, where a constant difference appears due to the $O(J^2)$ term in the action, namely 
\begin{align}
\label{appEq: var it delta g}
    \left\langle it_{\mu}it_{\nu}\right\rangle -\left\langle it_{\mu}\right\rangle \left\langle it_{\nu}\right\rangle = \frac{1}{\sigma^{4}}\left(\left\langle \hat{\delta}g_{\mu}\hat{\delta}g_{\nu}\right\rangle -\left\langle \hat{\delta}g_{\mu}\right\rangle \left\langle \hat{\delta}g_{\nu}\right\rangle \right) - \frac{1}{\sigma^{2}}\delta_{\mu\nu}
\end{align}
In the GP case the r.h.s. of Eq. \ref{appEq: var it delta g} would equal simply $-\tilde{K}^{-1}_{\mu\nu}$, since on the training set the posterior covariance of $\hat{\delta} g$ is the same as that of $f$ and for a GP takes the form
\begin{align}
\label{appEq: Sigma GP}
    \Sigma &= K - K\tilde{K}^{-1}K
	\\ &= \nonumber	
	K - \left(K+\sigma^{2}I-\sigma^{2}I\right)\tilde{K}^{-1}K
	\\ &= \nonumber	
	\sigma^{2}\tilde{K}^{-1}K =	
	\sigma^{2}\tilde{K}^{-1}\left(K+\sigma^{2}I-\sigma^{2}I\right)
	\\ &= \nonumber	
	\sigma^{2}I-\sigma^{4}\tilde{K}^{-1}
\end{align}
namely $\Sigma_{\mu\nu}=\sigma^{2}\delta_{\mu\nu}-\sigma^{4}\tilde{K}_{\mu\nu}^{-1}$ and thus 
$\frac{1}{\sigma^{4}}\Sigma_{\mu\nu}-\frac{1}{\sigma^{2}}\delta_{\mu\nu}=-\tilde{K}_{\mu\nu}^{-1}$.
The reader should not be alarmed by having a negative definite covariance matrix for $it_{\mu}$, since $it_{\mu}$ cannot be understood as a standard real random variable as its partition function contains imaginary terms.  

To make contact with GPs it is beneficial to expand $\cC(t_1, \dots, t_{n}, -iJ_{n+1})$ in terms of its cumulants, and split the second cumulant, describing the DNNs' NNGP kernel, from the rest. Namely, using Einstein summation
\begin{align}
\label{appEq: Ctilde def}
    \cC\left(t_{1}, \dots, t_{n}, -iJ_{n+1} \right) &= \frac{1}{2!}\kappa_{\mu_{1},\mu_{2}}it_{\mu_{1}}it_{\mu_{2}}+\frac{1}{3!}\kappa_{\mu_{1},\mu_{2},\mu_{3}}it_{\mu_{1}}it_{\mu_{2}}it_{\mu_{3}} + O((it)^4)
    \\ \nonumber
    \kappa_{\mu_1,\mu_2} &\equiv K(\x_{\mu_1},\x_{\mu_2}) \\ \nonumber
    \tilde{\cC}\left(t_{1}, \dots, t_{n}, -iJ_{n+1} \right) & \equiv \cC\left(t_{1}, \dots, t_{n}, -iJ_{n+1} \right) + \frac{1}{2} \kappa_{\mu_1, \mu_2} t_{\mu_1}t_{\mu_2}
\end{align}
Writing Eq. \ref{appEq: action of t,J} in this fashion gives the action:
\begin{align}
\label{appEq: S of t and J GP}
    \cS=-\tilde{\cC}\left(\vec{t}\right)-\frac{1}{2}\sum_{\mu_{1},\mu_{2}}\kappa_{\mu_{1},\mu_{2}}it_{\mu_{1}}it_{\mu_{2}}+\sum_{\mu=1}^{n}\left[-\frac{\sigma^{2}}{2}\left(it_{\mu}\right)^{2}+it_{\mu}g_{\mu}+J_{\mu}\left(i\sigma^{2}t_{\mu}-g_{\mu}\right)-\frac{\sigma^{2}}{2}J_{\mu}^{2}\right]
\end{align}

\subsection{Saddle point equation for the mean predictor}
\label{appendix: SP eqs}
Having arrived at the action \ref{appEq: S of t and J GP}, we can readily derive the saddle point equations for the training points by setting: 
\begin{equation}
\label{appEq: SP general}   
    \forall \nu\in\left\{ 1,\dots,n\right\}: \qquad 
    \eval{\pa_{it_{\nu}}\cS\left(\vec{t},\vec{J}\right)}_{\vec{J} = \vec{0}} = 0
\end{equation}
This corresponds to treating the variables $\left\{ it_{\mu}\right\} _{\mu=1}^{n}$ as non-fluctuating quantities, i.e. replacing them with their mean value: 
$it_\mu \to \left\langle it_{\mu}\right\rangle$. 
Performing this for the training set $\nu\in\left\{ 1,\dots,n\right\}$ yields 
\begin{equation}
\label{appEq: SP eq it train}
    \sum_{\mu=1}^{n}\left(\kappa_{\mu,\nu}+\sigma^{2}\delta_{\mu\nu}\right)\left\langle it_{\mu}\right\rangle =g_{\nu}-\Delta g_{\nu}
\end{equation}
where 
\begin{equation}
\label{appEq: SP eq Delta g}
    \Delta g_{\nu}=\sum_{r=3}^{\infty}\frac{1}{\left(r-1\right)!}\sum_{\mu_{1},\dots,\mu_{r-1}=1}^{n}\kappa_{\nu,\mu_{1},\dots,\mu_{r-1}}\left\langle it_{\mu_{1}}\right\rangle \cdots\left\langle it_{\mu_{r-1}}\right\rangle 
\end{equation}
where this target shift is related to $\cC$ of Eq. \ref{appEq: Ctilde def} by 
\begin{equation}
\label{appEq: Delta g Ctilde relation}    
    \Delta g_{\nu}=\partial_{it_{\nu}}\tilde{\cC}\left(t_{1},\dots,t_{n},t_{n+1}\right)
\end{equation}

Finally, we get the expression for the mean predictor at the test point by setting 
$\left\langle f_{*}\right\rangle = \eval{ \pa_{J_{*}}\log Z\left(\vec{J}\right)}_{\vec{J}=\vec{0}}$
and plugging in the SP values for $it_\mu$ on the training set from Eq. \ref{appEq: SP eq it train} and the target shift $\Delta g$ from Eq. \ref{appEq: SP eq Delta g}. This gives 
\begin{equation}
\label{appEq: mean pred test}
    \left\langle f_{*}\right\rangle =\Delta g_{*} + \sum^{n}_{\mu,\nu}K_{\mu}^{*}\tilde{K}_{\mu\nu}^{-1}\left(g_{\nu}-\Delta g_{\nu}\right)
\end{equation}

\subsection{Posterior covariance}
\label{appendix: cov SP + fluct}
\subsubsection{Posterior covariance on the test point}
The posterior covariance on the test point is important for determining the average MSE loss on the test-set, as the latter involves the MSE of the mean-predictor plus the posterior covariance. 
Concretely, we wish to calculate $\partial^2_{J_{*}} \log(Z)$, express it as an expectation value w.r.t $Z$ and calculate this using $Z$ with the self-consistent target shift. 
To this end, note that generally if $Z(J)= \int e^{-\cS(J)}$ then
\begin{equation}
    \partial^2_{J} \log(Z(J))=\langle \cS'(0)^2 \rangle_{Z(J=0)} - \langle \cS'(0) \rangle_{Z(J=0)}^2 - \langle \cS''(0)\rangle_{Z(J=0)}
\end{equation}
For the action in Eq. \ref{appEq: S of t and J GP} we have
\begin{equation}
    \cS'(0) = -\kappa_{*\nu}it_{\nu} - \underbrace{\partial_{it_{*}}\tilde{\cC}}_{\Delta g_{*}}
    \qquad
    \cS''(0) = -\kappa_{**} - \partial_{it_{*}}^{2}\tilde{\cC}
\end{equation}
where an Einstein summation over $\nu=1, \dots, n$ is implicit. 
Further recalling that on the training set we have
\begin{align}
    \langle it_{\mu} \rangle &= \sum_{\nu}\tilde{K}^{-1}_{\mu \nu} (g_{\nu} - \Delta g_{\nu}) \\
    \langle it_{\mu}it_{\nu} \rangle - \langle it_{\mu}\rangle \langle it_{\nu} \rangle &= - \tilde{K}^{-1}_{\mu,\nu}
\end{align}
where here $\Delta g$ is the full quantity without any SP approximations
\begin{equation}
\label{appEq: Delta g no approx}
    \Delta g_{\nu}=\sum_{r=3}^{\infty}\frac{1}{\left(r-1\right)!}\sum_{\mu_{1},\dots,\mu_{r-1}=1}^{n}\kappa_{\nu,\mu_{1},\dots,\mu_{r-1}}it_{\mu_{1}}\cdots it_{\mu_{r-1}}
\end{equation}
We obtain 
\begin{align}
\label{appEq: SP posterior cov test exact}    
    \Sigma_{**} &= K_{**} + \left\langle \partial_{it_{*}}^{2}\tilde{\cC}\left(it_{1},\dots,it_{n},it_{*}\right)|_{t_{*}=0}\right\rangle + \left\langle \left(\kappa_{*\nu}it_{\nu} + \Delta g_{*}\right)^{2}\right\rangle - \left\langle \kappa_{*\nu}it_{\nu} + \Delta g_{*}\right\rangle ^{2} 
    % \\ \label{appEq: SP posterior cov test short} &= 
    % K_{**}+\kappa_{*\mu}\kappa_{*\nu}\left(\left\langle it_{\mu}it_{\nu}\right\rangle -\left\langle it_{\mu}\right\rangle \left\langle it_{\nu}\right\rangle \right)+\left\langle \partial_{it_{*}}^{2}\tilde{\cC}\left(it_{1},\dots,it_{n},it_{*}\right)|_{t_{*}=0}\right\rangle 
\end{align}
where we can unpack the last two terms as 
\begin{align}
\label{appEq: SP posterior cov test unpack}
    & \left\langle \left(\kappa_{*\nu}it_{\nu}+\Delta g_{*}\right)^{2}\right\rangle -\left\langle \kappa_{*\nu}it_{\nu}+\Delta g_{*}\right\rangle ^{2} \\ &= 
    \kappa_{*\mu}\kappa_{*\nu}\left(\left\langle it_{\mu}it_{\nu}\right\rangle -\left\langle it_{\mu}\right\rangle \left\langle it_{\nu}\right\rangle \right)+2\left(\left\langle \Delta g_{*}\kappa_{*\mu}it_{\mu}\right\rangle -\left\langle \Delta g_{*}\right\rangle \left\langle \kappa_{*\mu}it_{\mu}\right\rangle \right)+\left(\left\langle \Delta g_{*}^{2}\right\rangle -\left\langle \Delta g_{*}\right\rangle ^{2}\right) \nonumber
\end{align}
One can verify that for the GP case where $\tilde{\cC}=0$, Eq. \ref{appEq: SP posterior cov test exact} simplifies to 
\begin{equation}
\label{appEq: SP posterior test cov GP}
    \Sigma_{**} = K_{**}-K_{\mu}^{*}\tilde{K}_{\mu,\nu}^{-1}K_{\nu}^{*}
\end{equation}
which is the standard posterior covariance of a GP \cite{Rasmussen2005}. 

The expressions in Eqs. \ref{appEq: SP posterior cov test exact}, \ref{appEq: SP posterior cov test unpack} are exact, but to evaluate them in a more compact form we approximate the $t_\mu$ distribution as a Gaussian centered around the SP value. 

\subsubsection{Posterior covariance on the training set}
Our target shift approach at the saddle-point level allows a computation of the fluctuations of $it_{\mu}$ using the standard procedure of expanding the action at the saddle point to quadratic order in $t_{\mu}$. 
Due to the saddle-point being an extremum this leads to $\cS \approx \cS_{\rm{saddle}} + \frac{1}{2}t_{\mu}A^{-1}_{\mu \nu} t_{\nu}$ and thus using the standard Gaussian integration formula, one finds that $A_{\mu \nu}$ is the covariance matrix of $it_{\mu}$. 
Performing such an expansion on the action of Eq. \ref{appEq: S of t and J GP} one finds 
\begin{align}
\label{appEq: kernel shift}
    A^{-1}_{\mu \nu} &= -\left(\sigma^{2}\delta_{\mu\nu} + K_{\mu\nu} + \Delta K_{\mu\nu}\right)  \\ \nonumber 
    \Delta K_{\mu \nu} &= \partial_{it_{\mu}}\underbrace{\partial_{it_{\nu}}\tilde{\cC}\left(it_{1},\dots,it_{n}\right)}_{\Delta g_{\nu}} 
\end{align}
where the $it_{\mu}$ on the r.h.s. are those of the saddle-point. Recalling Eq. \ref{appEq: var it delta g} we have 
\begin{align}
\label{appEq: SP posterior cov train}    
    \Sigma_{\mu\nu}=\left\langle f_{\mu}f_{\nu}\right\rangle -\left\langle f_{\mu}\right\rangle \left\langle f_{\nu}\right\rangle =-\sigma^{4}\left[\sigma^{2}I+K+\Delta K\right]_{\mu\nu}^{-1}+\sigma^{2}\delta_{\mu\nu}
\end{align}
and the r.h.s. coincides with the posterior covariance of a GP with a kernel equal to $K+\Delta K$.

%%%%%%%%%%%%%%%%%%%%%%%%%%%%%%%%%%%%%%%%%%%%%%%%%%%%%%%%%%%%
%%%%%%%%%%%%%%%%%%%%%%%%%%%%%%%%%%%%%%%%%%%%%%%%%%%%%%%%%%%%

\subsection{A criterion for the saddle-point regime}
\label{appendix: heuristic SP criterion}
Saddle point approximations are commonly used in statistics \cite{Daniels1954} and physics and often rely on having partition functions of the form $Z = \int dt e^{-n\cS(t)}$ where $n$ is a large number and $\cS$ is order $1$ ($O(1)$). 
In our settings we cannot simply extract such a large factor from the action and make it $O(1)$. Nonetheless, we argue that expanding the action to quadratic order around the saddle point is still a good approximation at  large $n$, with $n$ being the training set size. Concretely we give the following two consistency criteria based on comparing the saddle point results with their leading order beyond-saddle-point corrections. The first is given by the latter correction to the mean predictor over the scale of the saddle point prediction 
\begin{align}
\label{appEq: criterion 1}
    \frac{1}{2} \left[\partial_{\frac{\hat{\delta} g_{\nu}}{\sigma^2}}\partial_{\frac{\hat{\delta} g_{\eta}}{\sigma^2}} \Delta g_{\mu}\right] \tilde{K}^{-1}_{\mu_0 \mu}\tilde{K}^{-1}_{\nu \eta} \ll O(g)
\end{align}
where an Einstein summation over the training-set is implicit and the derivatives are evaluated at the saddle point value. 
This criterion can be calculated for any specific model to verify the appropriateness of the saddle point approach. We further provide a simpler criterion
\begin{align}
\label{appEq: criterion 2}
    n\left(\frac{\hat{\delta}g}{\sigma^{2}}\right)^{2}\gg1
\end{align}
which however relies on heuristic assumptions. The main purpose of this heuristic criterion is to provide a qualitative explanation for why we expect the first criterion to be small in many interesting large $n$ settings. 

To this end we first obtain the leading (beyond quadratic) correction to the mean. Consider the partition function in terms of $it$ and its expansion around the saddle point. 
As $P_0[f]$ is effectively bounded (by the Gaussian tails of the finite set of weights), the corresponding characteristic function ($e^{\cC(t_1,\dots,t_n)}$) is well defined over the entire complex plane. 
Given this, one can deform the integration contour, along each dimension ($\int_{-\infty}^{\infty} dt_{\mu}$), which originally laid on the real axis, to 
$\int_{-\infty+t_{\rm{SP}, \mu}}^{\infty+t_{\rm{SP}, \mu}} d t_{\mu}$  where $ t_{\rm{SP}}$ is purely imaginary and equals $-i\hat{\delta} g/\sigma^2$ so it crosses the saddle point (see also Ref. \cite{Daniels1954}). 
Next we expand the action in the deviation from the SP value: $\delta t_{\mu} = t_{\mu} - t_{\rm{SP}, \mu}$ to obtain
\begin{align}
    Z=\int_{-\infty}^{\infty}\delta t_{1}\cdots\delta t_{n}\exp\left(-\cS_{0}-\frac{1}{2!}\delta t_{\mu}\cS_{\mu\nu}\delta t_{\nu}-\frac{1}{3!}\cS_{\mu\nu\eta}\delta t_{\mu}\delta t_{\nu}\delta t_{\eta}+O\left(\delta t^{4}\right)\right)
\end{align}
where an Einstein summation over the training set is implicit and where we denoted for $m\geq3$
\begin{align}
    \cS_{\mu_{1}\dots\mu_{m}} \equiv 
    \eval{\partial_{t_{\mu_{1}}}\cdots\partial_{t_{\mu_{m}}}\cS}_{\vec{t}=\vec{t}_{\rm{SP}}} =
    -\eval{\partial_{t_{\mu_{1}}}\cdots\partial_{t_{\mu_{m}}}\tilde{\cC}}_{\vec{t}=\vec{t}_{\rm{SP}}} = 
    i \eval{\partial_{t_{\mu_{1}}}\cdots\partial_{t_{\mu_{m-1}}}\Delta g_{\mu_m}}_{\vec{t}=\vec{t}_{\rm{SP}}}
\end{align}

Next we consider first order perturbation theory in the cubic term and calculate the correction to the mean of $it_{\mu}$ or equivalently $\hat{\delta}g_{\mu} / \sigma^2$.
\begin{align}
\label{AppEq: Criteria}
    \langle i t_{\mu_0}\rangle &\approx \langle i t_{\mu_0} \rangle_{\rm{SP}} - \frac{1}{6} \cS_{\mu \nu \eta} \langle i \delta t_{\mu_0} \delta t_{\mu} \delta t_{\nu} \delta t_{\eta}\rangle_{\rm{SP, connected}} 
    \\ \nonumber &= 
    \frac{\hat{\delta} g_{\mu_0}}{\sigma^2}-\frac{i}{2} \cS_{\mu \nu \eta}\tilde{K}^{-1}_{\mu_0 \mu}\tilde{K}^{-1}_{\nu \eta} 
    \\ \nonumber &= 
    \frac{\hat{\delta} g_{\mu_0}}{\sigma^2} + 
    \frac{1}{2}\partial_{it_{\nu}}\partial_{it_{\eta}} \Delta g_{\mu}|_{\vec{t}=\vec{t}_{\rm{SP}}}
    \tilde{K}^{-1}_{\mu_0 \mu}\tilde{K}^{-1}_{\nu \eta}
\end{align}
where here the kernel is shifted: $\tilde{K} = \sigma^2 I + K + \Delta K$, as in Eq. \ref{appEq: kernel shift} and 
$\langle ... \rangle_{\rm{SP, connected}}$ 
means keeping terms in Wick's theorem which connect the operator being averaged ($i\delta t_{\mu_0}$) with the perturbation, as standard in perturbation theory. Comparing the last term on the right hand side (i.e. the correction) with the predictions which are $O(g)$ gives the first criterion, Eq. \ref{appEq: criterion 1}. 
Depending on context, it may be more appropriate to compare this term with the discrepancy rather than the prediction. 

Next we turn to study the scaling of this correction with $n$. 
To this end we first consider a single derivative of $\Delta g$ ($\partial_{it_{\nu}}\Delta g_{\mu}$). 
Note that $\Delta g_{\mu}$, by its definition, includes contributions from at least $n^3$ different $it$'s. 
In many cases, one expects that the value of this sum will be dominated by some finite fraction of the training set rather than by a vanishing fraction. 
This assumption is in fact implicit in our EK treatment where we replaced all $\sum_{\mu}$ with $n \int$. 
Given so, the derivative $\pa_{it_{\nu}} \Delta g_\mu$, which can be viewed as the sensitivity to changing $it_{\nu}$, is expected to go as one over the size of that fraction of the training set, namely as $1/n$. 
Under this collectivity assumption we expect the scaling 
\begin{align}
    \partial_{it_{\nu}} \Delta g_{\mu} = O(it^{-1} \Delta g n^{-1})
\end{align}
Making a similar collectivity assumption on higher derivatives yields 
\begin{align}
    \partial_{it_{\eta}} \partial_{it_{\nu}}\Delta g_\mu = O(it^{-2} \Delta g n^{-2})
\end{align}

Following this we count powers of $n$ in Eq. \ref{AppEq: Criteria} and find a $n^{-2}$ contribution from the second derivative of $\Delta g$ and a contribution from the summation over  $\sum_{\eta \nu}$. 
Despite containing two summations, we argue that the latter is in fact order $n$. 
To this end consider $n^2 \partial_{\nu} \partial_{\eta}\Delta g_{\mu}$ for fixed $\nu,\eta$, as an effective target function ($G_{\mu}(\nu,\eta)$) where we multiplied by the scaling of the second derivative to make $G$ order 1. 
The above summation appears then as $\sum_{\eta,\nu}\tilde{K}^{-1}_{\eta \nu} G_{\mu}(\nu,\eta)$. 
Next we recall that $i t_{\mu_0} = \tilde{K}^{-1}_{\mu_0 \mu} g_{\mu} = \hat{\delta}g_{\mu_0}/\sigma^2$, and so multiplication of a vector with $\tilde{K}^{-1}$ can be interpreted as the discrepancy w.r.t. the $G_{\mu}$ target. 
Accordingly the above summation over $\mu$ can be viewed as performing GP Regression on $G_{\mu}(\nu,\eta)$ leading to train discrepancy ($i t_{\mu}[G(\nu,\eta)]$) which is order $G_{\mu}(\nu,\eta)$ and hence order 1. 
The remaining summation has now a summand of the order $1$ and hence is $O(n)$ or smaller. 
We thus find that the correction to the saddle point scales as 
\begin{align}
    \langle i t_{\mu_0}\rangle - \langle i t_{\mu_0} \rangle_{\rm{SP}} &= O\left(\frac{\Delta g}{n (\hat{\delta}g/\sigma^2)^2}\right)
\end{align}
Generally we expect $\Delta g \approx 0$ at strong over-parameterization (as non-linear effects are suppressed by $C^{-1}$) and $\Delta g \sim O(g)$ at good performances (as this implies good performance on the training set). Thus we generally expect $\Delta g = O(g)=O(1)$ and hence large $n (\hat{\delta}g/\sigma^2)^2$ controls the magnitude of the corrections. Considering the $\sigma^2 \rightarrow 0$ limit, we note in passing that $\hat{\delta}g/\sigma^2$ typically remains finite. For instance it is simply $K^{-1} g$ for a Gaussian Process. 

Considering the linear CNN model of the main text, we estimate the above heuristic criterion for $n=650$ and $C=8$ where $\Delta g = O(g)$ and $\hat{\delta}g \approx 0.1 g$. This then gives $(6.5 O(g)^2)^{-1}$ as the small factor dominating the correction. As we choose $O(g)=3$ in that experiment, we find that the correction is roughly $1/60$. As the discrepancy is $0.1 g = O(0.3)$ we expect roughly a $5\%$ relative error in predicting the discrepancy.

%%%%%%%%%%%%%%%%%%%%%%%%%%%%%%%%%%%%%%%%%%%%%%%%%%%%%%%%%%%%%%%%%%%%%%%%%%%%%%%%%%%

\section{Review of the Edgeworth expansion}
\label{appendix: Edgeworth review}
In this section we give a review of the Edgeworth expansion, starting from the simplest case of a scalar valued RV and then moving on vector valued
RVs so we can write down the expansion for the output of a generic neural network on a fixed set of inputs.

\subsection{Edgeworth expansion for a scalar random variable}
\label{appendix: Edgeworth scalar RV}
Consider scalar valued continuous iid RVs $\{Z_i \}$ and assume WLOG $\left\langle Z_{i}\right\rangle =0, \hspace{5pt} \left\langle Z_{i}^{2}\right\rangle =1 $, with higher cumulants $\kappa_r^Z$ for $r \ge 3$. Now consider their normalized sum $Y_N = \frac{1}{\sqrt{N}}\sum_{i=1}^N Z_i$. 
Recall that cumulants are \textit{additive}, i.e. if $Z_1, Z_2$ are independent RVs then 
$
\kappa_r(Z_1 + Z_2) = \kappa_r(Z_1) + \kappa_r(Z_2)
$
and that the $r$-th cumulant is \textit{homogeneous} of degree $r$, i.e. if $c$ is any constant, then
$
\kappa_r(cZ) = c^r\kappa_r(Z)
$.
Combining additivity and homogeneity of cumulants we have a relation between the cumulants of $Z$ and $Y$
\begin{equation}
    \kappa_{r\ge2} := \kappa^Y_{r\ge2} = \frac{N \kappa_r^Z}{(\sqrt{N})^r} = 
    \frac{\kappa_r^Z}{N^{r/2 - 1}}    
\end{equation}
Now, let $\varphi(y):=(2\pi)^{-1/2}e^{-y^2/2}$ be the PDF of the standard normal distribution. The \textit{characteristic function} of $Y$ is given by the Fourier transform of its PDF $P(y)$ and is expressed via its cumulants
\begin{equation}
    \hat{P}(t) := \cF[P(y)] =
    \exp\left(\sum_{r=1}^\infty \kappa_r \frac{(it)^r}{r!} \right) =
    \exp\left(\sum_{r=3}^\infty \kappa_r \frac{(it)^r}{r!} \right) \hat{\varphi}(t)    
\end{equation}
where the last equality holds since $\kappa_1 = 0, \quad \kappa_2 = 1$ and
$\hat{\varphi}(t) = e^{-\frac{t^2}{2}}$. 
From the CLT, we know that $P(y) \to \varphi(y)$ as $N\to \infty$. 
Taking the inverse Fourier transform  $\cF^{-1}$ has the effect of mapping $it \mapsto -\partial_y$ thus
\begin{equation}
    P(y) = \exp\left( \sum_{r=3}^\infty \kappa_r \frac{(-\partial_y)^r}{r!} \right) \varphi(y) =
        \varphi(y)\left(1 + \sum_{r=3}^\infty \frac{\kappa_r}{r!}H_r(y) \right)    
\end{equation}
where $H_r(y)$ is the $r$th \textit{probabilist's Hermite polynomial}, defined by
\begin{equation}
\label{eq: scalar Hermite poly def}    
    H_r(y) = (-)^r e^{y^2/2} \frac{d^r}{dy^r} e^{-y^2/2}
\end{equation}
e.g. $H_4(y) = y^4 - 6y^2 + 3$.

\subsection{Edgeworth expansion for a vector valued random variable}
\label{appendix: Edgeworth vector RV}
Consider now the analogous procedure for vector-valued RVs in $\bbR^n$ (see \cite{Mccullagh2017}). We perform an Edgeworth expansion around a centered multivariate Gaussian distribution with covariance matrix $\kappa^{i,j}$
\begin{equation}
    \varphi(\vec{y}) = \frac{1}{(2\pi)^{d/2} \det(\kappa^{i,j})} \exp \left( -\frac{1}{2} \kappa_{i,j} y^i y^j \right)
\end{equation}
where $\kappa_{i,j}$ is the matrix inverse of $\kappa^{i,j}$ and Einstein summation is used. 
The $r$'th order cumulant becomes a tensor with $r$ indices, e.g. the analogue of $\kappa_4$ is $\kappa^{i,j,k,l}$. The Hermite polynomials are now multi-variate polynomials, so that the first one is 
$H_i = \kappa_{i,j} y^j$ and the fourth one is
\begin{equation} \label{eq: 4th Hermite multivar}
\begin{split}
    H_{ijkl}(\vec{y})
    & = 
    e^{\frac{1}{2} \kappa_{i',j'} y^{i'} y^{j'}} \pa_i \pa_j \pa_k \pa_l e^{-\frac{1}{2} \kappa_{i',j'} y^{i'} y^{j'}}
    \\ & = 
    H_i H_j H_k H_l - H_i H_j \kappa_{k,l} [6] + \kappa_{i,j} \kappa_{k,l} [3]
\end{split}
\end{equation}
where the postscript bracket notation is simply a convenience to avoid listing explicitly all possible partitions of the indices, e.g.
$
\kappa_{i,j} \kappa_{k,l} [3] = 
\kappa_{i,j} \kappa_{k,l} + \kappa_{i,k} \kappa_{j,l}
+ \kappa_{i,l} \kappa_{j,k}
$

In our context we are interested in even distributions where all odd cumulants vanish, so the Edgeworth expansion reads
\begin{equation}
\label{eq: multivar Edgeworth}
    P(\vec{y}) = \exp\left(\frac{\kappa^{i,j,k,l}}{4!}\pa_{i}\pa_{j}\pa_{k}\pa_{l} + \dots\right)\varphi(\vec{y}) = \varphi(\vec{y})\left(1+\frac{\kappa^{i,j,k,l}}{4!}H_{ijkl}+\dots\right)
\end{equation}

\subsection{Edgeworth expansion for the posterior of Bayesian neural network}
\label{appendix: Edgeworth posterior BNN}
Consider an on-data formulation, i.e. a distribution over a vector space - the NN output evaluated on the training set and on a single test point, rather than a distribution over the whole function space:
\begin{equation}
    f\left(\x\right)\to\vec{f}\equiv\left(f\left(\x_{1}\right),\dots,f\left(\x_{n}\right),f\left(\x_{n+1}\right)\right)\in\bbR^{n+1}\qquad \x_{n+1} = \x_{*}
\end{equation}
where $\x_{*}$ is the test point. 
Let $\kappa_{r}$ denote the $r$th cumulant of the prior $P_{0}\left(\vec{f}\right)$ of the network over this space:
\begin{equation}
    \left[\kappa_{r}\right]_{\mu_{1},...,\mu_{r}}=\left\langle f\left(\x_{\mu_{1}}\right),\dots,f\left(\x_{\mu_{r}}\right)\right\rangle -"\mathrm{disconnected\,averages}" \qquad \mu\in\left\{ 1,\dots,n+1\right\} 
\end{equation}
Take the baseline distribution to be Gaussian $P_{G}\left(\vec{f}\right)\propto\exp\left(-\frac{1}{2}\vec{f}^{\transpose}K^{-1}\vec{f}\right)$, around which we perform the Edgeworth expansion, thus the characteristic function of the prior reads
\begin{equation}
    \hat{P}_{0}\left(\vec{t}\right)=\exp\left(\sum_{r=4}^{\infty}\frac{\kappa_{r}\left(i\vec{t}\right)^{r}}{r!}\right)\hat{P}_{G}\left(\vec{t}\right)
\end{equation}
and thus 
\begin{equation}
    P_{0}\left(\vec{f}\right)=\exp\left(\sum_{r=4}^{\infty}\frac{\left(-\right)^{r}\kappa_{r}\vec{\pa}^{r}}{r!}\right)P_{G}\left(\vec{f}\right)
\end{equation}
where we used the shorthand notation:
\begin{equation}
    \kappa_{r}\vec{\pa}^{r}\equiv\sum_{\mu_{1},...,\mu_{r}}\left[\kappa_{r}\right]_{\mu_{1},...,\mu_{r}}\pa_{f_{\mu_{1}}}\cdots\pa_{f_{\mu_{r}}}
\end{equation}
and the indices range over both the train set and the test point
$\mu\in\left\{ 1,\dots,\underbrace{n+1}_{*}\right\}$. 
In our case, all odd cumulants vanish, thus 
\begin{equation}
    \exp\left(\sum_{r=4}^{\infty}\frac{\left(-\right)^{r}\kappa_{r}\vec{\pa}^{r}}{r!}\right)=\exp\left(\sum_{r=4}^{\infty}\frac{\kappa_{r}\vec{\pa}^{r}}{r!}\right)
\end{equation}
Introducing the data term and a source term, the partition function reads (denote
$f\left(\x_{\mu}\right)\equiv f_{\mu}, \quad f\left(\x_{*}\right)\equiv f_{*})$
\begin{equation}
\label{appEq: Z Edgeworth}
    Z\left(J\right)=\int d\vec{f}\left(\exp\left(\sum_{r=4}^{\infty}\frac{\kappa_{r}\vec{\pa}^{r}}{r!}\right)P_{G}\left(\vec{f}\right)\right)\exp\left(-\frac{1}{2\sigma^{2}}\sum_{\mu=1}^{n}\left(g_{\mu}-f_{\mu}\right)^{2} + \sum_{\mu=1}^{n+1} J_\mu f_{\mu} \right)
\end{equation}

%%%%%%%%%%%%%%%%%%%%%%%%%%%%%%%%%%%%%%%%%%%%%%%%%%%%%%%%%%%%
%%%%%%%%%%%%%%%%%%%%%%%%%%%%%%%%%%%%%%%%%%%%%%%%%%%%%%%%%%%%

\section{Target shift equations - alternative derivation}
\label{appendix: TS derivs}

Here we derive our self-consistent target shift equations from a different approach which does not require the introduction of the $it_{\mu}$ integration variables by transforming to Fourier space. 
While this approach requires an additional assumption (see below) it also has the benefit of being extendable to any smooth loss function comprised of a sum over training points. In particular, below we derive it for both MSE loss and cross entropy loss. 

To this end, we examine the Edgeworth expansion for the partition function given by Eq. \ref{appEq: Z Edgeworth}. 
By using a series of integration by parts and noting the boundary terms vanish, one can shift the action of the higher cumulants from the prior to the data dependent term
\begin{equation}
\label{appEq: Z int by parts}
    Z\left(\vec{J}\right) = \int d\vec{f}P_{G}\left(\vec{f}\right)\left[\exp\left(\sum_{r=3}^{\infty}\frac{1}{r!} \sum_{\mu_{1},...,\mu_{r}=1}^{n+1} \kappa_{\mu_{1},...,\mu_{r}}\pa_{f_{\mu_{1}}}\cdots\pa_{f_{\mu_{r}}}\right) \exp\left(-\frac{1}{2\sigma^{2}}\sum_{\mu=1}^{n}\left(g_{\mu} - f_{\mu}\right)^{2} + \sum_{\mu=1}^{n+1}J_{\mu}f_{\mu}\right)\right]
\end{equation}
Doing so yields an equivalent viewpoint on the problem, wherein the Gaussian data term and the non-Gaussian prior appearing in Eq. \ref{appEq: Z Edgeworth} are replaced in Eq. \ref{appEq: Z int by parts} by a Gaussian prior and a non-Gaussian data term. 

Next we argue that in the large $n$ limit, the non-Gaussian data-term can be expressed as a Gaussian-data term but on a shifted target. To this end we note that when $n$ is large, most combinations of derivatives in the exponents act on different data points. In such cases derivatives could simply be replaced as  $\partial_{\mu_i} \rightarrow \sigma^{-2} \hat{\delta}g_{\mu_i}$, where $\hat{\delta}g_{\mu_i} \equiv g_{\mu_i} - f_{\mu_i}$ denotes the discrepancy on the training point $\mu_i$. 

Consider next how $f_{\nu}$ on a particular training point ($\nu$) is affect by these derivative terms. Following the above observation, most terms in the exponent will not act on $f_{\nu}$ and a $1/n$ portion will contain a single derivative. The remaining rarer cases, where two derivatives act on the same $\nu$, are neglected. For each $f_{\nu}$ we thus replace $r-1$ derivatives in the order $r$ term in \ref{appEq: Z int by parts} by discrepancies, leaving a single derivative operator that is multiplied by the following quantity
\begin{equation}
\label{eq: Delta g def}
    \Delta g_{\nu}\equiv \sum_{r=3}^{\infty} \frac{1}{\left(r-1\right)!} \sum_{\mu_{1},...,\mu_{r-1}}^{n} \kappa_{\nu, \mu_{1}\dots\mu_{r-1}}(\sigma^{-2}\hat{\delta}g_{\mu_{1}})\cdots(\sigma^{-2}\hat{\delta}g_{\mu_{r-1}})
\end{equation}
Note that the summation indices span only the training set, not the test point: $\mu_{1},...,\mu_{r-1} \in \left\{ 1,\dots,n\right\} $, whereas the free index spans also the test point $\nu \in \left\{ 1,\dots, n+1 \right\} $.

Recall that an exponentiated derivative operator acts as a shifting operator, e.g. for some constant $a\in\bbR$, any smooth scalar function $\varphi$ obeys
$e^{a\pa_{x}}\varphi\left(x\right)=\varphi\left(x+a\right)$. 
If this $\Delta g$ was a constant, the differential operator could now readily act on the data term. 
Next we make again our collectivity assumption: as $\Delta g$ involves a sum over many data-points, it will be a weakly fluctuating quantity in the large $n$ limit provided the contribution to $\Delta g$ comes from a collective effect rather than by a few data points. We thus perform our second approximation, of the mean-field type, and replace $\Delta g$ by its average $\overline{\Delta g}$, leading to 
\begin{equation}
\label{eq: Z(J) Delta g}
    Z\left(\vec{J}; \overline{\Delta g} \right)= \int d\vec{f}P_{G}\left(\vec{f}\right)
    \exp \left(
    -\frac{1}{2\sigma^{2}}\sum_{\mu=1}^{n}\left(g_{\mu} - \overline{\Delta g}_{\mu} - f_{\mu}\right)^{2} 
    + \sum_{\mu=1}^{n+1} J_\mu \left(f_{\mu} + \overline{\Delta g}_{\mu} \right)
    \right)
\end{equation}
Given a fixed $\overline{\Delta g}$, \ref{eq: Z(J) Delta g} is the partition function corresponding to a GP with the train targets shifted by $\overline{\Delta g}_\mu$ and the test target shifted by $\overline{\Delta g}_*$. 
Following this we find that $\overline{\Delta g}$ depends on the discrepancy of the GP prediction which in turn depends on $\overline{\Delta g}$. In other words we obtain our self-consistent equation: $\overline{\Delta g} = \langle \Delta g_{\mu}\rangle_{Z\left(\vec{J};\overline{\Delta g}\right)}$. 

The partition function \ref{eq: Z(J) Delta g} reflects the correspondence between finite DNNs and a GP with its target shifted by $\overline{\Delta g}$. 
To facilitate the analytic solution of this self-consistent equation, we focus on the case
$\left\langle \hat{\delta}g_{\mu}\hat{\delta}g_{\nu}\right\rangle \ll\left\langle \hat{\delta}g_{\mu}\right\rangle \left\langle \hat{\delta}g_{\nu}\right\rangle$
at least for $\mu \neq \nu$. We note that this was the case for the two toy models we studied.

Given this, the expectation value over $\Delta g$ using the GP defined by $Z\left(\vec{J};\overline{\Delta g}\right)$, which consists of products of expectation values of individual discrepancies and correlations between two discrepancies, can then be expressed using only the former. Omitting correlations within the GP expectation value, one obtains a simplified self-consistent equation involving only the average discrepancies: 
\begin{equation}
\label{eq: delta g self consistent}
    \forall\mu\in\left\{ 1,\dots,n\right\} :\qquad 
    \langle \hat{\delta}g_{\mu} \rangle_{Z\left(\vec{J};\overline{\Delta g}\right)}
     \equiv g_{\mu} - \overline{\Delta g}_{\mu} -  \left\langle f_{\mu}\right\rangle _{Z\left(\vec{J};\overline{\Delta g}\right)} = g_{\mu} - \overline{\Delta g}_{\mu} - \sum_{\nu, \rho = 1}^{n} K_{\mu\nu}\tilde{K}_{\nu\rho}^{-1}\left(g_{\rho} - \overline{\Delta g}_{\rho}\right) 
\end{equation}
with $\hat{\delta}g_{\mu}$ now understood as a number, also within \ref{eq: Delta g def}. 
Lastly, we plug the solution to these equations to find the prediction on the test point:
$\left\langle f(\x_*)\right\rangle _{Z\left(\vec{J};\overline{\Delta g}\right)}$. These coincide with the self-consistent equations derived via the saddle point approximation in the main text. 

Notably the above derivation did not hinge on having MSE loss. For any loss given as a sum over training points, $\cL = \sum^n_{\mu} L_{\mu}(f_{\mu})$, the above derivation should hold with $\sigma^{-2} \hat{\delta} g_{\mu}$ in $\Delta g_{\nu}$ replaced by $\partial_{f_{\mu}} L_{\mu}$. 
In particular for the cross entropy loss where $f_{\nu,i}$ is the pre-softmax output of the DNN for class $i$ we will have 
\begin{align}
\partial_{f_{\nu,i}} L_{\nu} = -\delta_{i_{\nu},i} + \frac{e^{f_{\nu,i}}}{\sum_j e^{f_{\nu,j}}}
\end{align}
where $i$ and $j$ run over all classes, $i_{\nu}$ is the class of $\x_{\nu}$. Neatly, the above r.h.s. is again a form of discrepancy but this time in probability space. Namely it is $p_{\rm{model}}(i|\x_{\nu})-p_{\rm{data}}(i|\x_{\nu})$, where $p_{\rm{model}}$ is the distribution generated by the softmax layer, and $p_{\rm{data}}$ is the empirical distribution. 
Following this one can readily derive  self-consistent equations for cross entropy loss and solve them numerically. 
Further analytical progress hinges on developing analogous of the EK approximation for cross entropy loss.

%%%%%%%%%%%%%%%%%%%%%%%%%%%%%%%%%%%%%%%%%%%%%%%%%%%%%%%%%%%%
%%%%%%%%%%%%%%%%%%%%%%%%%%%%%%%%%%%%%%%%%%%%%%%%%%%%%%%%%%%%

\section{Review of the Equivalent Kernel (EK)}
\label{appendix: EK review}

In this appendix we generally follow \cite{Rasmussen2005}, see also \cite{sollich2004understanding} for more details. 
The posterior mean for GP regression 
\begin{equation}
\label{appEq: GP mean and var}
    \bar{f}_{\mathrm{GP}}(\x_*) = \sum_{\mu, \nu} K^*_{\mu} \tilde{K}^{-1}_{\mu\nu} y_\nu
\end{equation}
can be obtained as the function which minimizes the functional
\begin{equation}
\label{appEq: J[f] discrete}
    J\left[f\right]=\frac{1}{2\sigma^{2}}\sum_{\alpha=1}^{n}\left(y_{\alpha}-f\left(\x_{\alpha}\right)\right)^{2}+\frac{1}{2}  ||f||_{\cH}^{2}
\end{equation}
where $||f||_{\cH}$ is the RKHS norm corresponding to kernel $K$. 
Our goal is now to understand the behaviour of the minimizer of $J[f]$ as $n \to \infty$.
Let the data pairs $\left(\x_{\alpha},y_{\alpha}\right)$ be drawn from the probability measure $\mu(\x, y)$. 
The expectation value of the MSE is 
\begin{equation}
\label{eq: MSE expec val}    
   \bbE\left[\sum_{\alpha=1}^{n}\left(y_{\alpha}-f\left(\x_{\alpha}\right)\right)^{2}\right]=n\int\left(y-f\left(\x\right)\right)^{2}d\mu\left(\x,y\right)
\end{equation}

Let $g\left(\x\right)\equiv\bbE\left[y|\x\right]$ be the ground truth regression function to be learned. The variance around $g\left(\x\right)$ is denoted 
$\sigma^{2}\left(\x\right)=\int\left(y-g\left(\x\right)\right)^{2}d\mu\left(y|\x\right)$.
Then writing $y-f=\left(y-g\right)+\left(g-f\right)$ we find that the MSE on the data target $y$ can be broken up into the MSE on the ground truth target $g$ plus variance due to the noise
\begin{equation}
\label{eq: MSE split}
    \int\left(y-f\left(\x\right)\right)^{2}d\mu\left(\x,y\right)=\int\left(g\left(\x\right)-f\left(\x\right)\right)^{2}d\mu\left(\x\right)+\int\sigma^{2}\left(\x\right)d\mu\left(\x\right)
\end{equation}
Since the right term on the RHS of \ref{eq: MSE split} does not depend on $f$ we can ignore it when looking for the minimizer of the functional which is now replaced by 
\begin{equation}
\label{eq: J[f] contin}
    J_{\mu}\left[f\right]=\frac{n}{2\sigma^{2}}\int\left(g\left(\x\right)-f\left(\x\right)\right)^{2}d\mu\left(\x\right)+\frac{1}{2}
    ||f||_{\cH}^{2}
\end{equation}
To proceed we project $g$ and $f$ on the eigenfunctions of the kernel with respect to $\mu(\x)$ which obey 
$\int\mu\left(\x'\right)K\left(\x,\x'\right)\psi_{s}\left(\x'\right)=\lambda_{s}\psi_{s}\left(\x\right)$. 
Assuming that the kernel is non-degenerate so that the $\psi$'s form a complete orthonormal basis, for a sufficiently well behaved target we may write $g\left(\x\right)=\sum_{s}g_{s}\psi_{s}\left(\x\right)$ where 
$g_{s}=\int g\left(\x\right)\psi_{s}\left(\x\right)d\mu\left(\x\right)$, 
and similarly for $f$. 
Thus the functional becomes
\begin{equation}
\label{eq: J[f] contin eigenbasis}    
    J_{\mu}\left[f\right]=\frac{n}{2\sigma^{2}}\sum_{s}\left(g_{s}-f_{s}\right)^{2}+\frac{1}{2}\sum_{s}\frac{f_{s}^{2}}{\lambda_{s}}
\end{equation}
This is easily minimized by taking the derivative w.r.t. each $f_s$ to yield
\begin{equation}
\label{eq: f_s sol}
    f_{s} = \frac{\lambda_{s}}{\lambda_{s}+\sigma^{2}/n}g_{s}
\end{equation}
In the limit $n \to \infty$ we have $\sigma^2/n \to 0$ thus we expect that $f$ would converge to $g$.
The rate of this convergence will depend on the smoothness of $g$, the kernel $K$ and the measure $\mu(\x,y)$. 
From \ref{eq: f_s sol} we see that if $n\lambda_s \ll \sigma^2$ then $f_s$ is effectively zero. 
This means that we cannot obtain information about the coefficients of eigenfunctions with small eigenvalues until we get a sufficient amount of data.
Plugging the result \ref{eq: f_s sol} into
$f\left(\x\right) = \sum_{s}f_{s}\psi_{s}\left(\x\right)$ and recalling
$g_{s}=\int g\left(\x'\right)\psi_{s}\left(\x'\right)d\mu\left(\x'\right)$
we find
\begin{equation}
\label{appEq: f EK main res}
    \left\langle f\left(\x\right)\right\rangle _{\rm{EK}}=\sum_{s}\frac{\lambda_{s}g_{s}}{\lambda_{s}+\sigma^{2}/n}\psi_{s}\left(\x\right)=\int\underbrace{\sum_{s}\frac{\lambda_{s}\psi_{s}\left(\x\right)\psi_{s}\left(\x'\right)}{\lambda_{s}+\sigma^{2}/n}}_{h\left(\x,\x'\right)}g\left(\x'\right)d\mu\left(\x'\right)
\end{equation}
The term $h(\x,\x')$ it the \textit{equivalent kernel}.
Notice the similarity to the vector-valued equivalent kernel weight function 
$\mathbf{h}\left(\x_{*}\right)=\left(\mathbf{K}+\sigma^{2}I\right)^{-1}\mathbf{k}\left(\x_{*}\right)$
where $\mathbf{K}$ denotes the $n\times n$ matrix of covariances between the training points with entries $K\left(\x_{\mu},\x_{\nu}\right)$ and $\mathbf{k}\left(\x_{\ast}\right)$ is the vector of covariances with elements $K\left(\x_{\mu},\x_{\text{\textasteriskcentered}}\right)$. 
The difference is that in the usual discrete formulation the prediction was obtained as a linear combination of a finite number of observations
$y_i$ with weights given by $h_i(\x)$ while here we have instead a continuous integral.

%%%%%%%%%%%%%%%%%%%%%%%%%%%%%%%%%%%%%%%%%%%%%%%%%%%%%%%%%%%%
%%%%%%%%%%%%%%%%%%%%%%%%%%%%%%%%%%%%%%%%%%%%%%%%%%%%%%%%%%%%

\section{Additional technical details for solving the self consistent equations}
\label{appendix: technical SC eqs}

\subsection{EK limit for the CNN toy model}
\label{appendix: technical SC eqs EK limit}
In this subsection we show how to arrive at Eq. \ref{eq: SC alpha all kappas no SL} from the main text, which is a self consistent equation for the proportionality constant, $\alpha$, defined by $\hat{\delta}g = \alpha g$. 
We first show that both the shift and the discrepancy are linear in the target, and then derive the equation. 

\subsubsection{The shift and the discrepancy are linear in the target}
Recall that we assume a linear target with a single channel:
\begin{align}
    g\left(\x\right)=\sum_{k=1}^{N}a_{k}^{*}\left(\w^{*}\cdot\tilde{\x}_{k}\right)
\end{align}
A useful relation in our context is 
\begin{align}
    \int d\mu\left(\x^{2}\right)\left(\tilde{\x}_{i}^{1}\cdot\tilde{\x}_{j}^{2}\right)g\left(\x^{2}\right) 
    &=
    \int d\mu\left(\x^{2}\right)\left(\tilde{\x}_{i}^{1}\cdot\tilde{\x}_{j}^{2}\right)\sum_{k=1}^{N}a_{k}^{*}\left(\w^{*}\cdot\tilde{\x}_{k}^{2}\right)
    \\ \nonumber &= 
    \left(\tilde{\x}_{i}^{1}\right)^{\transpose}\left(\sum_{k=1}^{N}a_{k}^{*}\underbrace{\int d\mu\left(\x^{2}\right)\tilde{\x}_{j}^{2}\left(\tilde{\x}_{k}^{2}\right)^{\transpose}}_{\delta_{jk}I_{S}}\right)\w^{*}
    \\ \nonumber &= 
    a_{j}^{*}\left(\tilde{\x}_{i}^{1}\cdot\w^{*}\right)
\end{align}
The fact that $f$ is always a linear function of the input (since the CNN linear) and the fact that it is proportional to $g$ at $C\to\infty$ (since the GP is linear in the target), motivates the ansatz:
\begin{align}
    \hat{\delta}g \equiv g - f = \alpha g
\end{align}
Indeed we will show that this ansatz provides a solution to the non linear self consistent equations.

Notice that the target shift has a form of a geometric series. 
In the linear CNN toy model we are able to sum this entire series, whose first term is related to (using the notation introduced in §\ref{appendix: cumulants linear CNN}): 
\begin{align}
    & \int d\mu\left(\x^{2}\right)d\mu\left(\x^{3}\right)d\mu\left(\x^{4}\right)\kappa_{4}\left(\x^{1},\x^{2},\x^{3},\x^{4}\right)g\left(\x^{2}\right)g\left(\x^{3}\right)g\left(\x^{4}\right)
    \\ \nonumber &= 
    \frac{\lambda^{2}}{C}\int d\mu_{2:4}\sum_{i,j=1}^{N}\left\{ \left(1_{i}3_{j}\right)\left[\left(2_{i}4_{j}\right)+\left(4_{i}2_{j}\right)\right]+\left(1_{i}4_{j}\right)\left[\left(2_{i}3_{j}\right)+\left(3_{i}2_{j}\right)\right]+\left(1_{i}2_{j}\right)\left[\left(3_{i}4_{j}\right)+\left(4_{i}3_{j}\right)\right]\right\} g\left(\x^{2}\right)g\left(\x^{3}\right)g\left(\x^{4}\right)
    \\ \nonumber &= 
    \frac{\lambda^{2}}{C}\sum_{i,j=1}^{N}\left\{ 2a_{i}^{*}\left(a_{j}^{*}\right)^{2}\norm{\w^{*}}^{2}\left(\tilde{\x}_{i}^{1}\cdot\w^{*}\right)+2a_{i}^{*}\left(a_{j}^{*}\right)^{2}\norm{\w^{*}}^{2}\left(\tilde{\x}_{i}^{1}\cdot\w^{*}\right)+2a_{i}^{*}\left(a_{j}^{*}\right)^{2}\norm{\w^{*}}^{2}\left(\tilde{\x}_{i}^{1}\cdot\w^{*}\right)\right\} 
    \\ \nonumber &= 
    \frac{6\lambda^{2}}{C}\norm{\w^{*}}^{2}\left(\sum_{j=1}^{N}\left(a_{j}^{*}\right)^{2}\right)\underbrace{\sum_{i=1}^{N}a_{i}^{*}\left(\tilde{\x}_{i}^{1}\cdot\w^{*}\right)}_{g\left(\x^{1}\right)}
    \\ \nonumber &= 
    \frac{6\lambda^{2}}{C}\norm{\w^{*}}^{2}\underbrace{\left(\sum_{j=1}^{N}\left(a_{j}^{*}\right)^{2}\right)}_{\approx\sigma_{a}^{2} }g\left(\x^{1}\right)
\end{align}
For simplicity we can assume $\norm{\w^{*}}^{2} = 1$ and $\sigma_a^2 = 1$, thus getting a simple proportionality constant of $\frac{6\lambda^{2}}{C}$.
If we were to trade $g$ for $\hat{\delta}g$, as we have in $\Delta g$, we would get a similar result, with an extra factor of 
$\left(\frac{\alpha}{\sigma^{2}/n}\right)^{3}$. 
The factor of $6$ will cancel out with the factor of $1/(4-1)!$ appearing in the definition of $\Delta g$. 
Repeating this calculation for the sixth cumulant, one would arrive to the same result multiplied by a factor of 
$\frac{\lambda}{C}\left(\frac{\alpha}{\sigma^{2}/n}\right)^{2}$
due to the general form of the even cumulants (Eq. \ref{appEq: kappa_2m}) and the fact that there an extra two $(\sigma^{2}/n)^{-1}\hat{\delta}g$'s.

\subsubsection{Self consistent equation in the EK limit}
Starting from the proportionality relations $\hat{\delta}g = \alpha g$ and $\Delta g=\alpha_{\Delta}g$, we can now write the self consistent equation for the discrepancy as 
\begin{align}
\label{appEq: SC EK discrep}    
    \hat{\delta}g=\left(g-\Delta g\right)-q\frac{\lambda}{\lambda+\sigma_{n}^{2}}\left(g-\Delta g\right)
\end{align}
Dividing both sides by $g$ we get a scalar equation 
\begin{align}
\label{appEq: SC EK alpha}    
    \alpha &= \left(1-\alpha_{\Delta}\right)-q\frac{\lambda}{\lambda+\sigma_{n}^{2}}\left(1-\alpha_{\Delta}\right)
    \\ \nonumber &= 
    \frac{\lambda+\sigma_{n}^{2}}{\lambda+\sigma_{n}^{2}}-q\frac{\lambda}{\lambda+\sigma_{n}^{2}}+\left(q\frac{\lambda}{\lambda+\sigma_{n}^{2}}-1\right)\alpha_{\Delta}
    \\ \nonumber &= 
    \frac{\sigma_{n}^{2}}{\lambda+\sigma_{n}^{2}}+\left(1-q\right)\frac{\lambda}{\lambda+\sigma_{n}^{2}}+\left(q\frac{\lambda}{\lambda+\sigma_{n}^{2}}-1\right)\alpha_{\Delta}
\end{align}
The factor $\alpha_{\Delta}$ can be calculated by noticing that $\Delta g$ has the form of a geometric series. 
To better understand what follows next, the reader should first go over §\ref{appendix: cumulants linear CNN}.
The first term in this series is related to contracting the fourth cumulant $\kappa_4$ with three $\hat{\delta}g$'s thus yielding a factor of 
$\frac{\lambda^{2}}{C}\left(\frac{\alpha}{\sigma^{2}/n}\right)^{3}$
(recall that in the EK approximation we trade $\sigma^2 \to \sigma^2/n$). 
The ratio of two consecutive terms in this series is given by 
$\frac{\lambda}{C}\left(\frac{\alpha}{\sigma^{2}/n}\right)^{2}$. 
Using the formula for the sum of a geometric series we have 
\begin{align}
\label{appEq: SC alpha}
    \alpha=\frac{\sigma^{2}/n}{\lambda+\sigma^{2}/n}+\frac{\left(1-q\right)\lambda}{\lambda+\sigma^{2}/n}+\left(q\frac{\lambda}{\lambda+\sigma^{2}/n}-1\right)\frac{\lambda^{2}}{C}\left(\frac{\alpha}{\sigma^{2}/n}\right)^{3}\left[1-\frac{\lambda}{C}\left(\frac{\alpha}{\sigma^{2}/n}\right)^{2}\right]^{-1}
\end{align}

\subsection{Corrections to EK and estimation of the $q_{\rm{train}}$ factor in the main text}
\label{appendix: Fudge factor}
The EK approximation can be improved systematically using the field-theory approach of Ref. \cite{Cohen2019} where the EK result is interpreted as the leading order contribution, in the large $n$ limit, to the average of the GP predictor over many data-set draws from the dataset measure. 
However, that work focused on the test performance whereas for $q_{\rm{train}}$ we require the performance on the training set. 
We briefly describe the main augmentations needed here and give the sub-leading and sub-sub-leading corrections to the EK result on the training set, enabling us to estimate $q_{\rm{train}}$ analytically within a $16.3\%$ relative error compared with the empirical value. 
Further systematic improvements are possible but are left for future work. 

We thus consider the quantity $\sum_\mu \varphi(\x_{\mu}) f(\x_{\mu})$ where $\x_{\mu}$ is drawn from the training set, $f(\x_{\mu})$ is the predictive mean of the GP on that specific training set, and $\varphi(\x_{\mu})$ is some function which we will later take to be the target function ($\varphi(\x) = g(\x)$). 
We wish to calculate the average of this quantity over all training set draws of size $n$. 
We begin by adding a source term of the form $J \sum_{\mu} \varphi(\x_{\mu}) f(\x_{\mu})$ to the action and notice a similar term appearing in the GP action ($-\sum_{\mu} (f(\x_{\mu}) - g(\x_{\mu}))^2$) due to the MSE loss.
Examining this extra term one notices that it can be absorbed as a $J$ dependent shift to the target on training set ($g(\x_{\mu}) \rightarrow g(\x_{\mu}) + \frac{J \sigma^2}{2} \varphi(\x_{\mu})$) following which the analysis of Ref. \cite{Cohen2019} carries through straightforwardly. 
Doing so, the general result for the leading EK term and sub-leading correction are 
\begin{align}
    n \int d \mu(\x) \varphi(\x) \langle f(\x) \rangle_{\rm{EK}} - \frac{n}{\sigma^2}\int d \mu(\x) \varphi(\x) \left[ \cov(\x, \x) (\langle f(\x) \rangle_{\rm{EK}}-g(\x))\right]
\end{align}
where $\cov(\x, \x) = \langle f(\x) f(\x)\rangle_{\rm{EK}}-\langle f(\x) \rangle_{\rm{EK}}\langle f(\x) \rangle_{\rm{EK}}$, $\langle ... \rangle_{\rm{EK}}$ means averaging with $Z_{\rm{EK}}$ of Ref. \cite{Cohen2019}, and $\langle f(\x) \rangle_{\rm{EK}}$ is the EK prediction of the previous section, Eq. \ref{appEq: f EK main res}. 

Turning to the specific linear CNN toy model and carrying the above expansion up to an additional term leads to 
\begin{align}
\alpha_{\rm{train}} &\approx \alpha_{\rm{EK}} \left(1-\frac{\alpha_{\rm{EK}}}{\sigma^2}+\frac{3}{4}\frac{\alpha_{\rm{EK}}^2}{\sigma^4} \right) \\ \nonumber 
\alpha_{\rm{EK}} &= \frac{\sigma^2/n}{\sigma^2/n+\lambda} = \frac{\sigma^2/n}{\sigma^2/n+(NS)^{-1}}
\end{align}
Considering for instance $n=200, \sigma^2=1.0, N=30$ and $S=30$, we find $\alpha_{\rm{EK}}=0.818$ and so
\begin{align}
\alpha_{\rm{train}} &\approx 0.559
\end{align}
recalling that $q_{\rm{train}} =\frac{\lambda+\sigma^2/n}{\lambda}(1-\alpha_{\rm{train}})$ we have 
\begin{align}
q_{\rm{train}} &\approx 2.4255
\end{align}
whereas the empirical value here is $2.8995$.

%%%%%%%%%%%%%%%%%%%%%%%%%%%%%%%%%%%%%%%%%%%%%%%%%%%%%%%%%%%%%%%%%%%%%%%%%%%%%%%%%%%%%%%%%%

\section{Cumulants for a two-layer linear CNN}
\label{appendix: cumulants linear CNN}
In this section we explicitly derive the leading (fourth and sixth) cumulants of the toy model of §\ref{section: two layer linear CNN}, and arrive at the general formula for the even cumulant of arbitrary order. 

\subsection{Fourth cumulant}
\subsubsection{Fourth cumulant for a CNN with general activation function (averaging over the readout layer)}
For a general activation, we have in our setting for a 2-layer CNN
\begin{equation}
    f\left(\x^{\mu}\right)=\sum_{i=1}^{N}\sum_{c=1}^{C}a_{i,c}\phi\left(\w_{c}\cdot\tilde{\x}_{i}^{\mu}\right)=:\sum_{i=1}^{N}\sum_{c=1}^{C}a_{i,c}\phi_{i,c}^{\mu}
\end{equation}
The kernel is 
\begin{align}
    K\left(\x^{1},\x^{2}\right) &= 
    \left\langle f\left(\x^{1}\right)f\left(\x{}^{2}\right)\right\rangle 
    \\ &= \nonumber
    \left\langle \sum_{i,i'=1}^{N}\sum_{c,c'=1}^{C}a_{i,c}\phi_{i,c}^{1}a_{i',c'}\phi_{i',c'}^{2}\right\rangle 
    \\ &= \nonumber
    \sum_{i,i'=1}^{N}\sum_{c,c'=1}^{C}\underbrace{\left\langle a_{i,c}a_{i',c'}\right\rangle _{a}}_{\delta_{ii'}\delta_{cc'}\sigma_{a}^{2}/CN}\left\langle \phi_{i,c}^{1}\phi_{i',c'}^{2}\right\rangle _{\w}
    \\ &= \nonumber
    \frac{\sigma_{a}^{2}}{CN}\sum_{i=1}^{N}\sum_{c=1}^{C}\left\langle \phi_{i,c}^{1}\phi_{i,c}^{2}\right\rangle _{\w}
    =
    \frac{\sigma_{a}^{2}}{N}\sum_{i=1}^{N}\left\langle \phi_{i,c}^{1}\phi_{i,c}^{2}\right\rangle _{\w}
\end{align}
The fourth moment is
\begin{equation}
    \left\langle f\left(\x^{1}\right)f\left(\x^{2}\right)f\left(\x^{3}\right)f\left(\x^{4}\right)\right\rangle _{\bf{a},\w} = 
    \sum_{i_{1:4}}\sum_{c_{1:4}}\left\langle a_{i_{1},c_{1}}a_{i_{2},c_{2}}a_{i_{3},c_{3}}a_{i_{4},c_{4}}\right\rangle _{\bf{a}}\left\langle \phi_{i_{1},c_{1}}^{1}\phi_{i_{2},c_{2}}^{2}\phi_{i_{3},c_{3}}^{3}\phi_{i_{4},c_{4}}^{4}\right\rangle_{\w}
\end{equation}
Averaging over the last layer weights gives 
\begin{align}
    \left\langle a_{i_{1},c_{1}}a_{i_{2},c_{2}}a_{i_{3},c_{3}}a_{i_{4},c_{4}}\right\rangle _{\bf{a}}=\left(\frac{\sigma_{a}^{2}}{CN}\right)^{2}\left(\delta_{i_{1}i_{2}}\delta_{c_{1}c_{2}}\delta_{i_{3}i_{4}}\delta_{c_{3}c_{4}}+\left\{ \left(13\right)\left(24\right)+\left(14\right)\left(23\right)\right\} \right)
\end{align}
So this will always make two pairs out of four $\phi$'s, each with the same $i,c$ indices. 
Notice that, regardless of the input indices, for different channels $c\neq c'$ we have
\begin{align}
    \left\langle \phi_{i,c}^{\mu}\phi_{i,c}^{\nu}\phi_{j,c'}^{\mu'}\phi_{j,c'}^{\nu'}\right\rangle _{\w}=\left\langle \phi_{i,c}^{\mu}\phi_{i,c}^{\nu}\right\rangle _{\w}\left\langle \phi_{j,c'}^{\mu'}\phi_{j,c'}^{\nu'}\right\rangle _{\w}
\end{align}
so, e.g. the first term out of three is 
\begin{align}
    & \sum_{i_{1:4}}\sum_{c_{1:4}}\delta_{i_{1}i_{2}}\delta_{c_{1}c_{2}}\delta_{i_{3}i_{4}}\delta_{c_{3}c_{4}}\left\langle \phi_{i_{1},c_{1}}^{1}\phi_{i_{2},c_{2}}^{2}\phi_{i_{3},c_{3}}^{3}\phi_{i_{4},c_{4}}^{4}\right\rangle _{\w}
    \\ &= \nonumber
    \sum_{i_{1},i_{3}}\sum_{c_{1},c_{3}}\left\langle \phi_{i_{1},c_{1}}^{1}\phi_{i_{1},c_{1}}^{2}\phi_{i_{3},c_{3}}^{3}\phi_{i_{3},c_{3}}^{4}\right\rangle _{\w}
    \\ &= \nonumber
    \sum_{i_{1},i_{3}}\left\{ \sum_{c}\left\langle \phi_{i_{1},c}^{1}\phi_{i_{1},c}^{2}\phi_{i_{3},c}^{3}\phi_{i_{3},c}^{4}\right\rangle _{\w}+\sum_{\begin{array}{c}
    c_{1},c_{3}\\
    c_{1}\ne c_{3}
    \end{array}}\left\langle \phi_{i_{1},c_{1}}^{1}\phi_{i_{1},c_{1}}^{2}\right\rangle _{\w}\left\langle \phi_{i_{3},c_{3}}^{3}\phi_{i_{3},c_{3}}^{4}\right\rangle _{\w}\right\} 
\end{align}
where in the last line we separated the diagonal and off-diagonal terms in the channel indices. So
\begin{align}
    & \left(\frac{\sigma_{a}^{2}}{CN}\right)^{-2}\left\langle f\left(\x^{1}\right)f\left(\x^{2}\right)f\left(\x^{3}\right)f\left(\x^{4}\right)\right\rangle _{\bf{a},\w}
    \\ &= \nonumber
    \sum_{i_{1},i_{2}}\sum_{c}\left\{ \left\langle \phi_{i_{1},c}^{1}\phi_{i_{1},c}^{2}\phi_{i_{2},c}^{3}\phi_{i_{2},c}^{4}\right\rangle _{\w}+\left\langle \phi_{i_{1},c}^{1}\phi_{i_{1},c}^{3}\phi_{i_{2},c}^{2}\phi_{i_{2},c}^{4}\right\rangle _{\w}+\left\langle \phi_{i_{1},c}^{1}\phi_{i_{1},c}^{4}\phi_{i_{2},c}^{2}\phi_{i_{2},c}^{3}\right\rangle _{\w}\right\} 
    \\ &+ \nonumber
    \sum_{i_{1},i_{2}}\sum_{\begin{array}{c}
    c_{1},c_{2}\\
    c_{1}\ne c_{2}
    \end{array}}\left\{ \left\langle \phi_{i_{1},c_{1}}^{1}\phi_{i_{1},c_{1}}^{2}\right\rangle _{\w}\left\langle \phi_{i_{2},c_{2}}^{3}\phi_{i_{2},c_{2}}^{4}\right\rangle _{\w}+\left\langle \phi_{i_{1},c_{1}}^{1}\phi_{i_{1},c_{1}}^{3}\right\rangle _{\w}\left\langle \phi_{i_{2},c_{2}}^{2}\phi_{i_{2},c_{2}}^{4}\right\rangle _{\w}+\left\langle \phi_{i_{1},c_{1}}^{1}\phi_{i_{1},c_{1}}^{4}\right\rangle _{\w}\left\langle \phi_{i_{2},c_{2}}^{2}\phi_{i_{2},c_{2}}^{3}\right\rangle _{\w}\right\} 
\end{align}
On the other hand 
\begin{align}
    & \left\langle f^{1}f^{2}\right\rangle \left\langle f^{3}f^{4}\right\rangle 
    \\ &= \nonumber
    \left(\frac{\sigma_{a}^{2}}{CN}\sum_{i=1}^{N}\sum_{c=1}^{C}\left\langle \phi_{i,c}^{1}\phi_{i,c}^{2}\right\rangle _{\w}\right)\left(\frac{\sigma_{a}^{2}}{CN}\sum_{i'=1}^{N}\sum_{c'=1}^{C}\left\langle \phi_{i',c'}^{3}\phi_{i',c'}^{4}\right\rangle _{\w}\right)
    \\ &= \nonumber
    \left(\frac{\sigma_{a}^{2}}{CN}\right)^{2}\sum_{i,i'=1}^{N}\sum_{c,c'=1}^{C}\left\langle \phi_{i,c}^{1}\phi_{i,c}^{2}\right\rangle _{\w}\left\langle \phi_{i',c'}^{3}\phi_{i',c'}^{4}\right\rangle _{\w}
    \\ &= \nonumber
    \left(\frac{\sigma_{a}^{2}}{CN}\right)^{2}\sum_{i,i'=1}^{N}\left\{ \sum_{\begin{array}{c}
    c,c'=1\\
    c\neq c'
    \end{array}}^{C}\left\langle \phi_{i,c}^{1}\phi_{i,c}^{2}\right\rangle _{\w}\left\langle \phi_{i',c'}^{3}\phi_{i',c'}^{4}\right\rangle _{\w}+\sum_{c=1}^{C}\left\langle \phi_{i,c}^{1}\phi_{i,c}^{2}\right\rangle _{\w}\left\langle \phi_{i',c}^{3}\phi_{i',c}^{4}\right\rangle _{\w}\right\} 
\end{align}
Putting it all together, the off-diagonal terms in the channel indices cancel and we are left with
\begin{align}
    & \left(\frac{\sigma_{a}^{2}}{CN}\right)^{-2}\kappa_{4}\left(\x_{1},\x_{2},\x_{3},\x_{4}\right)
    \\ &= \nonumber
    \left(\frac{\sigma_{a}^{2}}{CN}\right)^{-2}\left(\left\langle f^{1}f^{2}f^{3}f^{4}\right\rangle -\left(\left\langle f^{1}f^{2}\right\rangle \left\langle f^{3}f^{4}\right\rangle +\left\langle f^{1}f^{3}\right\rangle \left\langle f^{2}f^{4}\right\rangle +\left\langle f^{1}f^{4}\right\rangle \left\langle f^{2}f^{3}\right\rangle \right)\right)
    \\ &= \nonumber
    \sum_{i_{1},i_{2}}\sum_{c}\left\{ \left\langle \phi_{i_{1},c}^{1}\phi_{i_{1},c}^{2}\phi_{i_{2},c}^{3}\phi_{i_{2},c}^{4}\right\rangle _{\w}+\left\langle \phi_{i_{1},c}^{1}\phi_{i_{1},c}^{3}\phi_{i_{2},c}^{2}\phi_{i_{2},c}^{4}\right\rangle _{\w}+\left\langle \phi_{i_{1},c}^{1}\phi_{i_{1},c}^{4}\phi_{i_{2},c}^{2}\phi_{i_{2},c}^{3}\right\rangle _{\w}\right\} 
    \\ &- \nonumber
    \sum_{i_{1},i_{2}}\sum_{c}\left\{ \left\langle \phi_{i_{1},c}^{1}\phi_{i_{1},c}^{2}\right\rangle _{\w}\left\langle \phi_{i_{2},c}^{3}\phi_{i_{2},c}^{4}\right\rangle _{\w}+\left\langle \phi_{i_{1},c}^{1}\phi_{i_{1},c}^{3}\right\rangle _{\w}\left\langle \phi_{i_{2},c}^{2}\phi_{i_{2},c}^{4}\right\rangle _{\w}+\left\langle \phi_{i_{1},c}^{1}\phi_{i_{1},c}^{4}\right\rangle _{\w}\left\langle \phi_{i_{2},c}^{2}\phi_{i_{2},c}^{3}\right\rangle _{\w}\right\} 
    \\ &:= \nonumber
    \sum_{i,j=1}^{N}\sum_{c=1}^{C}\left\{ \left\langle \phi_{i,c}^{1}\phi_{i,c}^{2}\phi_{j,c}^{3}\phi_{j,c}^{4}\right\rangle _{\w}-\left\langle \phi_{i,c}^{1}\phi_{i,c}^{2}\right\rangle _{\w}\left\langle \phi_{j,c}^{3}\phi_{j,c}^{4}\right\rangle _{\w}\right\} +\left[\left(1_{i}3_{i}\right)\left(2_{j}4_{j}\right)+\left(1_{i}4_{i}\right)\left(2_{j}3_{j}\right)\right]
\end{align}
where in the last line we introduced a short-hand notation to compactly keep track of the combinations of the indices.

\subsubsection{Fourth cumulant for linear CNN}
Here, 
$\phi_{i,c}^{\mu}:=\w_{c}\cdot\tilde{\x}_{i}^{\mu}=\sum_{s=1}^{S}w_{s}^{\left(c\right)}\tilde{x}_{s}^{\left(\mu,i\right)}$. 
The fourth moment is
\begin{align}
    & \left\langle \phi_{i,c}^{1}\phi_{i,c}^{2}\phi_{j,c}^{3}\phi_{j,c}^{4}\right\rangle _{\w}
    \\ &= \nonumber
    \sum_{s_{1:4}=1}^{S}\left\langle \left(w_{s_{1}}^{\left(c\right)}\tilde{x}_{s_{1}}^{\left(1,i\right)}\right)\left(w_{s_{2}}^{\left(c\right)}\tilde{x}_{s_{2}}^{\left(2,i\right)}\right)\left(w_{s_{3}}^{\left(c\right)}\tilde{x}_{s_{3}}^{\left(3,j\right)}\right)\left(w_{s_{4}}^{\left(c\right)}\tilde{x}_{s_{4}}^{\left(4,j\right)}\right)\right\rangle _{\w}
    \\ &= \nonumber
    \sum_{s_{1:4}=1}^{S}\underbrace{\left\langle w_{s_{1}}^{\left(c\right)}w_{s_{2}}^{\left(c\right)}w_{s_{3}}^{\left(c\right)}w_{s_{4}}^{\left(c\right)}\right\rangle _{\w}}_{\left(\sigma_{w}^{2}/S\right)^{2}\cdot\delta_{s_{1}s_{2}}\delta_{s_{3}s_{4}}\left[3\right]}\tilde{x}_{s_{1}}^{\left(1,i\right)}\tilde{x}_{s_{2}}^{\left(2,i\right)}\tilde{x}_{s_{3}}^{\left(3,j\right)}\tilde{x}_{s_{4}}^{\left(4,j\right)}
    \\ &= \nonumber
    \left(\frac{\sigma_{w}^{2}}{S}\right)^{2}\sum_{s_{1:4}=1}^{S}\left(\delta_{s_{1}s_{2}}\delta_{s_{3}s_{4}}+\delta_{s_{1}s_{3}}\delta_{s_{2}s_{4}}+\delta_{s_{1}s_{4}}\delta_{s_{2}s_{3}}\right)\tilde{x}_{s_{1}}^{\left(1,i\right)}\tilde{x}_{s_{2}}^{\left(2,i\right)}\tilde{x}_{s_{3}}^{\left(3,j\right)}\tilde{x}_{s_{4}}^{\left(4,j\right)}
    \\ &= \nonumber
    \left(\frac{\sigma_{w}^{2}}{S}\right)^{2}\left\{ \left(\tilde{\x}_{i}^{1}\cdot\tilde{\x}_{i}^{2}\right)\left(\tilde{\x}_{j}^{3}\cdot\tilde{\x}_{j}^{4}\right)+\left(\tilde{\x}_{i}^{1}\cdot\tilde{\x}_{j}^{3}\right)\left(\tilde{\x}_{i}^{2}\cdot\tilde{\x}_{j}^{4}\right)+\left(\tilde{\x}_{i}^{1}\cdot\tilde{\x}_{j}^{4}\right)\left(\tilde{\x}_{i}^{2}\cdot\tilde{\x}_{j}^{3}\right)\right\} 
    \\ &:= \nonumber
    \left(\frac{\sigma_{w}^{2}}{S}\right)^{2}\left\{ \left(1_{i}2_{i}\right)\left(3_{j}4_{j}\right)+\left(1_{i}3_{j}\right)\left(2_{i}4_{j}\right)+\left(1_{i}4_{j}\right)\left(2_{i}3_{j}\right)\right\} 
\end{align}
Similarly 
\begin{align}
    \left(\frac{\sigma_{w}^{2}}{S}\right)^{-2}\left\langle \phi_{i,c}^{1}\phi_{i,c}^{3}\phi_{j,c}^{2}\phi_{j,c}^{4}\right\rangle _{\w} &= 
    \left(1_{i}3_{i}\right)\left(2_{j}4_{j}\right)+\left(1_{i}2_{j}\right)\left(3_{i}4_{j}\right)+\left(1_{i}4_{j}\right)\left(3_{i}2_{j}\right)
    \\ \nonumber
    \left(\frac{\sigma_{w}^{2}}{S}\right)^{-2}\left\langle \phi_{i,c}^{1}\phi_{i,c}^{4}\phi_{j,c}^{2}\phi_{j,c}^{3}\right\rangle _{\w} &=
    \left(1_{i}4_{i}\right)\left(3_{j}2_{j}\right)+\left(1_{i}3_{j}\right)\left(4_{i}2_{j}\right)+\left(1_{i}2_{j}\right)\left(4_{i}3_{j}\right)
\end{align}
Notice that the 2nd and 3rd terms have $\left(ij\right)\left(ij\right)$ 
while the first term has
$\left(ii\right)\left(jj\right)$.
The latter will cancel out with the
$\left\langle \phi_{i,c}^{\mu}\phi_{i,c}^{\nu}\right\rangle _{\w}\left\langle \phi_{j,c}^{\mu'}\phi_{j,c}^{\nu'}\right\rangle _{\w}$
terms. Thus
\begin{align}
    & \left[\cancel{\left(1_{i}2_{i}\right)\left(3_{j}4_{j}\right)}+\left(1_{i}3_{j}\right)\left(2_{i}4_{j}\right)+\left(1_{i}4_{j}\right)\left(2_{i}3_{j}\right)\right]
    \\ &+ \nonumber
    \left[\cancel{\left(1_{i}3_{i}\right)\left(2_{j}4_{j}\right)}+\left(1_{i}2_{j}\right)\left(3_{i}4_{j}\right)+\left(1_{i}4_{j}\right)\left(3_{i}2_{j}\right)\right]
    \\ &+ \nonumber
    \left[\cancel{\left(1_{i}4_{i}\right)\left(3_{j}2_{j}\right)}+\left(1_{i}3_{j}\right)\left(4_{i}2_{j}\right)+\left(1_{i}2_{j}\right)\left(4_{i}3_{j}\right)\right]
    \\ &- \nonumber
    \left[\cancel{\left(1_{i}2_{i}\right)\left(3_{j}4_{j}\right)}+\cancel{\left(1_{i}3_{i}\right)\left(2_{j}4_{j}\right)}+\cancel{\left(1_{i}4_{i}\right)\left(3_{j}2_{j}\right)}\right]
    \\ &= \nonumber
    \left(1_{i}3_{j}\right)\left(2_{i}4_{j}\right)+\left(1_{i}4_{j}\right)\left(2_{i}3_{j}\right)+\left(1_{i}2_{j}\right)\left(3_{i}4_{j}\right)+\left(1_{i}4_{j}\right)\left(3_{i}2_{j}\right)+\left(1_{i}3_{j}\right)\left(4_{i}2_{j}\right)+\left(1_{i}2_{j}\right)\left(4_{i}3_{j}\right)
    \\ &= \nonumber
    \left(1_{i}3_{j}\right)\left[\left(2_{i}4_{j}\right)+\left(4_{i}2_{j}\right)\right]+\left(1_{i}4_{j}\right)\left[\left(2_{i}3_{j}\right)+\left(3_{i}2_{j}\right)\right]+\left(1_{i}2_{j}\right)\left[\left(3_{i}4_{j}\right)+\left(4_{i}3_{j}\right)\right]
\end{align}

Denote 
$\lambda:=\frac{\sigma_{a}^{2}}{N}\frac{\sigma_{w}^{2}}{S}$
The fourth cumulant is 
\begin{align}
\label{appEq: kappa_4}
    & \kappa_4\left(\x_{1},\x_{2},\x_{3},\x_{4}\right)
    \\ & = \nonumber
    \frac{\lambda^{2}}{C}\sum_{i,j=1}^{N}\left\{ \left(1_{i}2_{j}\right)\left[\left(3_{i}4_{j}\right)+\left(4_{i}3_{j}\right)\right]+\left(1_{i}3_{j}\right)\left[\left(2_{i}4_{j}\right)+\left(4_{i}2_{j}\right)\right]+\left(1_{i}4_{j}\right)\left[\left(2_{i}3_{j}\right)+\left(3_{i}2_{j}\right)\right]\right\} 
\end{align}

Notice that all terms involve inner products between $\tilde{\x}$'s with \textit{different indices} $i,j$, i.e. mixing different convolutional windows. 
This means that $\kappa_4$, and also all higher order cumulants, \textit{cannot} be written in terms of the linear kernel, which does not mix different conv-window indices. 
This is in contrast to the kernel (second cumulant) of this linear CNN which is identical to that of a corresponding linear fully connected network (FCN): 
$K\left(\x,\x'\right)=\frac{\sigma_{a}^{2}\sigma_{w}^{2}}{NS}\x^{\transpose}\x'$
It is also in contrast to the higher cumulants of the corresponding linear FCN, where all cumulants can be expressed in terms of products of the linear kernel.

\subsection{Sixth cumulant and above}
The even moments in terms of cumulants for a vector valued RV with zero odd moments and cumulants are (see \cite{Mccullagh2017}):
\begin{align}
    \kappa^{\mu_{1}\mu_{2}}
    &=
    \kappa^{\mu_{1},\mu_{2}}
    \\ \nonumber
    \kappa^{\mu_{1}\mu_{2}\mu_{3}\mu_{4}}
    &= 
    \kappa^{\mu_{1},\mu_{2},\mu_{3},\mu_{4}}+\kappa^{\mu_{1},\mu_{2}}\kappa^{\mu_{3},\mu_{4}}\left[3\right]
    \\ \nonumber
    \kappa^{\mu_{1}\mu_{2}\mu_{3}\mu_{4}\mu_{5}\mu_{6}}
    &=
    \kappa^{\mu_{1},\mu_{2},\mu_{3},\mu_{4},\mu_{5},\mu_{6}}+\kappa^{\mu_{1},\mu_{2},\mu_{3},\mu_{4}}\kappa^{\mu_{5},\mu_{6}}\left[15\right]+\kappa^{\mu_{1},\mu_{2}}\kappa^{\mu_{3},\mu_{4}}\kappa^{\mu_{5},\mu_{6}}\left[15\right]
\end{align}
where the moments are on the l.h.s. (indices with no commas) and the cumulants are on the r.h.s. (indices are separated with commas). Thus, the sixth cumulant is 
\begin{align}
    \kappa^{\mu_{1},\mu_{2},\mu_{3},\mu_{4},\mu_{5},\mu_{6}}=\kappa^{\mu_{1}\mu_{2}\mu_{3}\mu_{4}\mu_{5}\mu_{6}}-\kappa^{\mu_{1},\mu_{2},\mu_{3},\mu_{4}}\kappa^{\mu_{5},\mu_{6}}\left[15\right]-\kappa^{\mu_{1},\mu_{2}}\kappa^{\mu_{3},\mu_{4}}\kappa^{\mu_{5},\mu_{6}}\left[15\right]
\end{align}
In the linear case, the analogue of $\kappa^{\mu_{1},\mu_{2},\mu_{3},\mu_{4}}\kappa^{\mu_{5},\mu_{6}}$ is ($15$ such pairings, where only the numbers "move", not the $i,j,k$)
\begin{align}
    \frac{1}{\lambda^{3}}K\left(\x_{1},\x_{2}\right)K\left(\x_{3},\x_{4}\right)K\left(\x_{5},\x_{6}\right)={\color{blue}\left(1_{i}2_{i}\right)\left(3_{j}4_{j}\right)\left(5_{k}6_{k}\right)}
\end{align}
and the analogue of $\kappa^{\mu_{1},\mu_{2}}\kappa^{\mu_{3},\mu_{4}}\kappa^{\mu_{5},\mu_{6}}$ is
\begin{align}
    & \frac{C}{\lambda^{3}}\kappa_{4}\left(\x_{1},\x_{2},\x_{3},\x_{4}\right)K\left(\x_{5},\x_{6}\right)    
    \\ &= \nonumber
    \left\{ \left(1_{i}2_{j}\right)\left[\left(3_{i}4_{j}\right)+\left(4_{i}3_{j}\right)\right]+\left(1_{i}3_{j}\right)\left[\left(2_{i}4_{j}\right)+\left(4_{i}2_{j}\right)\right]+\left(1_{i}4_{j}\right)\left[\left(2_{i}3_{j}\right)+\left(3_{i}2_{j}\right)\right]\right\} \left(5_{k}6_{k}\right)
    \\ &= \nonumber
    \left\{ \left(1_{i}2_{j}\right)\left(3_{i}4_{j}\right)+\left(1_{i}2_{j}\right)\left(4_{i}3_{j}\right)+{\color{red}\left(1_{i}3_{j}\right)\left(2_{i}4_{j}\right)}+\left(1_{i}3_{j}\right)\left(4_{i}2_{j}\right)+{\color{red}\left(1_{i}4_{j}\right)\left(2_{i}3_{j}\right)}+\left(1_{i}4_{j}\right)\left(3_{i}2_{j}\right)\right\} {\color{red}\left(5_{k}6_{k}\right)}
\end{align}
Below, we found the 6th moment for a linear CNN to be
\begin{align}
    & \left\langle \phi_{i,c}^{1}\phi_{i,c}^{2}\phi_{j,c}^{3}\phi_{j,c}^{4}\phi_{k,c}^{5}\phi_{k,c}^{6}\right\rangle 
    \\ &= \nonumber
    {\color{blue}\left(1_{i}2_{i}\right)\left(3_{j}4_{j}\right)\left(5_{k}6_{k}\right)}+{\color{red}\left(1_{i}3_{j}\right)\left(2_{i}4_{j}\right)\left(5_{k}6_{k}\right)}+{\color{red}\left(1_{i}4_{j}\right)\left(2_{i}3_{j}\right)\left(5_{k}6_{k}\right)}
    \\ &+ \nonumber
    {\color{red}\left(1_{i}2_{i}\right)\left(3_{j}5_{k}\right)\left(4_{j}6_{k}\right)}+\left(1_{i}3_{j}\right)\left(2_{i}5_{k}\right)\left(4_{j}6_{k}\right)+\left(1_{i}5_{k}\right)\left(2_{i}3_{j}\right)\left(4_{j}6_{k}\right)
    \\ &+ \nonumber
    {\color{red}\left(1_{i}2_{i}\right)\left(4_{j}5_{k}\right)\left(3_{j}6_{k}\right)}+\left(1_{i}5_{k}\right)\left(2_{i}4_{j}\right)\left(3_{j}6_{k}\right)+\left(1_{i}4_{j}\right)\left(2_{i}5_{k}\right)\left(3_{j}6_{k}\right)
    \\ &+ \nonumber
    {\color{red}\left(1_{i}5_{k}\right)\left(3_{j}4_{j}\right)\left(2_{i}6_{k}\right)}+\left(1_{i}3_{j}\right)\left(4_{j}5_{k}\right)\left(2_{i}6_{k}\right)+\left(1_{i}4_{j}\right)\left(3_{j}5_{k}\right)\left(2_{i}6_{k}\right)
    \\ &+ \nonumber
    {\color{red}\left(2_{i}5_{k}\right)\left(3_{j}4_{j}\right)\left(1_{i}6_{k}\right)}+\left(3_{j}5_{k}\right)\left(2_{i}4_{j}\right)\left(1_{i}6_{k}\right)+\left(4_{j}5_{k}\right)\left(2_{i}3_{j}\right)\left(1_{i}6_{k}\right)
\end{align}
Notice that for every blue term we have exactly $6$ red terms, so all of the colored terms will exactly cancel out and only the uncolored terms will survive. There are $8$ such uncolored terms for each one of the $15$ pairings, thus we will ultimately have $120$ such pairs, thus the sixth cumulant is 
\begin{equation}
\label{appEq: kappa_6}
    \kappa_{6}\left(\x_{1},\dots,\x_{6}\right)=\frac{\lambda^{3}}{C^{2}}\sum_{i,j,k=1}^{N}\left(\bullet_{i}\bullet_{j}\right)\left(\bullet_{i}\bullet_{k}\right)\left(\bullet_{j}\bullet_{k}\right)\left[120\right]
\end{equation}
where the $\left[120\right]$ stands for the number of ways to pair the numbers $\left\{ 1,...,6\right\}$  into the form 
$\left(\bullet_{i}\bullet_{j}\right)\left(\bullet_{i}\bullet_{k}\right)\left(\bullet_{j}\bullet_{k}\right)$. 

We can thus identify a pattern which we conjecture to hold for any even cumulant of arbitrary order $2m$:
\begin{equation}
\label{appEq: kappa_2m}
    \kappa_{2m}\left(\x_{1},\dots,\x_{2m}\right)=\frac{\lambda^{m}}{C^{m-1}}\sum_{i_{1},\dots,i_{m}=1}^{N}\left(\bullet_{i_{1}},\bullet_{i_{2}}\right)\cdots\left(\bullet{}_{i_{m-2}},\bullet{}_{i_{m-1}}\right)\left(\bullet{}_{i_{m-1}},\bullet{}_{i_{m}}\right)\cdots\left[\left(2m-1\right)!\right]
\end{equation}
where the indices $i_{1},\dots,i_{m}$ obey the following: 
\begin{enumerate}
    \item 
    Each index appears exactly twice in each summand.
    \item 
    Each index cannot be paired with itself, i.e. $\left(\bullet_{i_{1}},\bullet_{i_{1}}\right)$  is not allowed. 
    \item 
    The same pairing can appear more than once, e.g. $\left(1_{i}2_{j}\right)\left(3_{i}4_{j}\right)\left(5_{k}6_{\ell}\right)\left(7_{k}8_{\ell}\right)$ is OK, in that $i,j$ are paired together twice, and so are $k,\ell$.
\end{enumerate}

%%%%%%%%%%%%%%%%%%%%%%%%%%%%%%%%%%%%%%%%%%%%%%%%%%%%%%%%%%%%%%%%%%%%%%%%%%%%%%%%%%%%%%%%%%

\section{Feature learning phase transition}
\label{appendix: FL phase trans}

\subsection{Field theory derivation of the statistics of the hidden weights covariance}
\label{appendix: Sigma_W field theory}
Although our main focus was on the statistics of the DNN outputs, our function-space formalism can also be used to characterize the statistics of the weights of the intermediate hidden layers. 
Here we focus on the linear CNN toy model given in the main text, where the learnable parameters of the student are given by $\theta = \left\{ w_{c,s},a_{i,c}\right\} $. 
Consider first a prior distribution in output space, where throughout this section we denote: 
$\vec{f}\equiv\left(f_{1},\dots,f_{n}\right)$, i.e. the vector of outputs on the training set alone (without the test point). 
Since we are interested in the statistics of the hidden weights, we will introduce an appropriate source term in weight space $J_{c,s}$
\begin{equation}
\label{appEq: prior J w-space}
    P_{0}\left[\vec{f},\left\{ J_{c,s}\right\} \right]\propto\int dw\int da\exp\left(-\frac{1}{2\sigma_{w}^{2}}\sum_{c,s}\left(w_{c,s}-\sigma_{w}^{2}J_{c,s}\right)^{2}\right)P_{0}\left(a\right)\prod_{\mu=1}^{n}\delta\left(f_{\mu}-z_{\theta,\mu}\right)
\end{equation}
where $z_{\theta,\mu}$ is the of output of the CNN parameterized by $\theta$ on the $\mu$'th training point.  
Given some loss function $\cL$, the posterior is given by
\begin{equation}
    P\left[\vec{f}, \left\{ J_{c,s}\right\} \right] = P_{0}\left[\vec{f}, \left\{ J_{c,s}\right\} \right]e^{-\cL/\sigma^{2}}
\end{equation}
The posterior mean of the hidden weights is thus
\begin{equation}
\label{appEq: w posterior mean}
    \eval{\pa_{J_{c,s}}\log\left(\int_{\bbR^n} d \vec{f} P\left[\vec{f}, \left\{ J_{c,s}\right\} \right]\right)}_{J=0} = \eval{\left\langle w_{c,s}\right\rangle _{P\left[\vec{f}, \left\{ J_{c,s}\right\} \right]}-\sigma_{w}^{2}J_{c,s}}_{J=0} = \left\langle w_{c,s}\right\rangle _{P\left[\vec{f}, \left\{ J_{c,s}\right\} \right]}
\end{equation}
and the posterior covariance can be extracted from taking the second derivative, namely 
\begin{align}
\label{appEq: w posterior covariance}
    & \eval{\pa_{J_{c_1,s_1}}\pa_{J_{c_2,s_2}}\log\left(\int_{\bbR^n} d \vec{f}P\left[\vec{f}, \left\{ J_{c,s}\right\} \right]\right)}_{J=0} 
    \\ &= \nonumber
    \eval{\left\langle w_{c_{1},s_{1}}w_{c_{2},s_{2}}\right\rangle _{P\left[\vec{f}, \left\{ J_{c,s}\right\} \right]}+\sigma_{w}^{4}J_{c_{1},s_{1}}J_{c_{2},s_{2}}}_{J=0}-\sigma_{w}^{2}\delta_{s_{1}s_{2}}\delta_{c_{1}c_{2}}
    \\ &= \nonumber
    \left\langle w_{c_{1},s_{1}}w_{c_{2},s_{2}}\right\rangle -\sigma_{w}^{2}\delta_{s_{1}s_{2}}\delta_{c_{1}c_{2}}
\end{align}

Our next task is to rewrite these expectation values over weights under the posterior as expectation values of DNN training outputs ($f(\x_\mu)$) under the posterior. 
To this end we write down the kernel of this simple CNN such that it depends on the source terms: 
\begin{align}
    K_{J}\left(\x,\x'\right) &= 
    \sum_{i,i',c,c',s,s'}\left\langle a_{i,c}w_{c,s}\tilde{x}_{i,s}a_{i',c'}w_{c',s'}\tilde{x}'_{i',s'}\right\rangle _{P\left[\vec{f}, J_{c,s}\right]}
    \\ &= \nonumber
    \underbrace{\sum_{i,i',c,c'}\left\langle a_{i,c}a_{i',c'}\right\rangle _{a}}_{\delta_{ii'}\delta_{cc'}/CN}\sum_{s,s'}\left\langle w_{c,s}\tilde{x}_{i,s}w_{c',s'}\tilde{x}'_{i',s'}\right\rangle _{w,J}
    \\ &= \nonumber
    \frac{1}{CN}\sum_{i,c}\sum_{s,s'}\left\langle w_{c,s}\tilde{x}_{i,s}w_{c,s'}\tilde{x}'_{i,s'}\right\rangle _{w,J}
    \\ &= \nonumber
    \frac{1}{N}\sum_{i}\sum_{s,s'}\frac{1}{C}\sum_{c}\underbrace{\left\langle w_{c,s}w_{c,s'}\right\rangle _{w,J}}_{\left(1/S\right)\delta_{ss'}+\left(1/S^{2}\right)J_{cs}J_{cs'}}\tilde{x}_{i,s}\tilde{x}'_{i,s'}
    \\ &= \nonumber
    \frac{1}{NS}\frac{1}{C}\sum_{i}\sum_{s}\sum_{c}\tilde{x}_{i,s}\tilde{x}'_{i,s}+\frac{1}{NS^{2}}\sum_{i}\sum_{s,s'}\underbrace{\left(\frac{1}{C}\sum_{c}J_{cs}J_{cs'}\right)}_{\equiv B_{ss'}}\tilde{x}_{i,s}\tilde{x}'_{i,s'}
    \\ &= \nonumber
    \frac{1}{NS}\x^{\transpose}\x'+\frac{1}{NS^{2}}\sum_{i}\tilde{\x}_{i}^{\transpose}B\tilde{\x}_{i}'
\end{align}
where $B\in\bbR^{S\times S}$. This can be written as ($d=NS$)
\begin{equation}
    K_{J}\left(\x,\x'\right)=\frac{1}{NS}\x^{\transpose}\left(I_{d}+\frac{1}{S}\left(\begin{matrix}B\\
 & \ddots\\
 &  & B
\end{matrix}\right)\right)\x'
\end{equation}
We can now write the second mixed derivatives of $K_J$ to leading order in $J$ as
\begin{align}
    -\pa_{J_{c_{1},s_{1}}}\pa_{J_{c_{2},s_{2}}}K_{J}^{-1}\left(\x,\x'\right) &= 
    -NS\x^{\transpose}\left[\pa_{J_{c_{1},s_{1}}}\pa_{J_{c_{2},s_{2}}}\left(I_{d}+\frac{1}{S}\left(\begin{matrix}B\\
 & \ddots\\
 &  & B
\end{matrix}\right)\right)^{-1}\right]\x'
    \\ &= \nonumber
    -NS\x^{\transpose}\left[\pa_{J_{c_{1},s_{1}}}\pa_{J_{c_{2},s_{2}}}\left(I_{d}-\frac{1}{S}\left(\begin{matrix}B\\
 & \ddots\\
 &  & B
\end{matrix}\right)\right)\right]\x'
    \\ &= \nonumber
    \frac{N}{C}\sum_{i}\sum_{s,s'}\tilde{x}_{i,s}\tilde{x}'_{i,s'}\pa_{J_{c_{1},s_{1}}}\pa_{J_{c_{2},s_{2}}}\sum_{c}J_{cs}J_{cs'}
    \\ &= \nonumber
    \frac{N}{C}\delta_{c_{1}c_{2}}\sum_{i}\sum_{s,s'}\tilde{x}_{i,s}\tilde{x}'_{i,s'}\left(\delta_{ss_{2}}\delta_{s_{1}s'}+\delta_{s's_{2}}\delta_{s_{1}s}\right)
    \\ &= \nonumber
    2\frac{N}{C}\delta_{c_{1}c_{2}}\sum_{i}\tilde{x}_{i,s_{1}}\tilde{x}'_{i,s_{2}}
\end{align}

Next we take the large $C$ limit and thus have a posterior of the form 
$P[\vec{f}, J] = P_0[\vec{f}, J]e^{-\cL/\sigma^2}$ 
where $P_0[\vec{f}]$ contains only $K_J^{-1}$ and none of the higher cumulants. 
Having the derivatives of $K_J^{-1}$ w.r.t. $J$ we can proceed in analyzing the derivatives of the log-partition function for the posterior w.r.t $J$. 
In particular the covariance matrix of the weights averaged over the different channels is 
\begin{align}
    & \frac{1}{C}\sum_{c_{1},c_{2}}\pa_{J_{c_{1},s_{1}}}\pa_{J_{c_{2},s_{2}}}\log\left(\int_{\bbR^n} d \vec{f}P\left[\vec{f},J\right]\right) 
    \\ &= \nonumber
    -\frac{1}{C}\sum_{\mu,\nu=1}^{n}\left\{ \sum_{c_{1},c_{2}}\left[\pa_{J_{c_{1},s_{1}}}\pa_{J_{c_{2},s_{2}}}K_{J}^{-1}\left(\x_{\mu},\x_{\nu}\right)\right]\frac{\int_{\bbR^{n}}d\vec{f}P[\vec{f}]f\left(\x_{\mu}\right)f\left(\x_{\nu}\right)}{\int_{\bbR^{n}}d\vec{f}P[\vec{f}]}\right\} 
    \\ &= \nonumber
    2\frac{N}{C^{2}}\sum_{i}\sum_{\mu,\nu=1}^{n}\tilde{x}_{i,s_{1}}^{\mu}\tilde{x}{}_{i,s_{2}}^{\nu}\frac{\int_{\bbR^{n}}d\vec{f}P[\vec{f}]f\left(\x_{\mu}\right)f\left(\x_{\nu}\right)}{\int_{\bbR^{n}}d\vec{f}P[\vec{f}]}\sum_{c_{1},c_{2}}\delta_{c_{1}c_{2}}
    \\ &= \nonumber
    2\frac{N}{C}\sum_{i}\sum_{\mu,\nu=1}^{n}\tilde{x}_{i,s_{1}}^{\mu}\tilde{x}{}_{i,s_{2}}^{\nu}\frac{\int_{\bbR^{n}}d\vec{f}P[\vec{f}]f\left(\x_{\mu}\right)f\left(\x_{\nu}\right)}{\int_{\bbR^{n}}d\vec{f}P[\vec{f}]}
\end{align}
The above result is one of the two main points of this appendix: we established a mapping between expectation values over outputs and expectation values over hidden weights. 
Such a mapping can in principle be extended to any DNN. 
On the technical level, it requires the ability to calculate the cumulants as a function of the source terms, $J$. 
As we argue below, it may very well be that unlike in the main text, only a few cumulants are needed here. 

To estimate the above expectation values we use the EK limit, where the sums over the training set are replaced by integrals over the measure $\mu(\x)$, the $f$'s are replaced as 
$f\left(\x_{\mu}\right)\to\frac{\lambda}{\lambda+\sigma^{2}/n}g\left(\x\right)$ 
and we assume the input distribution is normalized as 
$\int d\mu\left(\x\right)x_{i}x_{j}=\delta_{ij}$. 
Following this we find 
\begin{align}
    & 2\frac{N}{C}\left(\frac{\lambda}{\lambda+\sigma^{2}/n}\right)^{2}\int d\mu\left(\x\right)d\mu\left(\x'\right)g\left(\x\right)g\left(\x'\right)\sum_{i}\tilde{x}_{i,s_{1}}\tilde{x}'_{i,s_{2}}
    \\ &= \nonumber
    2\frac{N}{C}\left(\frac{\lambda}{\lambda+\sigma^{2}/n}\right)^{2}\sum_{i}\underbrace{\int d\mu\left(\x\right)g\left(\x\right)\tilde{x}_{i,s_{1}}}_{a_{i}^{*}w_{s_{1}}^{*}}\underbrace{\int d\mu\left(\x'\right)g\left(\x'\right)\tilde{x}'_{i,s_{2}}}_{a_{i}^{*}w_{s_{2}}^{*}}
    \\ &= \nonumber
    2\frac{N}{C}\left(\frac{\lambda}{\lambda+\sigma^{2}/n}\right)^{2}\underbrace{\sum_{i}\left(a_{i}^{*}\right)^{2}}_{1}w_{s_{1}}^{*}w_{s_{2}}^{*}
    \\ &= \nonumber
    2\frac{N}{C}\left(\frac{\lambda}{\lambda+\sigma^{2}/n}\right)^{2}w_{s_{1}}^{*}w_{s_{2}}^{*}
\end{align}
Comparing this to our earlier result for the covariance Eq.  \ref{appEq: w posterior covariance} we get 
\begin{align}
    2\frac{N}{C}\left(\frac{\lambda}{\lambda+\sigma^{2}/n}\right)^{2}w_{s_{1}}^{*}w_{s_{2}}^{*} &=
    \frac{1}{C}\sum_{c_{1},c_{2}}\left(\left\langle w_{c_{1},s_{1}}w_{c_{2},s_{2}}\right\rangle -\sigma_{w}^{2}\delta_{s_{1}s_{2}}\delta_{c_{1}c_{2}}\right)
    \\ &= \nonumber
    \frac{1}{C}\sum_{c_{1},c_{2}}\left\langle w_{c_{1},s_{1}}w_{c_{2},s_{2}}\right\rangle -\sigma_{w}^{2}\delta_{s_{1}s_{2}}
\end{align}
Multiplying by $S=1/\sigma_{w}^{2}$ and recalling that $\lambda=1/NS$ we get
\begin{equation}
    \left\langle \left[\Sigma_{W}\right]_{s_{1}s_{2}}\right\rangle =\delta_{s_{1}s_{2}}+\frac{2}{C}\frac{\lambda}{\left(\lambda+\sigma^{2}/n\right)^{2}}w_{s_{1}}^{*}w_{s_{2}}^{*}+O\left(1/C^{2}\right)
\end{equation}
Repeating similar steps while also taking into account diagonal fluctuations yields another factor of $\left(\frac{1}{\lambda}+\frac{n}{\sigma^{2}}\right)^{-1}$ on the diagonal, thus arriving at the result as it appears in the main text:
\begin{equation}
    \left\langle \left[\Sigma_{W}\right]_{s_{1}s_{2}}\right\rangle =\left(1+\left(\frac{1}{\lambda}+\frac{n}{\sigma^{2}}\right)^{-1}\right)\delta_{s_{1}s_{2}}+\frac{2}{C}\frac{\lambda}{\left(\lambda+\sigma^{2}/n\right)^{2}}w_{s_{1}}^{*}w_{s_{2}}^{*}+O\left(1/C^{2}\right)
\end{equation}
The above results capture the leading order correction in $1/C$ to the weights covariance matrix. 
However the careful reader may be wary of the fact that the results in the main text require $1/C$ corrections to all orders and so it is potentially inadequate to use such a low order expansion deep in the feature learning regime, as we do in the main text. 
Here we note that not all DNN quantities need to have the same dependence on $C$. 
In particular it was shown in Ref. \cite{li2020statistical}, that the weight's low order statistics is only weakly affected by finite-width corrections whereas the output covariance matrix is strongly affected by these. 
We conjecture that this is the case here and that only the cumulative effect of many weights, as reflected in the output of the DNN, requires strong $1/C$ corrections. 

This conjecture can be verified analytically by repeating the above procedure on the full prior (i.e. the one that contains all cumulants), obtaining the operator in terms of $f$'s corresponding the weight's covariance matrix, and calculating its average with respect to the saddle point theory. 
We leave this for future work.

\subsection{A surrogate quantity for the outlier}
\label{appendix: outlier surrogate}
Since we used moderate $S$ values in our simulations (to maintain a reasonable compute time), we aggregated the eigenvalues of many instances of $\Sigma_W$ across training time and across noise realizations. 
Although the empirical histogram of the spectrum of $\Sigma_W$ agrees very well with the theoretical MP distribution (solid smooth curves in Fig. \ref{fig: phase trans}A), there is a substantial difference between the two at the right edge of the support $\lambda_+$, where the empirical histogram has a tail due to finite size effects. 
Thus it is hard to characterize the phase transition using the largest eigenvalue $\lambda_{\max}$ averaged across realizations. 
Instead, we use the quantity 
$\cQ \equiv \w^{*\transpose} \Sigma_W \w^*$ 
as a surrogate which coincides with $\lambda_{\max}$ for $C\ll C_{\rm{crit}}$ but behaves sensibly on both sides of $C_{\rm{crit}}$, thus allowing to characterize the phase transition.

\section{Further details on the numerical experiments}
\label{appendix: numerics}
\subsection{Additional details of the numerical experiments}
\label{appendix: details numerics}
In our experiments, we used the following hyper-parameter values.
Learning rates of $\eta = 10^{-6}, 3\cdot 10^{-7}$ which yield results with no appreciable difference in almost all cases, when we scale the amount of statistics collected (training epochs after reaching equilibrium) so that both $\eta$ values have the same amount of re-scaled training  time: we used $10$ training seeds for $\eta = 10^{-6}$ and 30 for $\eta = 3\cdot10^{-7}$. 
We used a gradient noise level of $\sigma^2 = 1.0$, but also checked for $\sigma^2 \in \{ 0.1, 0.01 \}$ and got qualitatively similar results to those reported in the main text. 

\begin{figure}[h]
\vspace*{-0.1in}
\begin{center}
(A)
\includegraphics[height=4cm]{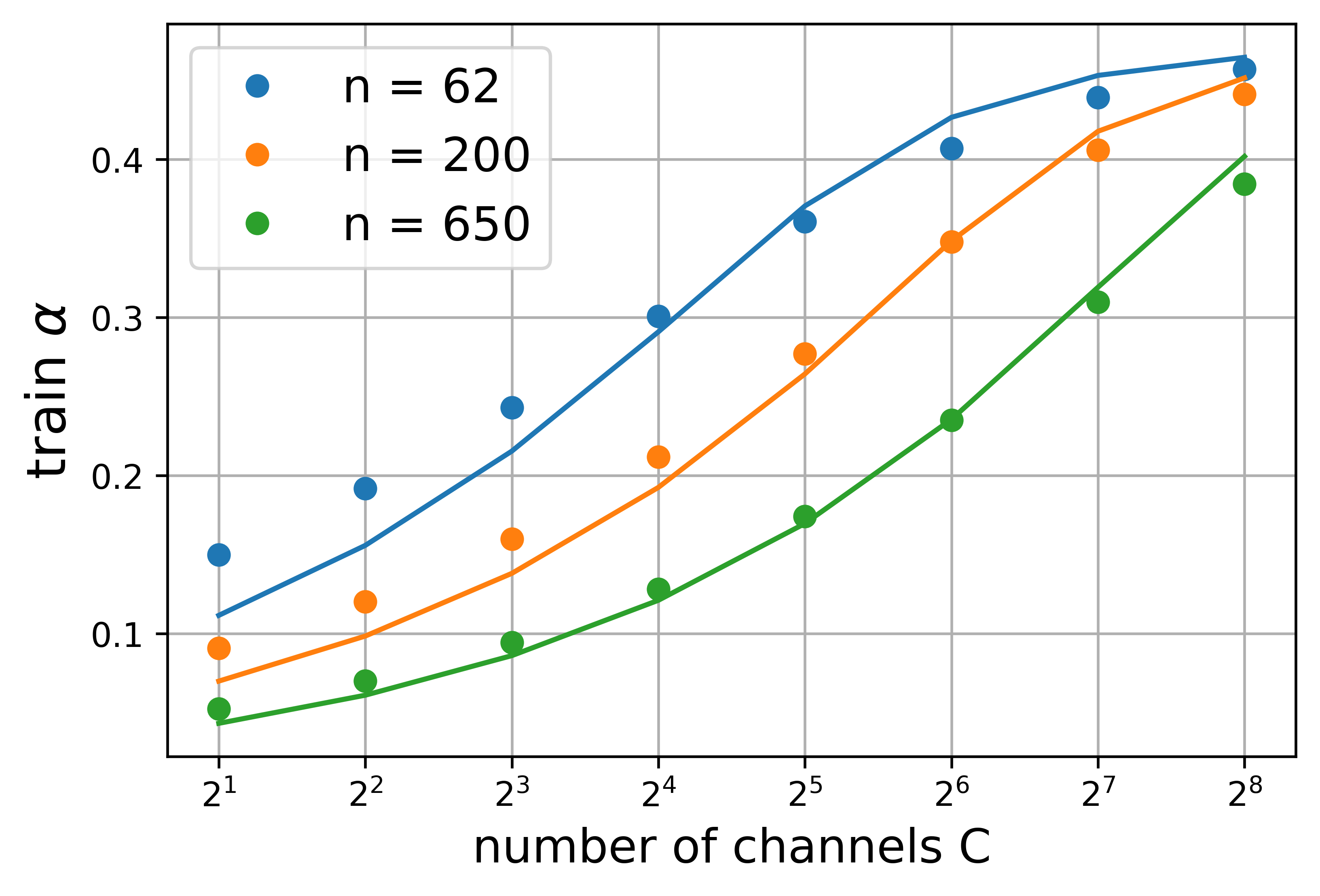}
(B)
\includegraphics[height=4cm]{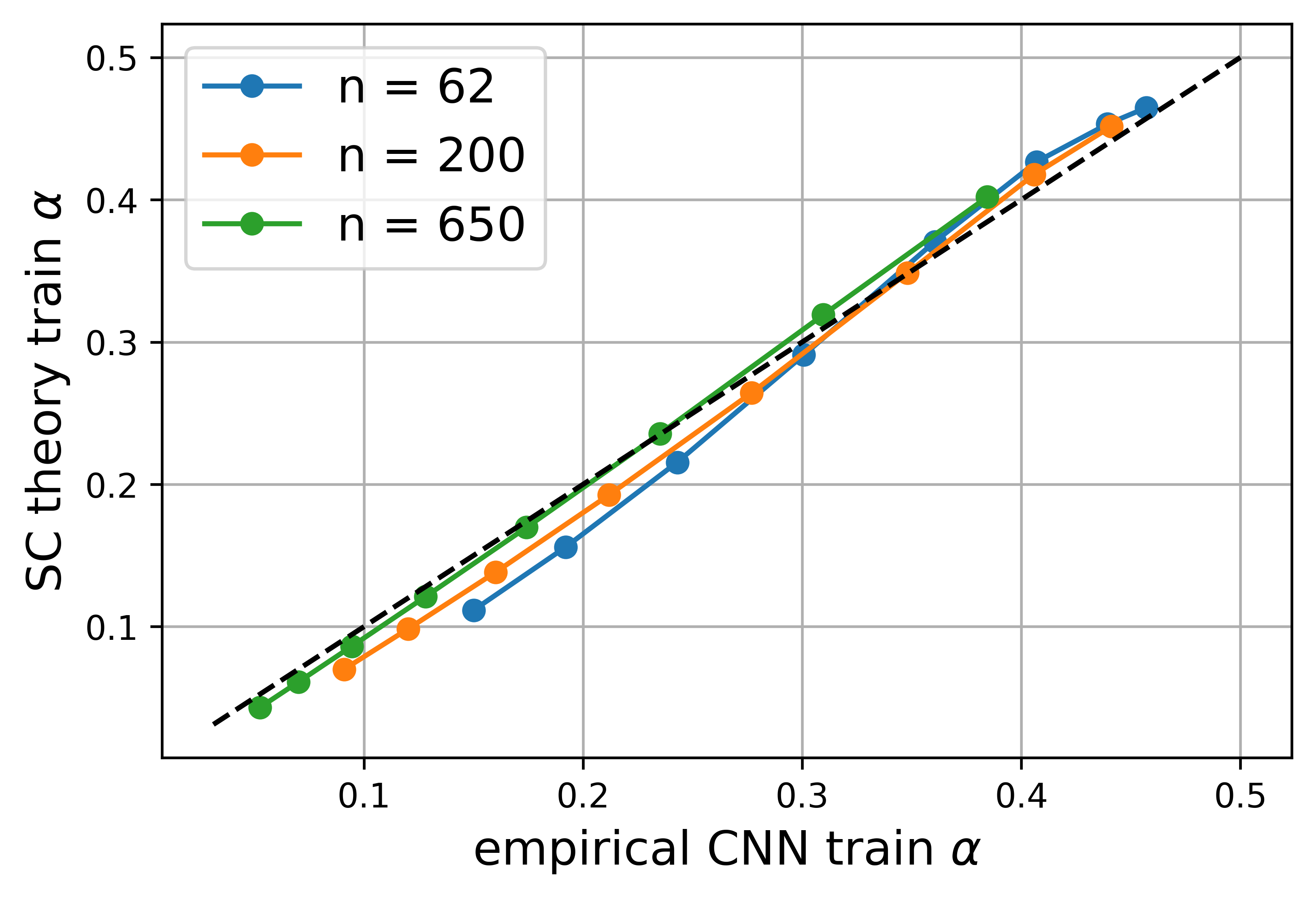}

\end{center}
\caption{ 
\textbf{(A)} 
The CNNs' cosine distance $\alpha$, defined by $\left\langle f\right\rangle  = (1 - \alpha) g$
between the ensemble-averaged prediction $\left\langle f\right\rangle $ and ground truth $g$ plotted vs. number of channels $C$ for the training set (for the test set see Fig. \ref{fig: SC theory verification} in the main text).  
As $n$ increases, the solution of the self consistent equation \ref{eq: SC alpha all kappas no SL} (solid line) yields an increasingly accurate prediction of these empirical values (dots).
\textbf{(B)} 
Same data as in (A), presented as empirical $\alpha$ vs. predicted $\alpha$.
As $n$ grows, the two converge to the identity line (dashed black line). Solid lines connecting the dots here are merely for visualization purposes.  
\label{appFig: SC theory verification}}
\end{figure}

In the main text and here we do not show error bars for $\alpha$ as these are too small to be appreciated visually. 
They are smaller than the mean values by approximately two orders of magnitude.
The error bars were found by computing the empirical standard deviation of $\alpha$ across training dynamics and training seeds. 

%%%%%%%%%%%%%%%%%%%%%%%%%%%%%%%%%%%%%%%%%%%%%%%%%%%%%%%%%%%%
%%%%%%%%%%%%%%%%%%%%%%%%%%%%%%%%%%%%%%%%%%%%%%%%%%%%%%%%%%%%

\subsection{Convergence of the training protocol to GP }
\label{appendix: GP convergence}
\begin{figure}[h]
\begin{center}

\includegraphics[height=3cm]{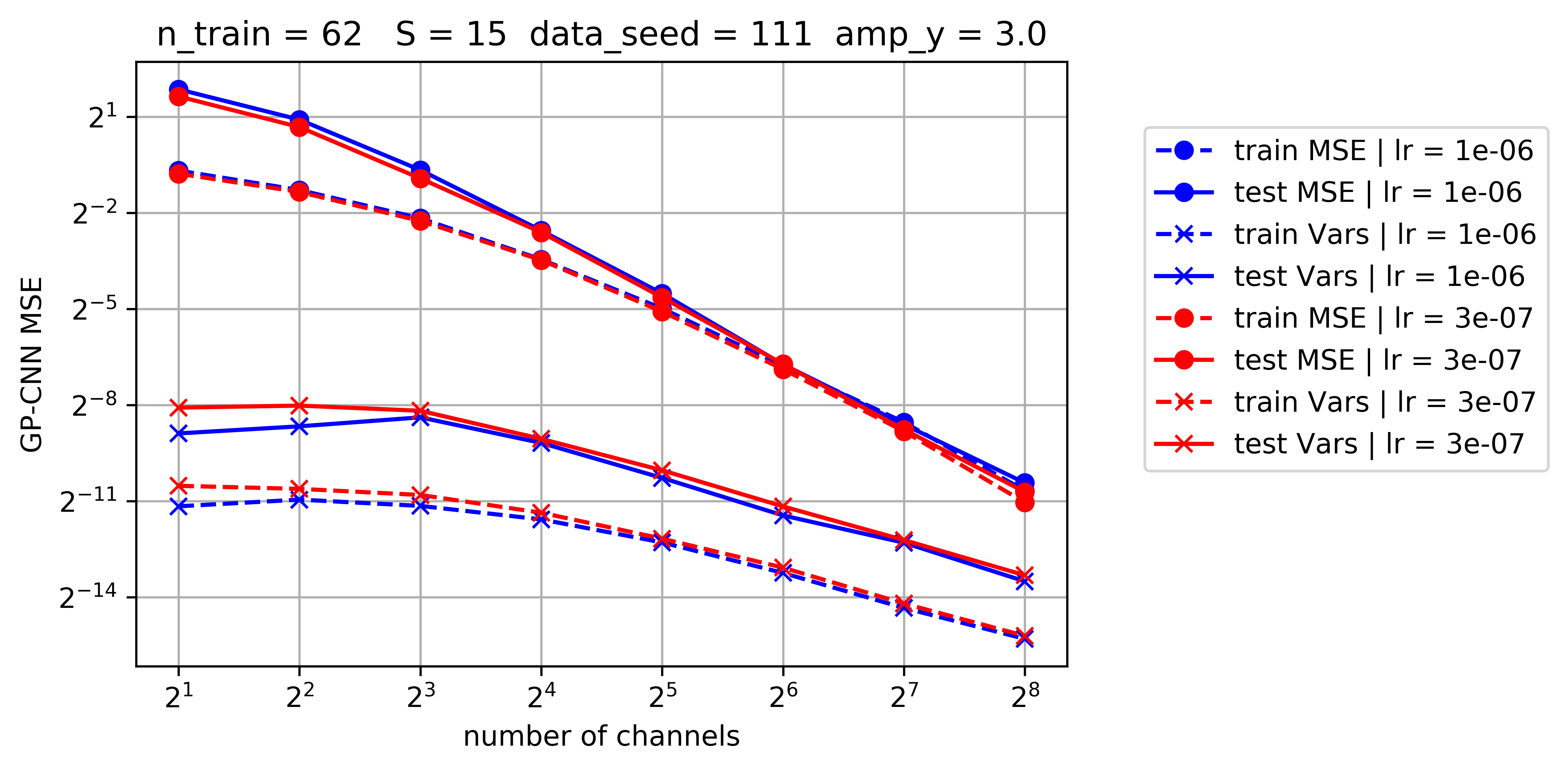}
\includegraphics[height=3cm]{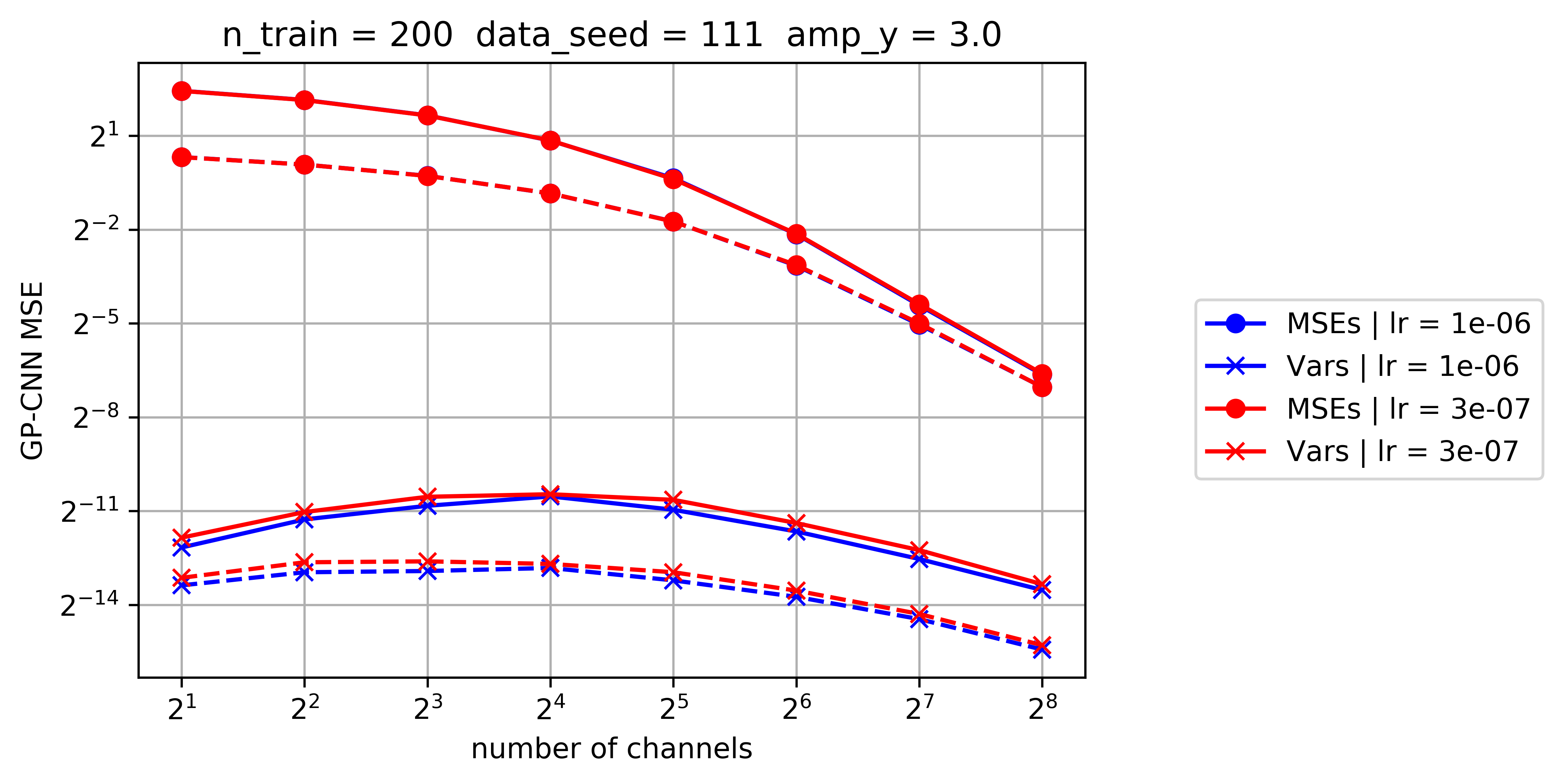}

\end{center}
\caption{ CNN-GP MSEs for different $S$, indicating where the perturbative regime starts (slope approaching $-2.0$). 
For $S=15$ this happens around $C=2^5$ whereas for $S=30$ this happens around $C=2^7$. 
\label{fig: CNN-GP MSEs}}
\end{figure}

In Fig. \ref{fig: CNN-GP MSEs} we plot the MSE between the outputs of the trained CNNs and the predictions of the corresponding GP. 
We see that as $C$ becomes large the slope of the MSE tends to $-2.0$ indicating the $O(1/C)$ scaling of the leading corrections to the GP.  
This illustrates where we enter the perturbative regime of GP, and we see that this happens for larger $C$ as we increase the conv-kernel size $S$, since this also increases the input dimension $d=NS$. 
Thus it takes larger $C$ to enter the highly over-parameterized regime. 

\section{Quadratic fully connected network}
\label{appendix: LenkasModel}
One of the simplest settings where GPs are expected to strongly under-perform finite DNNs is the case of quadratic fully connected DNNs \cite{mannelli2020optimization}. 
Here we consider some positive target of the form 
$g(\x)=(\w_* \cdot \x)^2 - \sigma_w^2 ||\x||^2$ where $\w_*,\x \in {\bbR}^d$ and a student DNN given by 
$f(\x) = \sum_{m=1}^M (\w_m \cdot \x)^2 - \sigma_w^2 ||\x||^2$ \footnote{The $||\x||^2$ shift is not part of the original model but has only a superficial shift effect useful for book-keeping later on}. 

At large $M$ and for $w_{m,i}$ drawn from ${\mathcal N}(0,\sigma_w^2/M)$, the student generates a GP prior. 
It is shown below that the GP kernel is simply $K(\x,\x')=\frac{2\sigma_w^4}{M}(\x \cdot \x')^2$. 
As such it is proportional to the kernel of the above DNN with an additional linear read-out layer. 
The above model can be written as 
$\sum_{ij} x_i [P_{ij}-\sigma_w^2 \delta_{ij}] x_j$ where $P_{ij}$ is a positive semi-definite matrix. The eigenvalues of the matrix appearing within the brackets are therefore larger than $-\sigma_w^2$ whereas no similar restriction occurs for DNNs with a linear read-out layer. This extra restriction is completely missed by the GP approximation and, as discussed in Ref. \cite{mannelli2020optimization}, leads to strong performance improvements compared to what one expects from the GP or equivalently the DNN with the linear readout layer. Here we demonstrate that our self-consistent approach at the saddle-point level captures this effects

We consider training this DNN on $n$ train points  $\left\{ \x_{\mu}\right\} _{\mu=1}^{n}$ using noisy GD training with weight decay $\gamma = M \sigma^2/\sigma_w^2$. 
We wish to solve for the predictions of this model with our shifted target approach. To this end, we first derive the cumulants associated with the effective Bayesian prior ($P_0(\vec{f})$) here. 
Equivalently stated, obtain the cumulants of the equilibrium distribution of $\vec{f}$ following training with no data, only a weight decay term. 
This latter distribution is given by
\begin{align}
P_{0}\left(\vec{f}\right)=\int d\w e^{-\frac{M}{2\sigma_{w}^{2}}\sum_{m=1}^{M}||\w_{m}||^{2}}\prod_{\mu=1}^{n+1}\delta\left(f_{\mu}-\sum_{m=1}^{M} \left(\w_{m}\cdot\x_{\mu}\right)^2 + \sigma_{w}^{2}||\x_{\mu}||^{2}\right)
\end{align}
To obtain the cumulants, we calculate the cumulant generating function of this distribution given by 
\begin{align}
    & \cC(t_1,...,t_{n+1}) \\ \nonumber
    &= \log\left(\int\prod_{m,i=1,1}^{M,d}\frac{dw_{m,i}}{\sqrt{2\pi M^{-1}\sigma_{w}^{2}}}e^{-\sum_{m,i=1,1}^{M,d}M\frac{w_{m,i}^{2}}{2\sigma_{w}^{2}}+\sum_{\mu=1}^{n}it_{\mu}\left[\sum_{m,i=1,1}^{M,d}\left(\w_{m}\cdot\x_{\mu}\right)^{2}-\sigma_{w}^{2}||\x_{\mu}||^{2}\right]}\right) 
    \\ \nonumber
    &= M\log\left(\int\prod_{i=1}^{d}\frac{dw_{i}}{\sqrt{2\pi M^{-1}\sigma_{w}^{2}}}e^{-M\frac{||\w||^{2}}{2\sigma_{w}^{2}}+\sum_{\mu=1}^{n+1}it_{\mu}\left[\left(\w\cdot\x_{\mu}\right)^{2}\right]}\right)-\sum_{\mu=1}^{n+1}it_{\mu}\sigma_{w}^{2}||\x_{\mu}||^{2}
     \\ \nonumber
    &= M\log\left(\int\prod_{i=1}^{d}\frac{dw_{i}}{\sqrt{2\pi M^{-1}\sigma_{w}^{2}}}e^{-\frac{\w^{\transpose}\left[I-2M^{-1}\sigma_{w}^{2}\sum_{\mu}it_{\mu}\x_{\mu}\x_{\mu}^{\transpose}\right]\w}{2M^{-1}\sigma_{w}^{2}}}\right)-\sum_{\mu=1}^{n+1}it_{\mu}\sigma_{w}^{2}||\x_{\mu}||^{2}
    \\ \nonumber
    &= -\frac{M}{2}\log\left(\det\left[I-2M^{-1}\sigma_{w}^{2}\sum_{\mu}it_{\mu}\x_{\mu}\x_{\mu}^{\transpose}\right]\right)-\sum_{\mu=1}^{n+1}it_{\mu}\sigma_{w}^{2}||\x_{\mu}||^{2}
    \\ \nonumber
    &= -\frac{M}{2}\Tr\left(\log\left[I-2M^{-1}\sigma_{w}^{2}\sum_{\mu}it_{\mu}\x_{\mu}\x_{\mu}^{\transpose}\right]\right)-\sum_{\mu=1}^{n+1}it_{\mu}\sigma_{w}^{2}||\x_{\mu}||^{2}
\end{align}

Taylor expanding this last expression is straightforward. For instance up to third order is gives
\begin{align}
    \cC(t_1,...,t_{n+1}) &= \frac{M}{2}\sum_{\mu_1,\mu_2}(2M^{-1} \sigma_w^2)^2\frac{it_{\mu_1} it_{\mu_2}}{2} (\x_{\mu_1}\cdot \x_{\mu_2})(\x_{\mu_2}\cdot \x_{\mu_1}) \\ \nonumber &+ \frac{M}{2}\sum_{\mu_1,\mu_2,\mu_3}(2M^{-1} \sigma_w^2)^3\frac{it_{\mu_1} it_{\mu_2} it_{\mu_3}}{3} (\x_{\mu_1}\cdot \x_{\mu_2})(\x_{\mu_2}\cdot \x_{\mu_3})(\x_{\mu_3}\cdot \x_{\mu_1}) + ...
\end{align}
from which the cumulants can be directly inferred, in particular the associated GP kernel given by 
\begin{align}
    K\left(\x_{\mu},\x_{\nu}\right)=2M^{-1}\sigma_{w}^{4}\left(\x_{\mu}\cdot\x_{\nu}\right)^{2}
\end{align}

Following this, the target shift equation, at the saddle point level, becomes 
\begin{align}
    \Delta g_{\nu} &= \partial_{t_{\nu}} \left( \cC(t_1..t_{n},t_{n+1}=0)- \sum_{\mu_1,\mu_2}\frac{K(\x_{\mu_1},\x_{\mu_2})}{2!} it_{\mu_1} it_{\mu_2} \right)|_{t_1..t_{n}=\frac{\hat{\delta} g_1}{\sigma^2}..\frac{\hat{\delta} g_n}{\sigma^2}}  \\ \nonumber
    &= -\sum_{\mu}K\left(\x_{\nu},\x_{\mu}\right)\frac{\hat{\delta}g_{\mu}}{\sigma^{2}}+\sigma_{w}^{2}\Tr\left(\x_{\nu}\x_{\nu}^{\transpose}\left[I-2M^{-1}\sigma_{w}^{2}\sum_{\mu}\frac{\hat{\delta}g_{\mu}}{\sigma^{2}}\x_{\mu}\x_{\mu}\right]^{-1}\right)-\sigma_{w}^{2}||\x_{\nu}||^{2}
    \\ \nonumber
    &= -\sum^n_{\mu=1}K\left(\x_{\nu},\x_{\mu}\right)\frac{\hat{\delta}g_{\mu}}{\sigma^{2}}+\sigma_{w}^{2}\x_{\nu}^{\transpose}\left[I-2M^{-1}\sigma_{w}^{2}\sum^n_{\mu=1}\frac{\hat{\delta}g_{\mu}}{\sigma^{2}}\x_{\mu}\x_{\mu}^{\transpose}\right]^{-1}\x_{\nu}-\sigma_{w}^{2}||\x_{\nu}||^{2} 
    \\ \nonumber 
    \hat{\delta}g_{\nu} &= \left(g_{\nu}-\Delta g_{\nu}\right)-\sum^n_{\mu,\mu'=1}K\left(\x_{\nu},\x_{\mu}\right)\tilde{K}_{\mu,\mu'}^{-1}\left(g_{\mu'}-\Delta g_{\mu'}\right) 
\end{align}
The above non-linear equation for the quantities $\hat{\delta} g_{1}, \dots, \hat{\delta} g_{n}$ could be solved numerically, with the most numerically demanding part being the inverse of $\tilde{K}_{\mu,\nu} = K(\x_{\mu},\x_{\nu}) + \sigma^2 \delta_{\mu,\nu}$ on the training set. 

\begin{figure}[h]
\vspace*{-0.1in}
\begin{center}

\includegraphics[height=8cm]{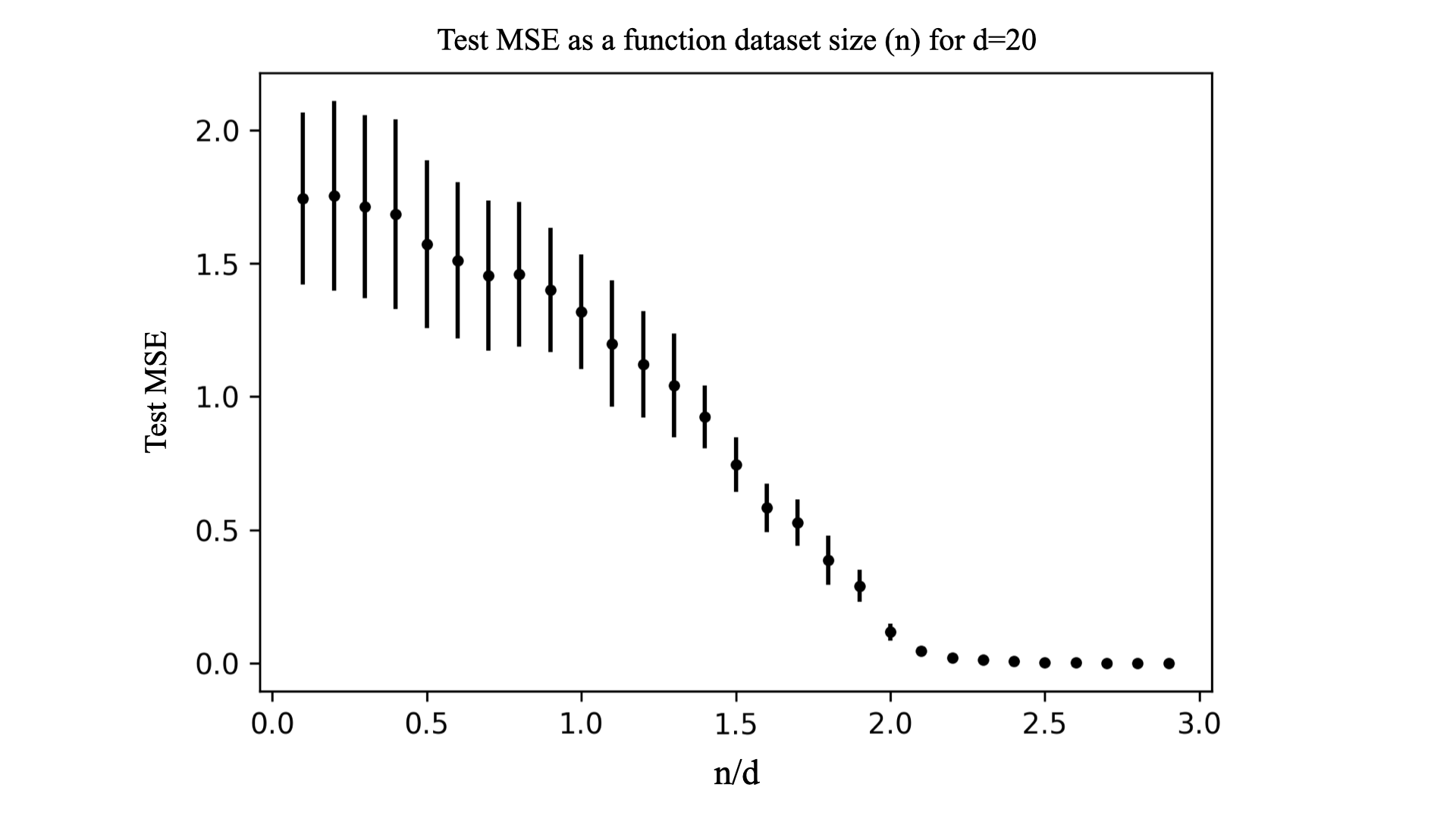}

\end{center}
\vspace*{-0.25in}
\caption{ Test MSE as a function of $n/d$ for the phase retrieval model as predicted by our self-consistent equation at the saddle-point level (without any EK-type approximation). 
Train and test data are drawn uniformly from the $d=20$ hypersphere with radius $1$. 
The graph shows the median test MSE of 60 different data sets. Our approach captures the desired $n = 2d$ threshold value \cite{mannelli2020optimization} whereas lazy-learning/GP will predict a cross over at $n=O(d^2)$. 
\label{Appfig: Lenka}}
\end{figure}

Figure \ref{Appfig: Lenka} shows the numerical results for the test MSE as obtained by solving the above equations for $\hat{\delta}g$ on the training set, taking $\nu=*$ in these equation together with the self-consistent $\hat{\delta}g_{\mu}$ to find the mean-predictor, and taking the average MSE of the latter over the test set. Both test and train data sets were random points sampled uniformly from a $d$ dimensional hypersphere of radius one. 
The test dataset contained $100$ points and the figure shows the test MSE as a function of $n/d$ where $d=20$, $\sigma^2_w=1$, $\sigma^2 = 2.76\cdot 10^{-6}$, $M=4d$, and $w^*_i$ drawn from $\cN(0,1)$. 
The non-linear equations were solved using the Newton-Krylov algorithm together with gradual annealing from $\sigma^2=1$ down to the above values. The figure shows the median over $60$ data sets.
Remarkably, our self-consistent approach yields the expected threshold values of $n/d=2$ \cite{mannelli2020optimization} separating good and poor performance. 
Discerning whether this is a threshold or a smooth cross-over in the large $d$ limit is left for future work. 

Turning to analytics, one can again employ the EK approximation as done for the CNN. However taking $\sigma^2$ to zero invalidates the EK approximation and requires a more advance treatment as in Ref. \cite{Cohen2019}. We thus leave an EK type analysis of the self-consistent equation at $\sigma^2=0$ for future work and instead focus on the simpler case of finite $\sigma^2$ where analytical predictions can again be derived in similar fashion to our treatment of the CNN. 

To simplify things further, we also commit to the distribution $[\x_{\mu}]_i  \sim \mathcal{ N}(0,1/d)$. In this setting $K(\x, \x')$ has two distinct eigenvalues w.r.t. to this measure, the larger one ($\lambda_0=2M^{-1} \sigma_w^4 \left(\frac{2}{d^2}+\frac{1}{d}\right)$) associated with $f(\x)=||\x||^2$ and a smaller one ($\lambda_2=2M^{-1} \sigma_w^4\frac{2}{d^2}$) associated with $x_i x_j$ (with $i \neq j$) and $\sum_i a_i x_i^2$ (with $\sum_{i=1}^d a_i = 0$) eigenfunctions. 

Next we argue that provided the discrepancy is of the following form 
\begin{align}
\hat{\delta}g_{\mu} &= \alpha g(\x_{\mu}) + \beta \sigma_w^2 ||\x||_{\mu}^2
\end{align}
then within the EK limit the target shift is also of the form of the r.h.s. with $\alpha_\Delta$ and $\beta_\Delta$ and the target shift equations reduce to two coupled non-linear equations for $\alpha$ and $\beta$. 
Following the EK approximation, we replace all $\sum_{\mu}$ in the target shift equation with $n\int d\mu_x$ and obtain 
\begin{align}
    \Delta g(\x) = -\frac{n}{\sigma^{2}}\int d\mu_{x'}K\left(\x,\x'\right)\hat{\delta}g(\x') + \sigma_{w}^{2}\x^{\transpose}\left[I-2M^{-1}\sigma_{w}^{2}\frac{n}{\sigma^{2}}\int d\mu_{x'}\hat{\delta}g(\x')\x'\x'^{\transpose}\right]^{-1}\x-\sigma_{w}^{2}\norm{\x}^{2}
\end{align}
Next we note that the $i\neq j$ element of the matrix $\int d\mu_x g(\x) \x \x^\transpose$ is given by 
\begin{align}
    \int d\mu_{x}g(\x)x_{i}x_{j}=\int d\mu_{x}\left(\left(\w_{*}\cdot\x\right)^{2}-\sigma_{w}^{2}\norm{\x}^{2}\right)x_{i}x_{j}=2d^{-2}w_{i}^{*}w_{j}^{*}
\end{align}
whereas for $i=j$ we obtain 
\begin{align}
    \int d\mu_x g(\x) x_i x_i &= 
    3 d^{-2}((w^*_i)^2-\sigma_w^2)+\sum_{j\neq i} d^{-2}((w^*_j)^2-\sigma_w^2) \\ \nonumber &= 3 d^{-2}((w^*_i)^2-\sigma_w^2)+\sum_{j} d^{-2}((w^*_j)^2-\sigma_w^2) - d^{-2}((w^*_i)^2-\sigma_w^2) \\ \nonumber
    &= 2 d^{-2}((w^*_i)^2-\sigma_w^2)
\end{align}
taking this together with the simpler term ($\int d \mu(\x) \frac{\beta}{\alpha} \sum_i x_i x_i \x \x^\transpose=\frac{\beta(d+2)}{\alpha d^2}I$)
\begin{multline}
\label{AppEq:DeltaGEK}
    \Delta g(y) = -\frac{n}{\sigma^{2}}\int d\mu_{x'}K(\x,\x')\left[\alpha g(\x')+\beta\sigma_{w}^{2}\norm{\x'}^{2}\right] + \dots \\
    \sigma_{w}^{2}\x^{\transpose}\left[I-2\lambda_{2}\frac{n}{\sigma_{w}^{2}\sigma^{2}}\sigma_{w}^{2}d^{-2}\left[\w_{*}\w_{*}^{\transpose}-\sigma_{w}^{2}\left(1-\frac{\beta(d+2)}{2\alpha}\right)I\right]\right]^{-1}\x-\sigma_{w}^{2}\norm{\x}^{2}
\end{multline}

Consider the matrix $(\w_* \w_*^\transpose + b I)$, appearing in the above denominator with $b=\left(\frac{\beta \sigma_w^2 (d+2)}{2 \alpha}-\sigma_w^2\right)$, and note that 
\begin{align}
\label{AppEq:Hint}
    \x^\transpose \cdot (\w_* \w_*^\transpose + b I)^n \x &= (\w_* \cdot \x)^2 \frac{ (\norm{\w_*}^2 + b)^n - b^n}{\norm{\w_*}^2} + b^n \norm{\x}^2  
\end{align}
Plugging this equation into a Taylor expansion of the denominator of Eq. \ref{AppEq:DeltaGEK} one finds that all the resulting terms are of the desired form of a linear superposition of $g(\x)$ and $\norm{\x}^2$. 
Considering the first term on the r.h.s. of Eq. \ref{AppEq:DeltaGEK}, $\beta \norm{x}^2$ is already an eigenfunction of the kernel whereas $g(\x)$ can be re-written as
\begin{align}
    g(\x) &\equiv \sum_{i \neq j} w^*_i w^*_j x_i x_j + \sum_i  (w^*_i)^2 x_i^2 - \sigma_w^2 \norm{\x}^2
    \\ \nonumber &= 
    \sum_{i\neq j}w_{i}^{*}w_{j}^{*}x_{i}x_{j}+\sum_{i}\left((w_{i}^{*})^{2}-\frac{\norm{\w_{*}}^{2}}{d}\right)x_{i}^{2}+\left(\frac{\norm{\w_{*}}^{2}}{d}-\sigma_{w}^{2}\right)\norm{\x}^{2}
\end{align}
so that the first two terms on the r.h.s. are $\lambda_2$ eigenfunctions and the last one is a $\lambda_0$ eigenfunctions. 
Summing these different contributions along with the aforementioned Taylor expansion, one finds that $\Delta g (\x)$ is indeed a linear superposition of $g(\x)$ and $\norm{\x}^2$.

Next we wish to write down the saddle-point equations for $\alpha$ and $\beta$. For simplicity we focus on the case where $g(\x)$ is chosen orthogonal to $\norm{\x}^2$ under $d \mu(\x)$, namely $\norm{\w_*}^2 = d\sigma_w^2$. Under this choice the self-consistent equations become
\begin{align}
\label{AppEq:SelfConsistent 1}
    \alpha &= \frac{\frac{\sigma^2}{n}}{\lambda_2 + \frac{\sigma^2}{n}} \left[1- \frac{\alpha c \sigma_w^2}{1-\alpha b c} \left(\frac{1}{1-\frac{ \alpha \norm{\w_*}^2 c}{1-\alpha b c}}\right)+ \frac{n}{\sigma^2} \alpha \lambda_2\right]\\ \nonumber
    \beta &= -\frac{\frac{\sigma^2}{n}}{\lambda_0 + \frac{\sigma^2}{n}}\left[ \frac{\alpha c}{1-\alpha b c} \left(b + \sigma_w^2 \frac{1}{1-\frac{ \alpha \norm{\w_*}^2 c}{1-\alpha b c}}  \right)- \frac{n}{\sigma^2} \beta \lambda_0\right]
\end{align}
where the constant $b$ was defined above and $c=\frac{n \lambda_2}{\sigma_w^2\sigma^2}$. 

Next we perform several straightforward algebraic manipulations with the aim of extracting their asymptotic behavior at large $n$. 
Noting that $c \sigma_w^2 =  \frac{n}{\sigma^2}\lambda_2$, $c \norm{\w_*}^2 = 2 d \frac{n}{\sigma^2}\lambda_2$, and $\alpha c b = -\frac{n}{\sigma^2}(\alpha\lambda_2-2\beta\lambda_0)$ we have 
\begin{align}
\label{AppEq:SelfConsistent 2}
\alpha &= \frac{1}{\lambda_2 + \frac{\sigma^2}{n}} \left[\frac{\sigma^2}{n}- \frac{\alpha\lambda_2}{1+\frac{n}{\sigma^2}(\alpha\lambda_2-2\beta\lambda_0)} \left(\frac{1}{1-\frac{\alpha d \frac{n}{\sigma^2}\lambda_2}{1+\frac{n}{\sigma_2}(\alpha\lambda_2-2\beta\lambda_0)}}\right)+ \alpha \lambda_2\right]\\ \nonumber
\beta &= \frac{-1}{\lambda_0 + \frac{\sigma^2}{n}}\left[ \frac{2\alpha\frac{\lambda_0} {d+2}}{1+\frac{n}{\sigma^2}(\alpha\lambda_2-2\beta\lambda_0)} \left(\frac{\beta (d+2)}{2\alpha} - 1 + \frac{1}{1-\frac{ \alpha d \frac{n}{\sigma^2}\lambda_2}{1+\frac{n}{\sigma^2}(\alpha\lambda_2-2\beta\lambda_0)}}  \right)- \beta \lambda_0\right]
\end{align}
Further simplifications yield 
\begin{align}
\label{AppEq:SelfConsistent 3}
\alpha &= \frac{1}{\lambda_2 + \frac{\sigma^2}{n}} \left[\frac{\sigma^2}{n}- \frac{\alpha\lambda_2}{1+\frac{n}{\sigma^2}(\alpha \lambda_2 - 2 \beta \lambda_0 - \alpha d \lambda_2)} + \alpha \lambda_2\right]\\ \nonumber
\beta &= \frac{-1}{\lambda_0 + \frac{\sigma^2}{n}}\left[ \frac{2\alpha\frac{\lambda_0} {d+2}}{1+\frac{n}{\sigma^2}(\alpha\lambda_2-2\beta\lambda_0)} \left(\frac{\beta (d+2)}{2\alpha} + \frac{\alpha \frac{n}{\sigma^2}d \lambda_2}{1+\frac{n}{\sigma^2}(\alpha \lambda_2 - 2 \beta \lambda_0 - \alpha d \lambda_2)}\right)- \beta \lambda_0\right]
\end{align}
noting that $d \lambda_2 = 2( \lambda_0 - \lambda_2)$ we find 
\begin{align}
\label{AppEq:SelfConsistent 4}
\alpha &= \frac{1}{\lambda_2 + \frac{\sigma^2}{n}} \left[\frac{\sigma^2}{n}- \frac{\alpha\lambda_2}{1-\frac{n}{\sigma^2}(\alpha \lambda_2 + 2 (\beta +\alpha)\lambda_0)} + \alpha \lambda_2\right]\\ \nonumber
\beta &= \frac{-1}{\lambda_0 + \frac{\sigma^2}{n}}\left[ \frac{2\alpha\frac{\lambda_0} {d+2}}{1+\frac{n}{\sigma^2}(\alpha\lambda_2-2\beta\lambda_0)} \left(\frac{\beta (d+2)}{2\alpha} + \frac{\alpha \frac{n}{\sigma^2}d \lambda_2}{1-\frac{n}{\sigma^2}(\alpha \lambda_2 + 2 (\beta +\alpha)\lambda_0)}\right)- \beta \lambda_0\right]
\end{align}
The first equation above is linear in $\beta$ and yields in the large $d$ limit  
\begin{align}
\label{appEq: 1stsim}
\beta &= -\alpha - \frac{\alpha}{d(1-\alpha)} + \frac{\sigma^2}{2\lambda_0 n} 
\end{align}
It can also be used to show that 
\begin{align}
\frac{\alpha\lambda_2}{1-\frac{n}{\sigma^2}(\alpha \lambda_2 + 2 (\beta +\alpha)\lambda_0)} &= (1-\alpha) \frac{\sigma^2}{n}
\end{align}
which when placed in the second equation yields 
\begin{align}
\beta &= \frac{\lambda_0}{\lambda_0 + \frac{\sigma^2}{n}} \frac{ \frac{n}{\sigma^2}\beta(\alpha \lambda_2 - 2 \beta \lambda_0)+2(1-\alpha)\alpha }{1+\frac{n}{\sigma^2}(\alpha\lambda_2-2\beta\lambda_0)} 
\end{align}

At large $n$, we expect $\alpha$ and $\beta$ to go to zero. Accordingly to find the asymptotic decay to zero, one can approximate $\alpha(1-\alpha)\approx \alpha$, and similarly $\alpha/(1-\alpha) \approx \alpha$. This along with the large $d$ limit simplifies the equations to a quadratic equation in $\beta$ 
\begin{align}
\beta^2\frac{2\lambda_0 n}{\sigma^2}\left(1-\frac{\lambda_0}{\lambda_0+\sigma^2/n}\right)+\beta\left(-1-2\frac{\lambda_0}{\lambda_0+\sigma^2/n}\right)+\frac{\lambda_0}{\lambda_0+\sigma^2/n}\frac{\sigma^2}{\lambda_0 n}
\end{align}
which for $\sigma^2/n \ll \lambda_0$ simplifies further into 
\begin{align}
2\beta^2 - 3 \beta + \frac{\sigma^2}{2\lambda_0 n} &= 0
\end{align}
yielding 
\begin{align}
\beta &= \frac{4}{18}\frac{\sigma^2}{\lambda_0 n} \\
\alpha &= \frac{5}{18}\frac{\sigma^2}{\lambda_0 n}
\end{align}
We thus find that both $\alpha$ and $\beta$ are of the order of $\frac{\sigma^2/n}{2 \lambda_0}=n^{-1} \frac{Md\sigma^2}{4\sigma_w^4}$. Hence $n$ scaling as $Md$ ensures good performance. This could have been anticipated as for small yet finite $\sigma^2$ each $n$ can be seen as a soft constrained on the parameters of the DNN and since the DNN contains $Md$ parameters $n=O(Md)$ should provide enough data to fix the student's parameters close to the teacher's.

\end{document}